%% file: main.tex
\documentclass[10pt,twocolumn,letterpaper]{article}

\usepackage{3dv}
\usepackage{times}
\usepackage{epsfig}
\usepackage{graphicx}
\usepackage{amsmath}
\usepackage{amssymb}
\usepackage{gensymb}
\usepackage[table]{xcolor}
\usepackage{float}
\usepackage{booktabs}
\usepackage{enumitem}
\usepackage{nicefrac}
\usepackage{microtype}
\usepackage{subfig}
\usepackage{multirow}
\usepackage{xspace}
\usepackage{sidecap}
\usepackage[export]{adjustbox}
\usepackage[hang,flushmargin]{footmisc}

\newcommand{\momo}[1]{\textcolor{black}{#1}}

\usepackage[pagebackref=true,breaklinks=true,colorlinks,bookmarks=false]{hyperref}

\newcommand{\thename}{ImmerseGAN\xspace}

\usepackage{cleveref}
\crefname{section}{sec.}{secs.}
\Crefname{section}{Sec.}{Secs.}
\crefname{table}{tab.}{tabs.}
\Crefname{table}{Tab.}{Tabs.}
\crefname{figure}{fig.}{figs.}
\Crefname{figure}{Fig.}{Figs.}
\crefname{equation}{eq.}{eqs.}
\Crefname{equation}{Eq.}{Eqs.}

\DeclareMathOperator*{\argmin}{arg\,min}

\definecolor{best}{RGB}{255, 220, 200}
\definecolor{second}{RGB}{255, 255, 200}

\threedvfinalcopy % *** Uncomment this line for the final submission

\def\threedvPaperID{152} % *** Enter the 3DV Paper ID here

\ifthreedvfinal\fi
\begin{document}

%%%%%%%%% TITLE
\title{Guided Co-Modulated GAN for 360\degree{} Field of View Extrapolation}

\author{Mohammad Reza Karimi Dastjerdi$^{1}$\thanks{Research partly done when Mohammad Reza was an intern at Adobe.},
Yannick Hold-Geoffroy$^{2}$,
Jonathan Eisenmann$^{2}$,\\
Siavash Khodadadeh$^{2}$,
Jean-Fran\c{c}ois Lalonde$^{1}$\\
$^1$Université Laval, $^2$Adobe\\
\small{\texttt{\url{https://lvsn.github.io/ImmerseGAN/}}}
} 

\maketitle

\input{0_abstract}
\input{1_intro}

\input{2_relwork}

\input{4_method_new}

\input{5_experiments}

\input{6_applications}

\input{7_discussion}
{\small
\bibliographystyle{ieee_fullname}
\bibliography{pubs}
}

\end{document}

%% file: 0_abstract.tex
\begin{abstract}

We propose a method to extrapolate a 360\degree{} field of view from a single image that allows for user-controlled synthesis of the out-painted content. To do so, we propose improvements to an existing GAN-based in-painting architecture for out-painting panoramic image representation. Our method obtains state-of-the-art results and outperforms previous methods on standard image quality metrics. To allow controlled synthesis of out-painting, we introduce a novel guided co-modulation framework, which drives the image generation process with a common pretrained discriminative model. Doing so maintains the high visual quality of generated panoramas while enabling user-controlled semantic content in the extrapolated field of view. We demonstrate the state-of-the-art results of our method on field of view extrapolation both qualitatively and quantitatively, providing thorough analysis of our novel editing capabilities. Finally, we demonstrate that our approach benefits the photorealistic virtual insertion of highly glossy objects in photographs.
\end{abstract}

%% file: 1_intro.tex
\section{Introduction}
\label{sec:intro}

\begin{figure}[t] 
  \centering
  \includegraphics[width=1.0\linewidth]{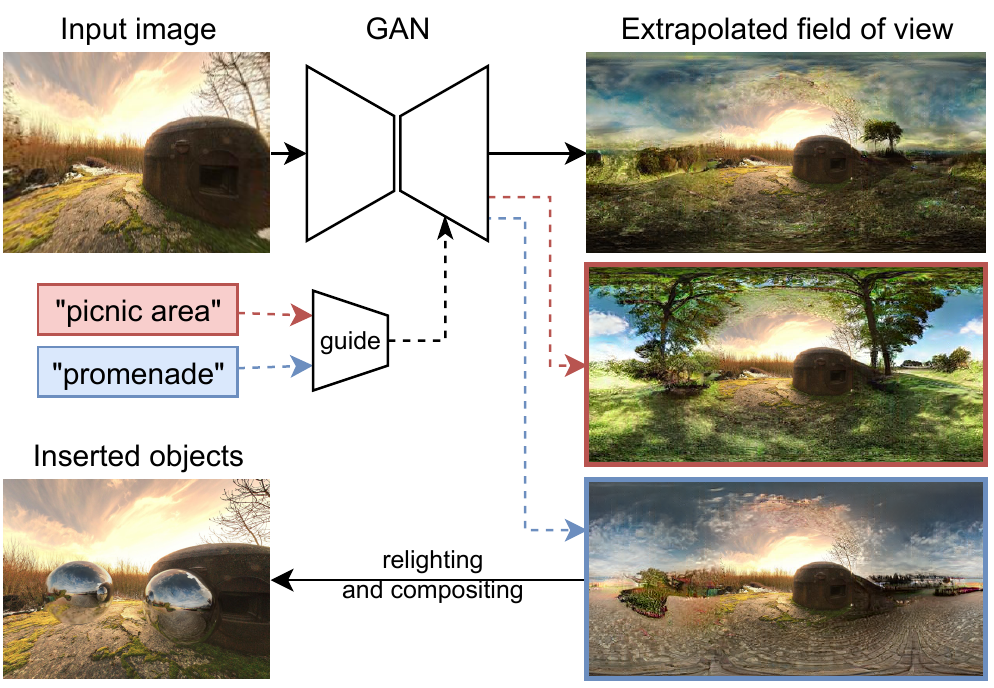}
  \caption{We present a method for high quality 360\degree{} field of view extrapolation. Using a novel guided style co-modulation, a user can drive the generation process through the use of intuitive global labels, allowing the generation of a varied set of results consistent with the input image and semantically matching the desired label. The high quality of our extrapolated panoramas makes them useful for relighting and compositing shiny virtual objects into the scene. }
  \label{fig:teaser}
\end{figure}

Photographs show a glimpse of reality captured when the shutter is pressed: they are but a small window on a full 360\degree{} scene. Despite the limits of cameras, one can easily imagine the full scene in which the image was captured: surely there is a large tree casting this shadow on the lawn; undoubtedly there must be other vehicles passing by this busy street. In computer vision, extrapolating content outside the frame boundaries is known as image out-painting. 

While image synthesis methods~\cite{efros1999texture,efros2001image,barnes2009patchmatch} have long been used as a solution to this problem, more recently learning-based methods which leverage learned priors for this task~\cite{lin2019coco} have been shown to yield more promising results. For example, methods have been trained to generate images that would likely arise if one were to continuously pan (\ie, translate) the camera~\cite{cheng2021out,lin2021infinitygan,yang2019very,liu2021infinite}. \momo{These methods expand the field of view (FOV) solely in front of the camera assuming the scene is a planar grid.}

% Extending the image beyond its sensed pixels has been tackled as an image-to-image task in the recent literature \cite{cheng2021out,lin2021infinitygan,yang2019very,liu2021infinite}. These methods are interested in hallucinating translations (or panning) in existing photographs. 

We instead consider the case of generating the entire 360\degree{} around the camera, \ie, what would happen if one would \emph{rotate} the camera about its center of projection. In other words, we wish to extrapolate the FOV of the camera to span a sphere around the camera. In the literature, generative adversarial networks have emerged as the method of choice for image generation and extrapolation~\cite{somanath2021hdr,song2018im2pano3d}. More recently, \momo{Hara et al.}~\cite{Hara_Mukuta_Harada_2021} explicitly enforce symmetries in FOV extrapolation. In parallel, CoModGAN~\cite{zhao2021comodgan} proposed a method for conditioning a StyleGAN2 generator~\cite{karras2020_stylegan2} and demonstrated that impressive in-painting results can be achieved even when large portions of the image are masked out. 

In this paper, we present an approach which leverages the CoModGAN architecture and makes it suitable to the problem of FOV extrapolation, which we frame as an image \emph{out}-painting task. We show that our proposed changes are critical to achieve state-of-the-art results in this context. Because it provides an immersive view of the entire 360\degree{} from an image, we name our approach \thename. While this technique generates high quality panoramas, it allows for very limited editing capabilities. Indeed, a user can generate new out-painting results by sampling different style vectors (\ie, running the mapping network on different random inputs), but they offer no explicit way of controlling the output since they are only conditioned on the input image. 

Therefore, we also present an approach that enables class-driven editability for FOV extrapolation (\cref{fig:teaser}). To do so, we introduce a novel ``guidance'' mechanism for style co-modulation in our \thename. Our approach relies on a discriminative network, the \emph{guide}, pre-trained for scene classification on a standard labeled dataset of regular photographs~\cite{zhou2017places}. Its output latent vector drives the style co-modulation mechanism in the generator. After training, a user can simply determine a target label, and the best matching latent vector for the guide is found via a simple and fast optimization procedure which does not require back-propagating through the synthesis network as is often the case. The resulting optimized latent co-modulates the style of the generator, which produces an output panorama that both 1) extrapolates a scene which semantically matches the desired label; and 2) seamlessly blends with the input image. 

Our contributions are summarized as follows. First, we present an end-to-end trainable pipeline specifically tailored to the 360\degree{} FOV extrapolation task, which generates high quality panoramas from a single input image with a limited FOV. Second, we introduce a novel \emph{guided} co-modulation mechanism which leverages a pre-trained discriminative model to guide the extrapolation process and allow users to determine the semantic content of the out-painted pixels. Third, we demonstrate state-of-the-art results both quantitatively and qualitatively which outperform previous methods, and provide a thorough evaluation of our novel editing capabilities. Finally, we demonstrate that our approach can be used for realistically inserting virtual objects into photos. 

% %
% \begin{itemize}[noitemsep,topsep=0pt]
%     \item We propose an end-to-end trainable pipeline specifically tailored to generate high-quality and high-resolution 360\degree{} panoramas from a single image with limited field-of-view; 
%     \item We develop a novel ``guided'' co-modulation mechanism which leverages a pre-trained discriminative model to guide the extrapolation process and allow users to determine the semantic content of the out-painted pixels; 
%     \item We demonstrate our state-of-the-art results both quantitatively using the FID measure and qualitatively on several applications including virtual object insertion. 
% \end{itemize}

%% file: 2_relwork.tex
\section{Related work}
\paragraph{Image synthesis}
The generation of photorealistic synthetic images has been studied for several decades~\cite{efros1999texture,efros2001image,barnes2009patchmatch}.
Recently, most image synthesis methods have been propelled by Generative Adversarial Networks (GANs)~\cite{goodfellow2014generative} due to their unparalleled capacity to model natural images. Methods evolved rapidly from the low-resolution DCGAN~\cite{radford2015unsupervised} to the high-quality results of StyleGAN and its successors~\cite{karras2019_stylegan,karras2020_stylegan2,Karras2021_stylegan3}, which greatly push the photorealism of synthesized images. 
Despite their mesmerizing results, these unconditional GANs are formulated to generate random images, with no or limited control over the generated image. 

To solve this problem, conditional GANs \cite{mirza2014conditional} on images were proposed~\cite{isola2017image,wang2018high}. These methods train an hourglass-like encoder-decoder architecture to translate an image from a specific domain to another, such as day to night. In contrast, Shocher et al.~\cite{shocher2020semantic} propose to leverage VGG~\cite{simonyan2014very}, a discriminative model trained on a large-scale dataset of images, to control the images synthesized by the generator. We draw inspiration from the latter to take advantage of transfer learning to add editing capabilities to our model. 
%From these ideas stemmed multiple. SPADE~\cite{park2019semantic} notably realized that segmentation maps were detrimental to the normalization layer and proposed to re-inject the statistics of the input throughout the network. 
%LoGANv2~\cite{oeldorf2019loganv2} conditional logo generation

Several methods were proposed to add editing capabilities~\cite{chai2021using,bau2019semantic} to GANs, and to invert pre-trained models~\cite{xia2021gan,abdal2019image2stylegan,zhu2020domain}. Another approach recently explored is to perform editing directly on the latent space instead of image space, see \cite{zhang2021survey} for a survey. In a nutshell, these methods discover properties of the latent space of a pre-trained GAN. Work in this field has focused on uncovering semantic directions, either in a supervised \cite{goetschalckx2019ganalyze,jahanian2019steerability,shen2020interpreting} or unsupervised way~\cite{harkonen2020ganspace}, or on spatially editing images using GANs~\cite{zhu2016generative,bau2018gan,park2019semantic,cheng2020segvae}. We draw inspiration from these insightful ideas, and propose a method for editing the generated image that does not require a labelled dataset \momo{of panoramas} while providing explicit control over the desired output (as opposed to unsupervised). Our method also does not require inversion, a process which recovers good but often imperfect reconstructions~\cite{zhu2020domain}. 

% We remark that several of these work rely on the GAN inversion producing a perfect reconstruction of the image, which is not always the case and often leads to visible seams between the observed and hallucinated regions of the resulting panorama. We devise a method that produces a panorama by seamlessly embedding the original image in a single feed-forward execution. 
% --> JF: this is a dangerous claim to make since we do not demonstrate  it. 

\paragraph{Inpainting}

The goal of image inpainting is to fill the gaps in a partially observed (or masked) image~\cite{efros1999texture,efros2001image,barnes2009patchmatch}. 
Recently, many approaches were developed to perform inpainting using GANs, for example using a patchGAN discriminator~\cite{demir2018patch} or an attention mechanism~\cite{yu2018generative}. The current state-of-the-art, CoModGAN~\cite{zhao2021comodgan}, proposes to convert a StyleGAN generator to a conditional model using co-modulation. In our method, we propose improvements to the CoModGAN architecture to generate panoramas. Inpainting is also needed in the related task of novel view synthesis~\cite{huang2020semantic,abbasi2019deep,qi2018semi,jo2021n}, where filling disocclusions and generating new textures is necessary to produce a plausible image. 

\paragraph{Field of View (FOV) extrapolation}

Similar to inpainting, FOV extrapolation can be framed as outpainting, which aims at extending the images beyond the original camera frame. Several GAN-based methods have been proposed~\cite{sabini2018painting}, notably by extending the original image over the peripheral vision~\cite{kimura2018extvision,ying2020180}. Other works propose to generate realistic translations (or pans) of the camera \cite{cheng2021out,lin2021infinitygan,yang2019very}. \momo{These methods expand the scene on a planar grid, which allows for interesting and useful artistic effects. However, this planar grid representation cannot model the entire 360\degree{} field of view as this is represented by a \emph{sphere} rather than a plane.} 

Many methods have tackled the full 360\degree{} FOV extrapolation. PIINET~\cite{han2020piinet} uses a cubemap projection to limit the amount of distortion when performing inpainting directly on 360\degree{} panoramas. 
Im2Pano3D~\cite{song2018im2pano3d} estimates a plausible 360\degree{} segmentation map from a regular image, giving hints about the content around the camera.
\momo{Sumantri~\etal}~\cite{sumantri2020360} present a method to recover a 360\degree{} panorama from four images taken uniformly along the horizon. \momo{Srinivasan~\etal~\cite{lighthouse} estimate a 360\degree{} panorama for any location of a scene using a narrow-baseline stereo image pair.}
Closer to our work, \momo{Akimoto~\etal}~\cite{akimoto2019360}, and \momo{Somanath and Kurz}~\cite{somanath2021hdr} suggest using a GAN to generate 360\degree{} panoramas. \momo{Hara ~\etal}~\cite{Hara_Mukuta_Harada_2021} goes one step further and propose to leverage symmetries usually present in environments to generate a full 360\degree{} panorama from a single image. \momo{Recently, Akimoto~\etal~\cite{akimoto2022diverse} utilize a transformer-based architecture \cite{esser2021taming} to predict environment maps for creating 3DCG backgrounds.}
% We compare our method against these other GAN-based techniques and show that we significantly outperform them both visually and quantitatively. 

%% file: 4_method_new.tex
\section{Method}

\subsection{Co-modulated GANs}

\begin{figure}[t]
    \centering
    \footnotesize
    \renewcommand{\tabcolsep}{0pt}
    \begin{tabular}{cc}
    \includegraphics[height=4.5cm]{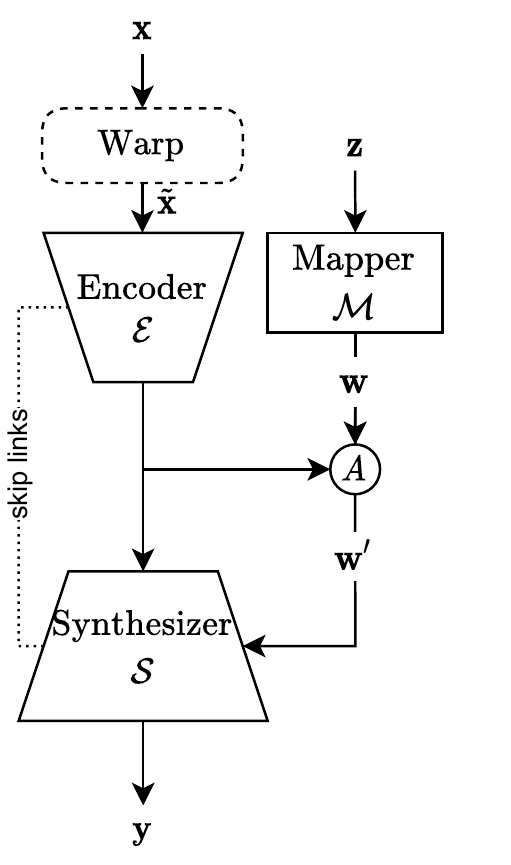} & 
    \includegraphics[height=4.5cm]{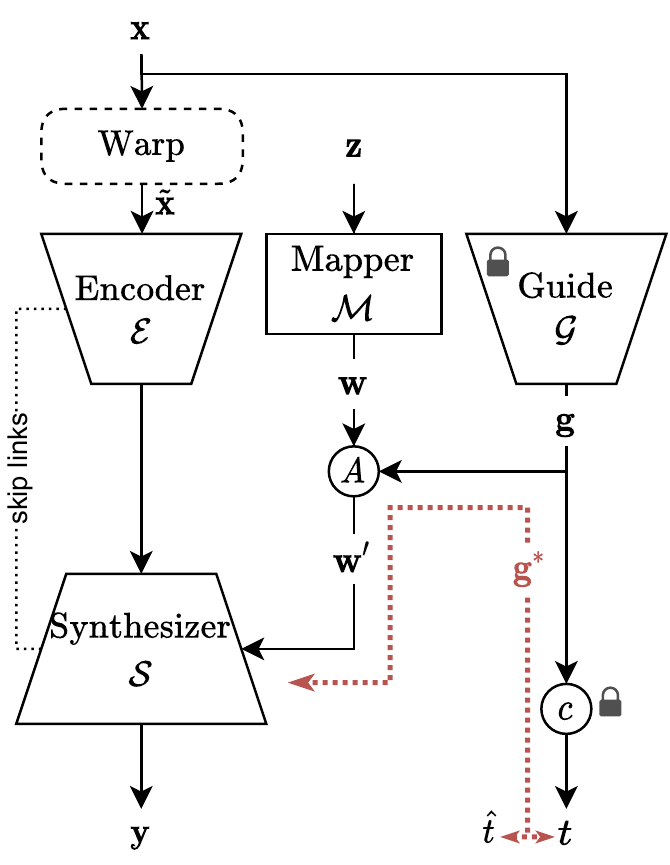} \\ 
    (a) Co-modulation~\cite{zhao2021comodgan} & (b) Guided co-modulation
    \end{tabular}
    \includegraphics[width=1.8cm]{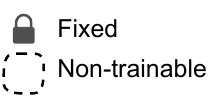}
    \caption[]{Overview of our guided co-modulation method. The input image $\mathbf{x}$ is first warped according to its (known) camera parameters to a 360\degree{} panorama $\tilde{\mathbf{x}}$ via a (non-trainable) warp. (a)~As proposed in CoModGAN~\cite{zhao2021comodgan}, an encoder $\mathcal{E}$ co-modulates (with the mapper $\mathcal{M}$) the synthesis network $\mathcal{S}$ according to $\tilde{\mathbf{x}}$. Here, $A$ denotes a (learnable) affine transformation. (b)~In contrast, our method guides the co-modulation process with a pre-trained ``Guide'' model $\mathcal{G}$, which operates directly on the input image $\mathbf{x}$. $\mathcal{G}$ produces a latent vector $\mathbf{g}$ which is used for co-modulation, and which can also be run through a classification layer $c$ to produce a label map $t$. This gives the user control over the generation process by defining a new label $\hat{t}$ and optimizing to determine a modified $\mathbf{g}^*$ that modulates the synthesizer to generate the desired label (red dashed line).}
    \label{fig:method}
\end{figure}

Our work builds on the recently-proposed CoModGAN architecture~\cite{zhao2021comodgan}, which we briefly present here for completeness and illustrate in \cref{fig:method}a. The warped image $\tilde{\mathbf{x}}$ is given as input to an encoder $\mathcal{E}$, whose output is combined to that of the mapper $\mathcal{M}$ via an affine transform $A$:
\begin{equation}
    \label{eqn:co-modulation}
    \mathbf{w}^\prime = A(\mathcal{E}(\tilde{\mathbf{x}}), \mathcal{M}(\mathbf{z})) \,,
\end{equation}
where $\mathbf{z} \sim \mathcal{N}(0, \mathbf{I})$ is a random noise vector, and  $\mathbf{w}^\prime$ is the style vector modulating the synthesis network $\mathcal{S}$. The output of $\mathcal{E}(\tilde{\mathbf{x}})$ is provided as the input tensor to $\mathcal{S}$. 

\subsection{\thename}
Although CoModGAN produces high-quality results for image completion, it is not suitable for FOV extrapolation due to two reasons. First, CoModGAN generates random masks during the training that do not represent typical scenarios of FOV extrapolation. Therefore, CoModGAN produces blurry results at test time. 
Second, since it is designed for inpainting in perspective images, it cannot cope with the ``wrap-around'' nature of panoramic images---rotating the generated panorama by 180\degree{} about the vertical axis will reveal a strong vertical edge. Potential solutions to these artifacts have been proposed in the literature, including complex changes such as circular padding~\cite{Hara_Mukuta_Harada_2021} or sphere-projected convolutions~\cite{fernandez2020corners} (dubbed ``EquiConvs''). 

We introduce \thename, which is built on the CoModGAN architecture with three modifications to address the mentioned problems. We replace random masks by generating FOV masks during training. This helps the network generate better quality results by introducing FOV biases. To address the problem of generating vertical edges, we leverage the nature of generative adversarial networks and horizontally shift the generator output before feeding it to the discriminator. This operator encourages the generator to produce panoramas with no discontinuities at the edges. Finally, we modify the architecture to yield a 2:1 aspect ratio to avoid anisotropic upsampling artifacts when mapping the output to the equirectangular representation. 

\subsection{Guided co-modulation}
While our proposed \thename produces state-of-the-art results for FOV extrapolation (\cref{sec:comparison}), the main limitation of this method is the lack of control over the output. Our goal is to leverage the knowledge of a ``guide'' network pre-trained on a large-scale scene classification task. This guide network produces two outputs: a feature vector (used to modulate our generator) and the image class (identifying the image content). For a given image at test time, we optimize the feature vector by backpropagating a user-provided desired class, as shown by the red path in \cref{fig:method}b. We repeat this operation around 2000 iterations, until convergence. We then feed this optimized feature vector to the generator. \\
We assume the guide network $\mathcal{G}$ has been trained for classification, and produces an intermediate latent vector $\mathbf{g} = \mathcal{G}(\mathbf{x})$ from the input (unwarped) image $\mathbf{x}$. $\mathbf{g}$ is subsequently fed to a classification subnetwork, $t = c(\mathbf{g})$. Here, $t \in \mathbb{R}^N$ is the vector of predicted probabilities over $N$ classes. The guided co-modulated style vector $\mathbf{w}^\prime$ is
\begin{equation}
    \label{eqn:guided-co-modulation}
    \mathbf{w}^\prime = A(\mathcal{G}(\mathbf{x}), \mathcal{M}(\mathbf{z})) \,.
\end{equation}

We can tune the output appearance $\mathbf{y}$ by modifying the latent vector $\mathbf{g}$ to represent \emph{another} class by optimizing a one-hot vector $\hat{t}$ with the desired class
\begin{equation}
    \label{eqn:opt}
    \mathbf{g}^* = \argmin_\mathbf{g} \ell (c(\mathbf{g}), \hat{t}) \,,
\end{equation}
where $\ell$ is a binary cross-entropy loss function. A panorama, whose appearance outside the FOV of the input image should better match $\hat{t}$ is produced by replacing $\mathbf{g} \leftarrow \mathbf{g}^*$, \ie, 
\begin{equation}
    \label{eqn:opt-comod}
    \mathbf{w}^\prime = A(\mathbf{g}^*, \mathcal{M}(\mathbf{z})) \,.
\end{equation}
%
% This process is illustrated by the red dashed line in \cref{fig:method}b. 
In contrast to existing editing methods, our pipeline does not require training the guide model on the domain output by the synthesizer (panoramas), any model pre-trained on regular images can do. It also does not require any analysis of the learned latent space. 

% \todo{why didn't we compare against other latent code editing? because they don't apply directly, in some cases you need all pairs of 365 (365 squared) classes analyzed, which is prohibitive. Also, pretrained models are made for crops and not panoramas.}

\subsection{Implementation details}
\label{sec:implementation}

%As in \cite{zhao2021comodgan}, the mask of the input panorama $\tilde{\mathbf{x}}$ (corresponding to regions outside the field of view of the camera) is concatenated with the input before being fed to the network. Finally, the known pixel values in $\tilde{\mathbf{x}}$ are composited over the output $\mathbf{y}$ before being passed to the discriminator (not shown in \cref{fig:method}). The resulting network is trained with regular discriminator losses, without requiring direct guidance like an $\ell_1$ loss as in \cite{isola2017image}.
% Momo's comment: I remove the todo since the loss is not novel

We assume the camera parameters to be known a priori. Other works, \ie~\cite{hold2017perceptual,xian2019uprightnet} could be used to estimate them otherwise.
Similar to \cite{zhao2021comodgan}, we concatenate the image's FOV mask with the input panorama $\tilde{\mathbf{x}}$ before being fed to the network. The known pixel values in $\tilde{\mathbf{x}}$ are copied over the output $\mathbf{y}$ before being passed to the discriminator (not shown in \cref{fig:method}). 
The guide network $\mathcal{G}$ is an 18-layer ResNet~\cite{he2016deep} pre-trained on the Places365~\cite{zhou2017places} dataset. The output of the guide network, $\mathbf{g} \in \mathbb{R}^{512}$, is produced by the penultimate layer of the ResNet, after the 1D-flatten. \momo{Please refer to supp. materials for more details regarding the network architecture.}

%% file: 5_experiments.tex
\section{Field of view extrapolation experiments}
\label{sec:experiments}

% \yhg{note: make extra-sure it's clear for the reader that the input image MUST be in the center of the panorama. If the hallucinated region does not fit the input image due to an edit, it is NOT a failure case.}
% We \emph{enforce} the input image to appear unchanged in the center of the panorama.

% \input{fig_qualresults}
\input{fig_qualresults_rot1_}
\subsection{Datasets}
\label{sec:dataset-training}

To generate training data, we employ a similar strategy as previous works~\cite{Hara_Mukuta_Harada_2021,somanath2021hdr} and extract rectified crops from a large dataset of 360\degree{} panoramas. We leverage a dataset of 250,000 (unlabelled) panoramas obtained from 360Cities\footnote{\tiny \url{https://www.360cities.net}\scriptsize, acquired under proprietary license with right to publish.}. The dataset is split into 248K/1K/1K train/validation/test subsets. When training (and for validation), random crops are computed on the fly to ensure as diverse a set as possible. For this, the parameters are sampled as $h_\theta \sim \mathcal{U}(40, 120)$ for the FOV and $\beta \sim \mathcal{N}(0,30)$ for the elevation angle, where $\mathcal{U}$ and $\mathcal{N}$ are uniform and normal distributions respectively. After sampling, $\beta$ is clipped to $[-30,30]^\circ$. For the test set, we use a set of 1,170 panoramas balanced between outdoor and indoor scenes that are not used during the training. We extract a set of 4,680 images at fixed FOV $h_\theta \in \{40, 60, 90, 120\}^\circ$, with elevation $\beta=0^\circ$. We also extract another set of 1,170 images, dubbed ``mixed'', where the parameters are sampled randomly (according to the same distribution as above).
% The test set contains 5,850 images and will be released publicly.

\begin{table}[t]
    \centering
    \footnotesize
    \setlength{\tabcolsep}{3pt}
    \begin{tabular}{lccccc}
    \toprule
    Method & $40^\circ$ & $60^\circ$ & $90^\circ$ & $120^\circ$ & Mixed\\
    \midrule
    pix2pixHD~\cite{wang2018high} & 226.41 & 163.19 & 100.18 & 58.09 & 122.18  \\
    Symmetry~\cite{Hara_Mukuta_Harada_2021}-R & 106.89 & 79.86 & 64.91 & 62.24 & 62.28 \\
    Symmetry~\cite{Hara_Mukuta_Harada_2021}-G & 92.97 & 75.66 & 61.60 & 62.15 & 56.04 \\
    CoModGAN-1x1~\cite{zhao2021comodgan} & 82.80 & 68.02 & 47.36 & 37.84 & 48.68 \\
    CoModGAN-2x1~\cite{zhao2021comodgan} & 79.05 & 67.72 & \cellcolor{second}{46.33} & 35.34 & 47.91 \\
    \thename (ours) & \cellcolor{second}{37.90} &  \cellcolor{second}{35.55} & \cellcolor{best}{32.25} & \cellcolor{best}{28.92} & \cellcolor{best}{32.48} \\
    % \begin{tabular}{@{}l@{}}Guided \\ \thename (ours)\end{tabular} 
    \thename-Guided (ours) & \cellcolor{best}{37.15} & \cellcolor{best}{34.65} & \cellcolor{best}{32.41} & \cellcolor{second}{32.97} & \cellcolor{second}{35.01}\\
    \bottomrule
    \end{tabular}
    % }
    \caption{FID computed on the test set for various methods at varying FOVs. The $40^\circ$, $60^\circ$, $90^\circ$, and $120^\circ$ columns report results on images with $\beta=0$ and the corresponding fixed FOV. The ``mixed'' column represents a set of images with mixed parameters, see text for details. Each row is color-coded as \colorbox{best}{best} and \colorbox{second}{second best}.}
    \label{tab:quantitative-results}
\end{table}

\subsection{Evaluation of field of view extrapolation}
\label{sec:comparison}

We now proceed to compare our method quantitatively and qualitatively with the state-of-the-art on our test set. First, we train a pix2pixHD~\cite{wang2018high} model on our training set. Instead of an all-black mask, we fill the background with random noise as suggested in \cite{somanath2021hdr}, which also improved results in our experiments. To compare with the symmetry-based approach of Hara~\etal~\cite{Hara_Mukuta_Harada_2021}, the authors graciously ran their code on our test set. We also train two different versions of CoModGAN~\cite{zhao2021comodgan}: an original version without any modifications (dubbed ``CoModGAN-1x1'') and a modified architecture to work on 2:1 aspect ratios (``CoModGAN-2x1'').
% (for transparency, this can potentially negatively bias this method since it was not trained on the same data as ours). 
All other methods were trained on our train set (\cref{sec:dataset-training}) for five days on eight V100 GPUs. All methods produce outputs of $512 \times 256$ resolution, which was chosen to conduct a fair quantitative evaluation against Hara~\etal~\cite{Hara_Mukuta_Harada_2021}. However, our method can generate outputs up to $2048 \times 1024$ resolution (see supp. materials).

Comparative quantitative results are reported in \cref{tab:quantitative-results}, and corresponding qualitative results in \cref{fig:qual-comparison}. Results are grouped according to the test subsets (\cf \cref{sec:dataset-training}). Unsurprisingly, all methods perform better when the FOV of the input image increases (the task is indeed easier since fewer pixels need to be out-painted). We note that pix2pixHD~\cite{wang2018high} results in catastrophic FIDs at $h_\theta < 120^\circ$, which could be partially explained by the mode collapse visible in \cref{fig:qual-comparison} where the generated panoramas all have the same specific pattern on the left. 
For \cite{Hara_Mukuta_Harada_2021}, we report both their ``reconstruction'' and ``generation'' settings, denoted by the ``-R'' and ``-G'' suffixes respectively. The ``generation'' results being superior to ``reconstruction'' in \cref{tab:quantitative-results}, only those results are therefore shown in \cref{fig:qual-comparison}. Overall, \thename, or its Guided version, yield the lowest FIDs throughout all test subsets. 
%Note that CoModGAN yields similar FIDs to ours at $h_\theta=120^\circ$, but its performance severely degrades with smaller FOVs. 
In addition, as shown in \cref{tab:quantitative-results}, only changing the architecture of CoModGAN to 2:1 ratio is not enough to achieve SOTA: all of our proposed modifications are necessary. For the rest of the paper, we use CoModGAN-2x1 for the comparisons. Despite known limitations with FID~\cite{chong2020effectively}, this behavior is qualitatively corroborated in \cref{fig:qual-comparison}.  

\paragraph{CoModGAN baseline vs \thename}
To demonstrate the importance of our proposed changes in \thename over CoModGAN~\cite{zhao2021comodgan}, we present in \cref{fig:comodgan-vs-ours} a qualitative and quantitative comparison for generated panoramas after being rotated by 180\degree{} in azimuth. CoModGAN creates a visible vertical seam at the edges, which is not present with ours.

% \begin{table} [t]
%     \centering
%     \scriptsize
%     \setlength{\tabcolsep}{2pt}
%     \begin{tabular}{lcccc}
%     \toprule
%      & Gardner~\etal~\cite{gardner2017learning} & EnvMapNet~\cite{somanath2021hdr} & Akimoto~\etal~\cite{akimoto2022diverse} & \thename (ours)\\
%     \midrule
%     FID & 197.4 & 52.7 & 46.15 & \textbf{42.78}\\
%     \bottomrule
%     \end{tabular}
%     \caption{Comparison with lighting estimation methods on the test split of Laval Indoor HDR dataset~\cite{gardner2017learning}}
%     \label{tab:sFID}
% \end{table}
\begin{table}
    \small
    \centering
    \setlength{\tabcolsep}{2pt}
    \begin{tabular}{cc}
    \toprule
    Method & FID \\
    \midrule
  Gardner~\etal~\cite{gardner2017learning} & 197.4 \\
  EnvMapNet~\cite{somanath2021hdr} & 52.7 \\ Akimoto~\etal~\cite{akimoto2022diverse} & 46.15 \\
  \thename (ours) & \textbf{42.78}\\
    \bottomrule
    \end{tabular}
    \caption[]{Comparison with lighting estimation methods on the test split of Laval Indoor HDR dataset~\cite{gardner2017learning}.}
    \label{tab:sFID}
\end{table}

\paragraph{Comparison with lighting estimation methods}
Since some lighting estimation methods predict a 360\degree panorama for image-based lighting, we conduct a quantitative analysis to compare their performance to our proposed method. We used the Laval indoor HDR dataset \cite{gardner2017learning} test set and followed the same protocols as in \cite{somanath2021hdr, akimoto2022diverse}. The results are reported in \cref{tab:sFID}. Although all other methods have the advantage of being trained on the train set of the Laval indoor HDR dataset, our method still managed to have the best performance even against the recently published method of Akimoto~\etal~\cite{akimoto2022diverse}.
\input{fig_comodgan_vanilla_vs_360}
\section{Guided editing experiments}
\label{sec:editing-experiments}

\subsection{Qualitative results} We now proceed to evaluate the capabilities of our novel guided co-modulation mechanism to influence the extrapolated FOV. First, several qualitative results are shown in \cref{fig:teaser,fig:qual-editing}. Focusing on \cref{fig:qual-editing}, we observe that the generated outputs obey the semantics of the target labels while seamlessly blending in the input image. We also note that labels are responsible for specific effects, \eg, for outdoor: ``sky'' draws a more dramatic sky, ``lawn'' adds green grass, ``snowfield'' lets it snow, ``promenade'' creates a more open space with paved walk. For indoor: ``entrance hall'' adds doors, ``corridor'' elongates the scene, ``artists loft'' decorates with wall art, and ``throne room'' embellishes with banners. 
 % creates an ambiance by adding neon lighting, and ``hangar'' increases the apparent size of the room (by pushing the back walls farther away, making them appear smaller). 

\input{fig_editingresult}

\subsection{Quantitative analysis} While \cref{fig:qual-editing} showcases the high quality results obtained with the novel editing capabilities of our network, it is not clear that they should work for all combinations of image/target labels. Surely, the semantic consistency between the input image content and the target label should have an impact on the final result. For example, while it is intuitive that a $\text{``field''} \mapsto \text{``lawn''}$ should work (first row of \cref{fig:qual-editing}), what about less meaningful mappings such as $\text{``church''} \mapsto \text{``sky''}$ or $\text{``street''} \mapsto \text{``artist loft''}$? 

To study this, we first record the top-1 label predicted by the guide network for each image in the training set, and retain the 45 most frequent category instances. We manually assign these 45 categories to either of the generic ``indoor'' (22) and ``outdoor'' (23) sets. We then apply our guided co-modulation on 1K random images from the $h_\theta=90^\circ$ test set (\cref{sec:dataset-training}) towards each of these 45 labels (generating 45K panoramas). 
We then extract a ``backwards-looking'' image, that is, an image $\mathbf{x}_\text{back}$ with parameters $h_\theta=90^\circ, \beta=0^\circ$, azimuth angle $\alpha=-180^\circ$, from the generated panorama. Every pixel in this image is generated by our method, as there is no overlap between this image and the input image.

We evaluate whether the generated content is indeed of the desired label. To do so, the image $\mathbf{x}_\text{back}$ is classified by $\mathcal{G}$ to obtain class probabilities $t_\text{back}$. It is considered a success when the target label is contained within the top-10 (top 2.7\% of labels) of $t_\text{back}$. As shown in \cref{tab:editing-percentage}, staying within the domains ($\text{``outdoor''} \mapsto \text{``outdoor''}$, $\text{``indoor''} \mapsto \text{``indoor''}$) generates the best results. Cross-domain scenarios are more difficult because these conversions do not correspond to real-world cases. This also highlights a bias in our train set, which indeed contains more outdoor scenes than indoor, leading to better performance for outdoor (52.8\%) vs. indoor (37.2\%). 

% \Cref{fig:editing-quant}a shows the FID of the $\mathbf{x}_\text{back}$ images for all combinations of ``indoor/outdoor image'' $\mapsto$ ``indoor/outdoor target'' mappings, as well as for ``unedited'' results. Unsurprisingly, 
% Note that the higher FIDs here cannot be compared to \cref{tab:quantitative-results} since they are being computed on fewer images (500 vs. 1,170). 

% and percentage of time for all combinations of $\text{indoor/outdoor image} \mapsto \text{indoor/outdoor target}$ mappings. 
% Unsurprisingly, staying within the domain ($\text{outdoor} \mapsto \text{outdoor}$, $\text{indoor} \mapsto \text{indoor}$) generates the best results, both in terms of diversity (FID) and generation of the desired target (\%). The FID is only slightly increased by the editing compared to the ``unedited'' (image-conditioned) results. Cross-domain scenarios are more difficult because the semantic meaning of these conversions does not correspond to real-world scenarios. This also highlights a bias in our training set, which indeed contains more outdoor scenes than indoors, thus resulting in better performance for outdoors (52.8\%) vs. indoors (37.2\%). 

% Beyond extrapolated image diversity captured by the FID, 

\begin{figure}[t]
\centering
\renewcommand{\tabcolsep}{0pt}
\begin{tabular}{c}
\includegraphics[width=1.0\linewidth,trim={5.8cm 3cm 15.5cm 1cm},clip]{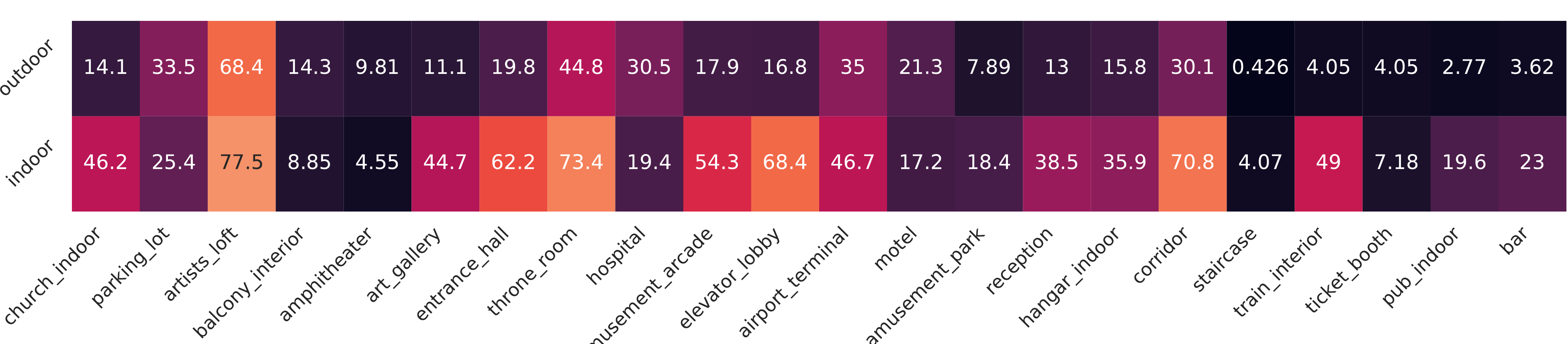} \\
\includegraphics[width=1.0\linewidth,trim={5.8cm 3cm 15.5cm 1cm},clip]{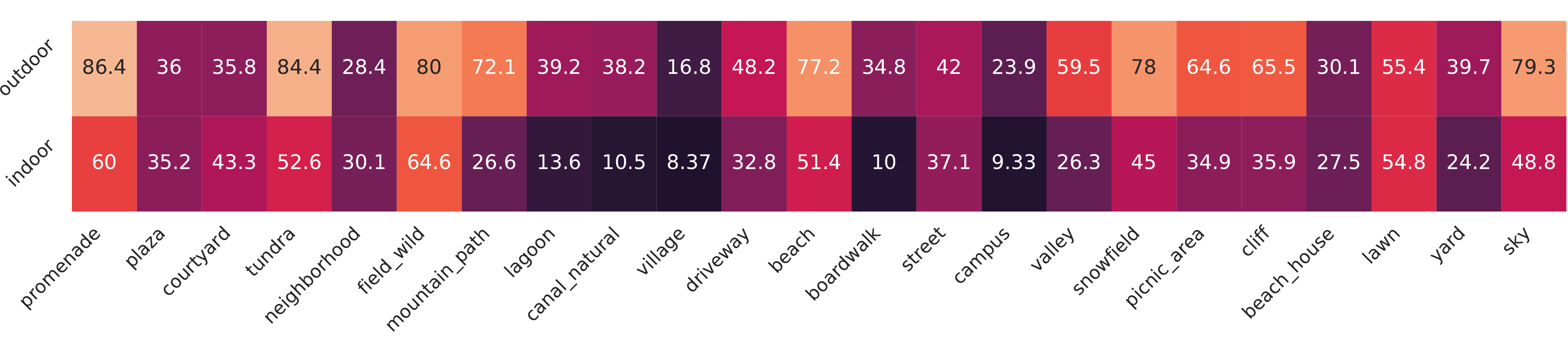} \\
\end{tabular}
% \vspace{-0.5em}
\caption[]{Percentage of success of the guidance process as a function of the target label. From indoor/outdoor categories of input images (rows), we guide the extrapolation results to different indoor (top) and outdoor (bottom) labels and evaluate performance on classifying the extrapolated part. }
\label{fig:editing-details}
\end{figure}

\Cref{fig:editing-details} provides a fine-grained analysis for each type of target label using the classification metric. While many labels offer good performance, others do not generate  results that the guide network can successfully recognize. Those are mostly indoor labels (\eg, ``ticket booth'',  and ``bar''), but the outdoor ``boardwalk'' label also yields lower performance. This label meta-analysis is particularly useful to determine which labels are likely to generate believable results ahead of time, which is confirmed by the visual results in \cref{fig:qual-editing}. 

\begin{SCtable}
\footnotesize
\begin{tabular}{lcc}
\toprule
& \multicolumn{2}{c}{target label} \\
img. label & outdoor & indoor \\
\midrule
outdoor & 52.8\% & 19\% \\
indoor & 34.1\% & 37.2\% \\
\bottomrule
\end{tabular}
\caption[]{Quantitative evaluation for guided editing, computed on an image $\mathbf{x}_\text{back}$ ``looking backwards''.}
\label{tab:editing-percentage}
\end{SCtable}

% \begin{table}[t]
% \centering
% \footnotesize
% \renewcommand{\tabcolsep}{3pt}
% \begin{tabular}{lcc}
% \toprule
% & \multicolumn{2}{c}{target label} \\
% img. label & outdoor & indoor \\
% \midrule
% outdoor & 52.8\% & 19\% \\
% indoor & 34.1\% & 37.2\% \\
% \bottomrule
% \end{tabular}
% \begin{tabular}{cc}
% \begin{tabular}{lccc}
% \toprule
% & \multicolumn{2}{c}{target label} & unedited\\
% img. label & outdoor & indoor & \\
% \midrule
% outdoor & 66.3 & 145.2 & 61.8\\
% indoor & 110.1 & 82.8 & 77.5 \\
% \bottomrule
% \end{tabular}
% &
% \begin{tabular}{lcc}
% \toprule
% & \multicolumn{2}{c}{target label} \\
% img. label & outdoor & indoor \\
% \midrule
% outdoor & 52.8\% & 19\% \\
% indoor & 34.1\% & 37.2\% \\
% \bottomrule
% \end{tabular}
% \\
% (a) FID (600 images) & (b) \% of success
% \end{tabular}
% \caption[]{Quantitative evaluation for guided editing, computed on an image $\mathbf{x}_\text{back}$ ``looking backwards'' in the extrapolated FOV. Percentage of success of the guidance process, indicated by the \% of successful edits according to a classification metric (see text).}
% \label{tab:editing-percentage}
% \end{table}

%

% \begin{figure}[t]
% \centering
% \footnotesize
% \renewcommand{\tabcolsep}{0pt}
% \newcommand{\myheight}{2.05cm}
% \begin{tabular}{ccc}
% \includegraphics[width=0.5\linewidth,trim={5.8cm 3cm 15.5cm 1cm},clip]{figs/editing/quant/confusion_labels_indoor_2xn-top10.pdf} &
% \includegraphics[width=0.5\linewidth,trim={5.8cm 3cm 15.5cm 1cm},clip]{figs/editing/quant/confusion_labels_outdoor_2xn-top10.pdf} \\
% (a) Indoor target labels &
% % \includegraphics[height=\myheight,trim={47.8cm 2cm 11.7cm 1cm},clip]{figs/editing/quant/confusion_labels_outdoor_2xn-top10.pdf} \\
% (b) Outdoor target labels 
% \end{tabular}
% \caption{Percentage of success of the guidance process as a function of the target label. From indoor/outdoor categories of input images (rows), we guide the extrapolation results to different (a) indoor and (b) outdoor labels and evaluate performance on classifying the generated part of the image. }
% \label{fig:quant-editing}
% \end{figure}

\subsection{Comparison to InterFaceGAN}

\begin{figure}[t]
\scriptsize
\centering
\newcommand{\mywidth}{0.23\linewidth}
\setlength{\tabcolsep}{1pt}
\begin{tabular}{lcccc}
& InterFaceGAN & Ours & InterFaceGAN & Ours \\
\rotatebox{90}{\tiny $\mapsto \text{``sky''}$} &
\includegraphics[width=\mywidth]{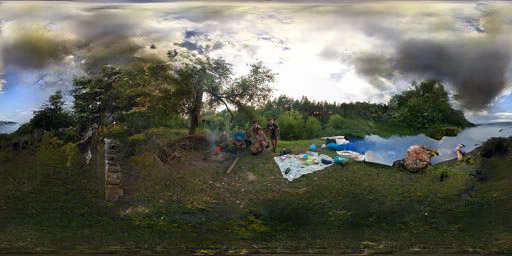} & 
\includegraphics[width=\mywidth]{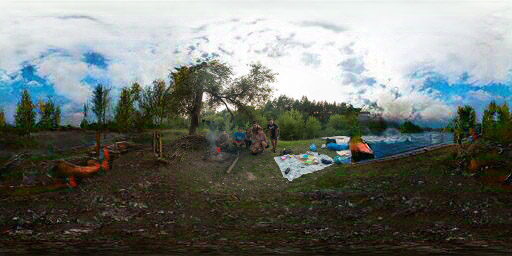} &
\includegraphics[width=\mywidth]{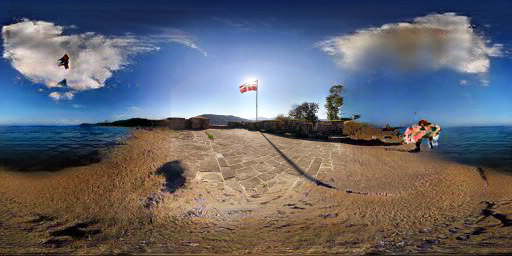} & 
\includegraphics[width=\mywidth]{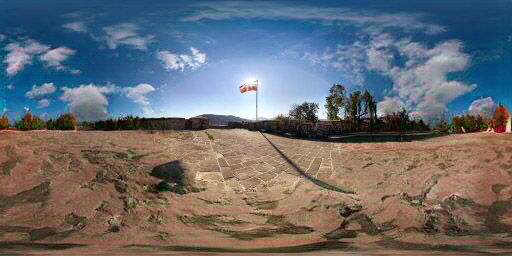} \\ 
\rotatebox{90}{\tiny $\mapsto \text{``snowfld''}$} &
\includegraphics[width=\mywidth]{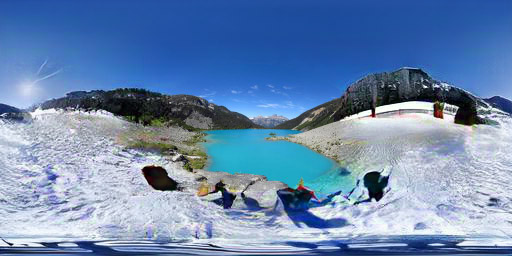} & 
\includegraphics[width=\mywidth]{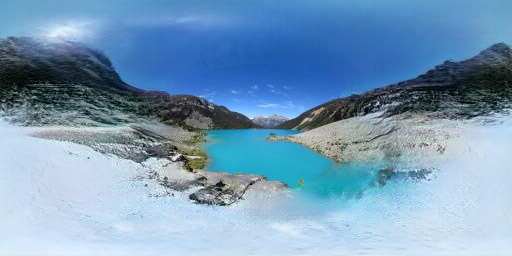} &
\includegraphics[width=\mywidth]{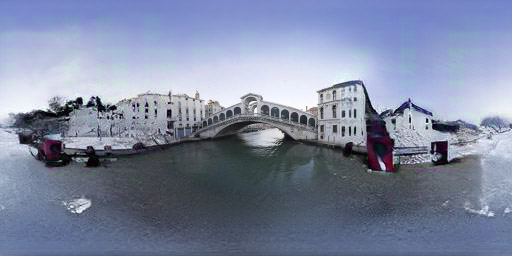} & 
\includegraphics[width=\mywidth]{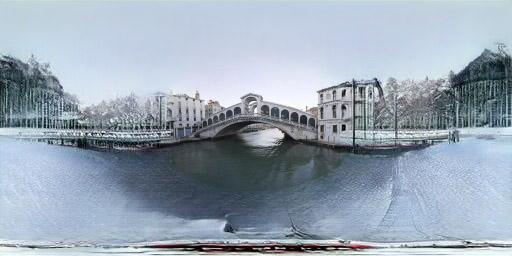} \\ 
\rotatebox{90}{\tiny $\mapsto \text{``church''}$} &
\includegraphics[width=\mywidth]{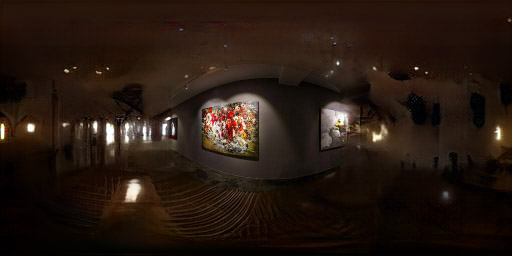} & 
\includegraphics[width=\mywidth]{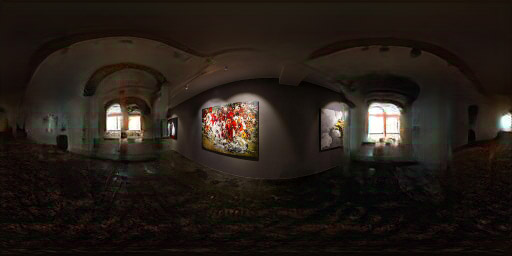} &
\includegraphics[width=\mywidth]{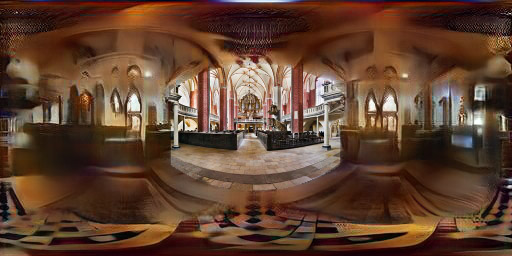} & 
\includegraphics[width=\mywidth]{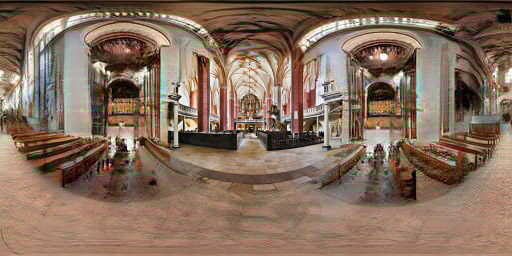} \\ 
% \rotatebox{90}{\tiny $\mapsto \text{``promend''}$} &
% \includegraphics[width=\mywidth]{figs/editing_comparison/scores_promenade/ig_39665_promenade_generated.jpg} & 
% \includegraphics[width=\mywidth]{figs/editing_comparison/scores_promenade/ours_39665_promenade_generated.jpg} &
% \includegraphics[width=\mywidth]{figs/editing_comparison/scores_promenade/ig_41323_promenade_generated.jpg} & 
% \includegraphics[width=\mywidth]{figs/editing_comparison/scores_promenade/ours_41323_promenade_generated.jpg} \\ 
\end{tabular}
\caption[]{Comparison to InterFaceGAN~\cite{shen2020interpreting}, with ``sky'', ``snowfield'', and ``church'' target labels, from top to bottom respectively. While InterFaceGAN yields reasonable results, we observe it more often generates visual artifacts and results are less semantically coherent with the input image.}
\label{fig:comparison-ig}
\end{figure}

We compare our editing results with those obtained with InterFaceGAN~\cite{shen2020interpreting}, a StyleGAN-based face editing method which first classifies generated images using a pretrained attribute classifier, and finds semantic boundaries by fitting a linear SVM on the latent codes of images of two different classes (\eg, smile and no-smile). As mentioned in \cref{sec:intro}, there exists no attribute classifier for panoramas---nor is there a labelled dataset to train one---so the method cannot be applied directly. Nevertheless, we adapt it by extracting the ``backwards-looking'' image $\mathbf{x}_\mathrm{back}$ (\cf \cref{sec:editing-experiments}) from a panorama generated by \thename on a random input image and employ the guide network $\mathcal{G}$ to obtain the label. We then fit the linear SVM in the $\mathbf{w}$ space (\cref{fig:method}), before the affine transformation $A$. \Cref{fig:comparison-ig} shows examples of results obtained with \cite{shen2020interpreting} (left) and ours (right). Results with \cite{shen2020interpreting} are plausible, but note how the consistency between the input crop and the extrapolated environment is weaker than with our method, resulting in strong visual artifacts (input image clearly visible in bottom-right of \cref{fig:comparison-ig}) or large semantic changes (pavement turns to water in top-right of \cref{fig:comparison-ig}). % \momo{Please note that a quantitative analysis using standard image quality metric such as FID is not feasible here because the ground truth panoramas corresponding to the target labels do not exist.}

%% file: fig_qualresults_rot1_.tex
\begin{figure*}
    \centering
    \renewcommand{\tabcolsep}{1pt}
    \newcommand{\mywidth}{0.2\linewidth}
    \resizebox{\linewidth}{!}{% take the[] entire width, but still find a good per-image width otherwise text gets compressed
    \begin{tabular}{ccccccc}
    Input 
    & Ground truth
    & pix2pixHD~\cite{wang2018high} 
    % & Symmetry~\cite{Hara_Mukuta_Harada_2021}-R 
    & Symmetry~\cite{Hara_Mukuta_Harada_2021}-G
    & CoModGAN~\cite{zhao2021comodgan}
    & \thename
    & Guided 
    \\
    % \rotatebox{90}{\parbox{1.3cm}{\centering $h_\theta = 40^\circ$ \\ $\beta=0^\circ$}} 
    % & \includegraphics[width=\mywidth]{figs/quant/ours/40d/partial/19659_0_full_partial.jpg}
    % & \includegraphics[width=\mywidth]{figs/quant/ours/40d/original/19659_0_full_original.jpg}
    % & \includegraphics[width=\mywidth]{figs/quant/pix2pixHD/40d/generated/19659_0_full_0_synthesized_image.jpg}
    % % & \includegraphics[width=\mywidth]{figs/quant/symmetry/40d/rec/19659_0_partial_out_0.jpg} 
    % & \includegraphics[width=\mywidth]{figs/quant/symmetry/40d/gen/19659_0_partial_out_0.jpg} 
    % & \includegraphics[width=\mywidth]{figs/quant/vanilla_comodgan/40d/19659_original.jpg}
    % & \includegraphics[width=\mywidth]{figs/quant/CoModGAN++/40d/generated/19659_0_full_generated.jpg}
    % & \includegraphics[width=\mywidth]{figs/quant/ours/40d/generated/19659_generated.jpg}
    % \\
    % \parbox[b]{1.5cm}{\centering $h_\theta = 40^\circ$ \\ $\beta=0^\circ$}
    % \begin{tabular}{c}$h_\theta = 40^\circ$ \\ $\beta=0^\circ$\end{tabular} &
    \includegraphics[width=\mywidth]{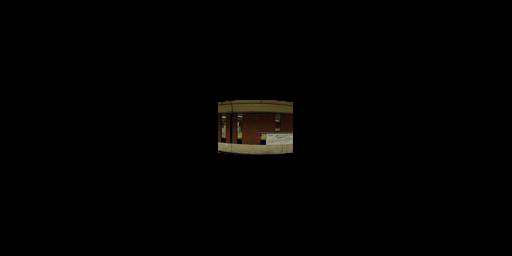}
    & \includegraphics[width=\mywidth]{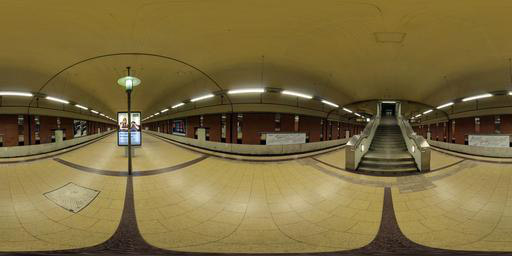}
    & \includegraphics[width=\mywidth]{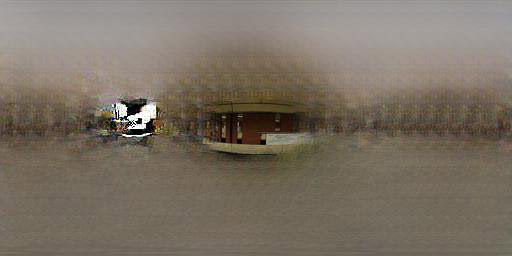}
    % & \includegraphics[width=\mywidth]{figs/quant/symmetry/40d/rec/21967_0_partial_out_0.jpg} 
    & \includegraphics[width=\mywidth]{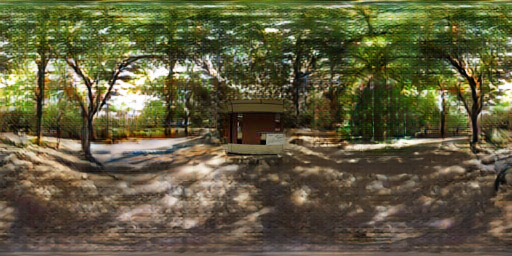} 
    & \includegraphics[width=\mywidth]{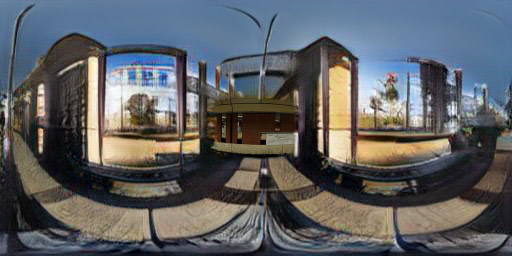}
    & \includegraphics[width=\mywidth]{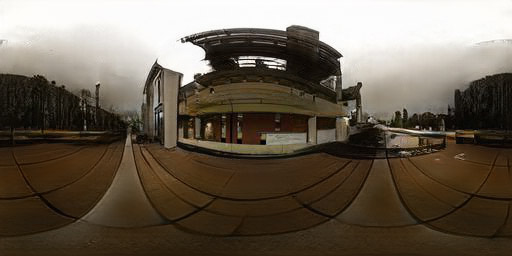}
    & \includegraphics[width=\mywidth]{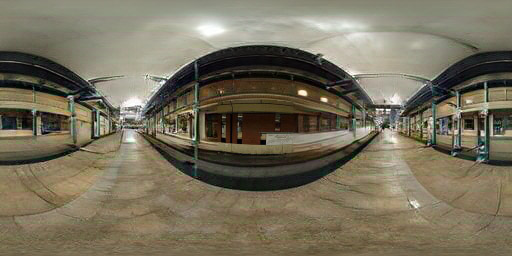}
    \\
    % \parbox[b]{1.5cm}{\centering $h_\theta = 60^\circ$ \\ $\beta=0^\circ$} &
    \includegraphics[width=\mywidth]{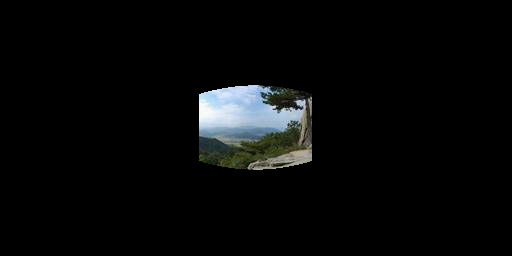}
    & \includegraphics[width=\mywidth]{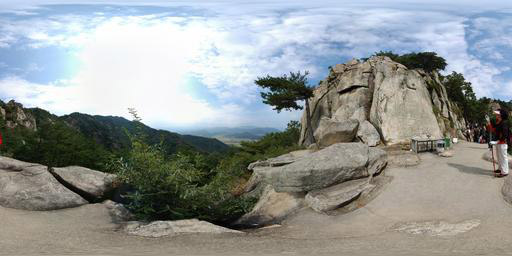}
    & \includegraphics[width=\mywidth]{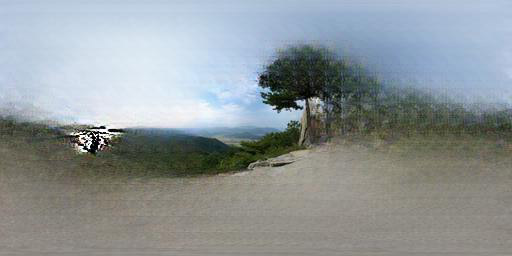}
    % & \includegraphics[width=\mywidth]{figs/quant/symmetry/60d/rec/180086_0_partial_out_0.jpg} 
    & \includegraphics[width=\mywidth]{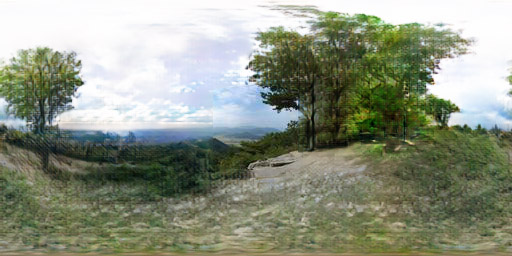} 
    & \includegraphics[width=\mywidth]{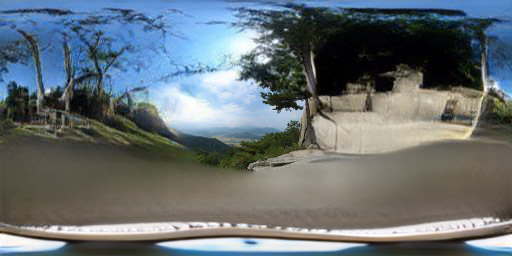}
    & \includegraphics[width=\mywidth]{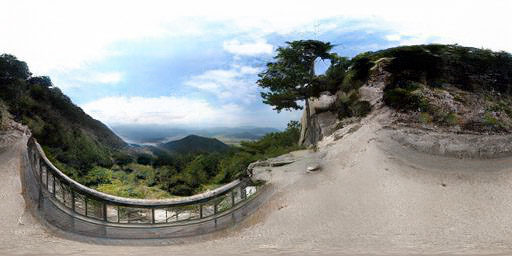}
    & \includegraphics[width=\mywidth]{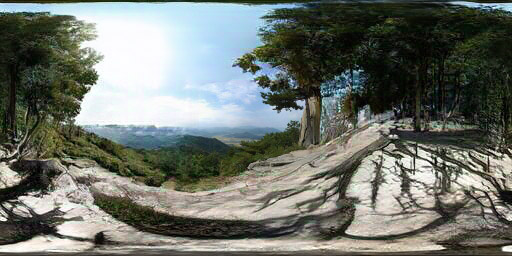}
    \\
    % \rotatebox{90}{\parbox{1.3cm}{\centering $h_\theta = 60^\circ$ \\ $\beta=0^\circ$}} 
    % & \includegraphics[width=\mywidth]{figs/quant/ours/60d/partial/8582_0_full_partial.jpg}
    % & \includegraphics[width=\mywidth]{figs/quant/ours/60d/original/8582_0_full_original.jpg}
    % & \includegraphics[width=\mywidth]{figs/quant/pix2pixHD/60d/generated/8582_0_full_0_synthesized_image.jpg}
    % % & \includegraphics[width=\mywidth]{figs/quant/symmetry/60d/rec/8582_0_partial_out_0.jpg} 
    % & \includegraphics[width=\mywidth]{figs/quant/symmetry/60d/gen/8582_0_partial_out_0.jpg} 
    % & \includegraphics[width=\mywidth]{figs/quant/vanilla_comodgan/60d/8582_original.jpg} 
    % & \includegraphics[width=\mywidth]{figs/quant/CoModGAN++/60d/generated/8582_0_full_generated.jpg}
    % & \includegraphics[width=\mywidth]{figs/quant/ours/60d/generated/8582_generated.jpg}
    % \\
    % \parbox[b]{1.5cm}{\centering $h_\theta = 90^\circ$ \\ $\beta=0^\circ$} &
    \includegraphics[width=\mywidth]{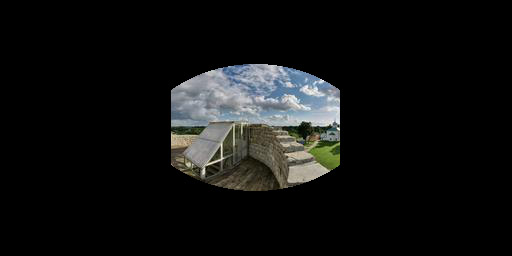}
    & \includegraphics[width=\mywidth]{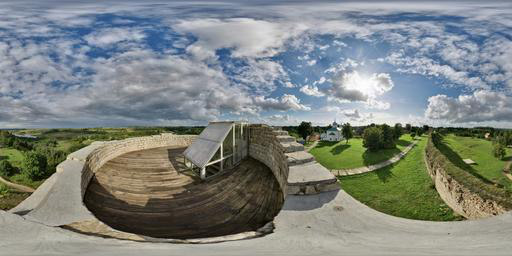}
    & \includegraphics[width=\mywidth]{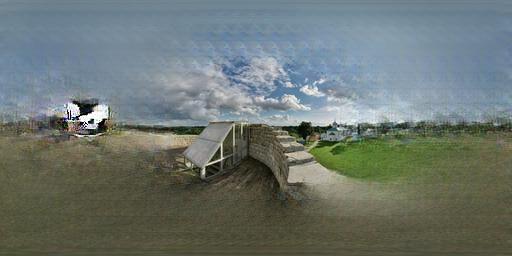}
    % & \includegraphics[width=\mywidth]{figs/quant/symmetry/90d/rec/8008_0_partial_out_0.jpg}
    & \includegraphics[width=\mywidth]{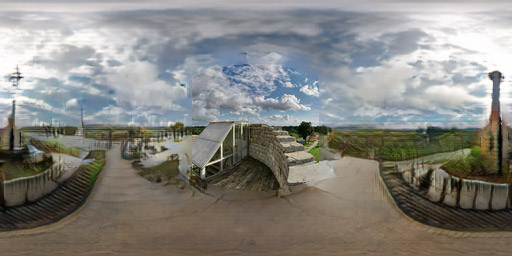}
    & \includegraphics[width=\mywidth]{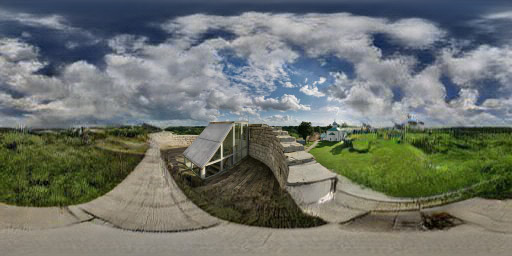}
    & \includegraphics[width=\mywidth]{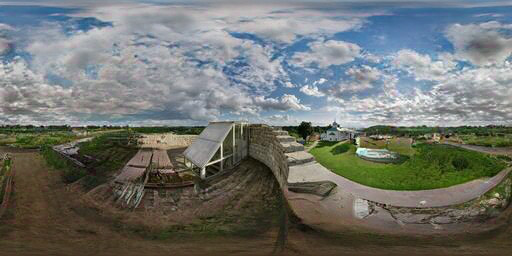}
    & \includegraphics[width=\mywidth]{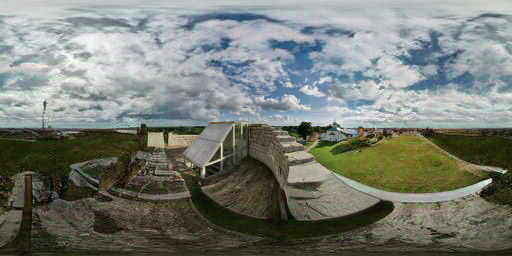}
    \\
    \includegraphics[width=\mywidth]{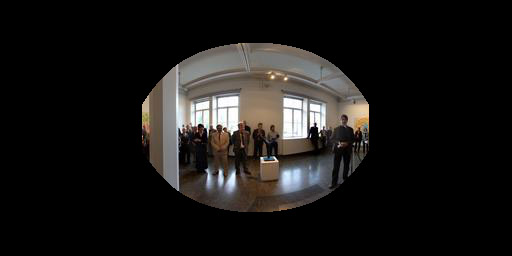}
    & \includegraphics[width=\mywidth]{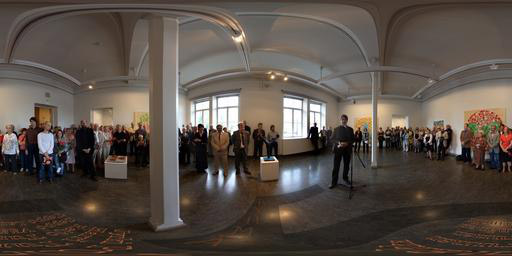}
    & \includegraphics[width=\mywidth]{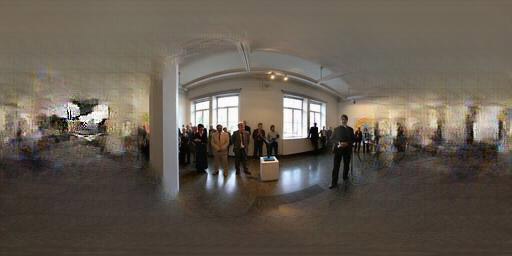}
    % & \includegraphics[width=\mywidth]{figs/quant/symmetry/120d/rec/41272_0_partial_out_0.jpg}
    & \includegraphics[width=\mywidth]{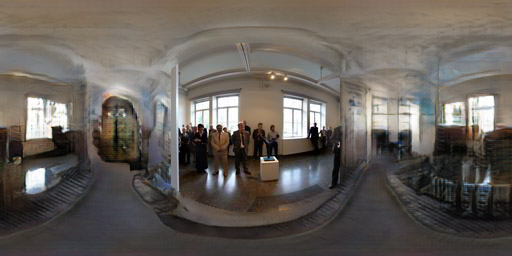}
    & \includegraphics[width=\mywidth]{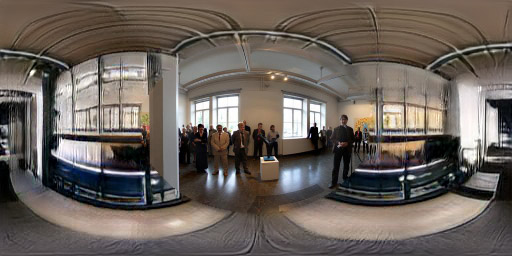}
    & \includegraphics[width=\mywidth]{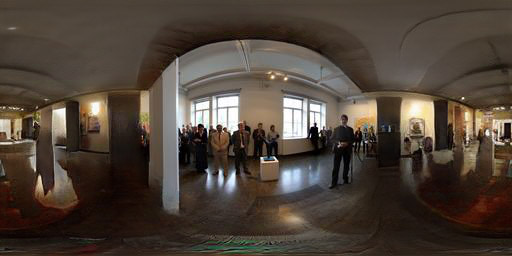}
    & \includegraphics[width=\mywidth]{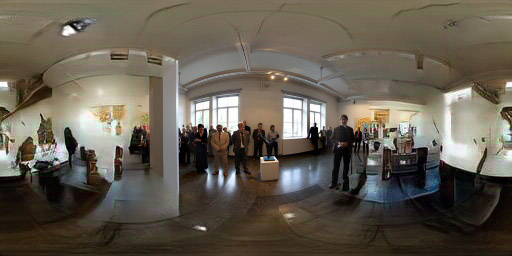}
    \\
    % \parbox[b]{1.5cm}{\centering $h_\theta = 44^\circ$ \\ $\beta=-8^\circ$} &
    \includegraphics[width=\mywidth]{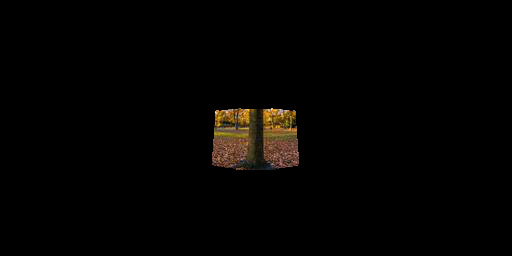}
    & \includegraphics[width=\mywidth]{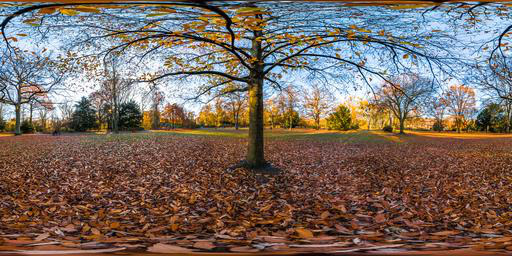}
    & \includegraphics[width=\mywidth]{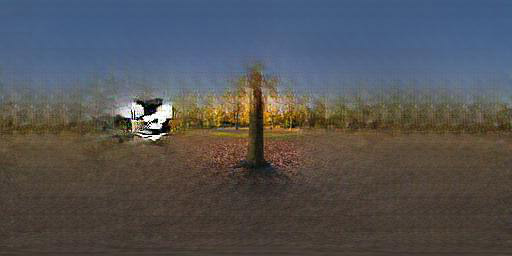}
    % & \includegraphics[width=\mywidth]{figs/quant/symmetry/random/rec/782658_0_partial_out_0.jpg}
    & \includegraphics[width=\mywidth]{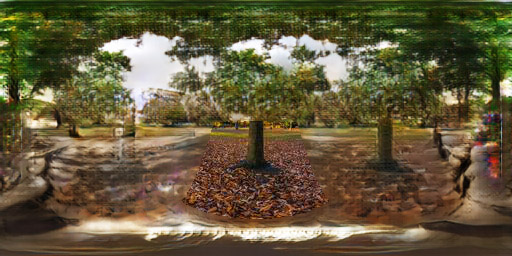}
    & \includegraphics[width=\mywidth]{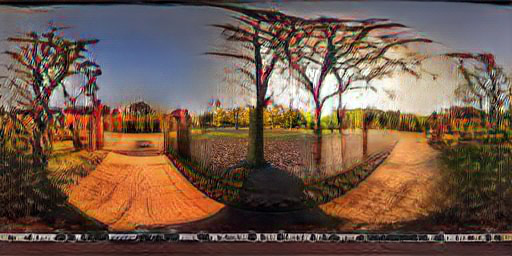}
    & \includegraphics[width=\mywidth]{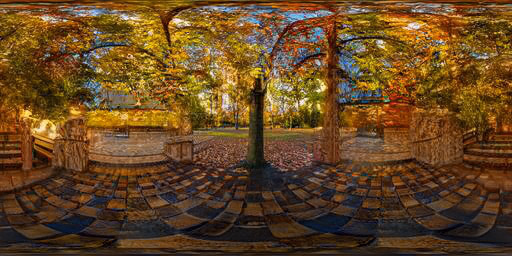}
    & \includegraphics[width=\mywidth]{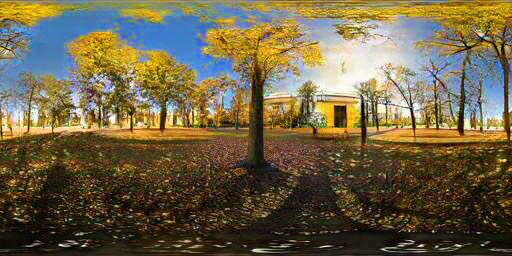}
    \\
    % \parbox[b]{1.7cm}{\centering $h_\theta = 110^\circ$ \\ $\beta=23^\circ$} &
    \includegraphics[width=\mywidth]{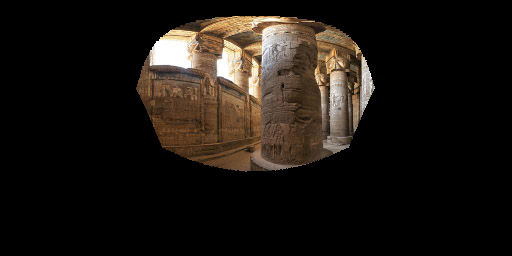}
    & \includegraphics[width=\mywidth]{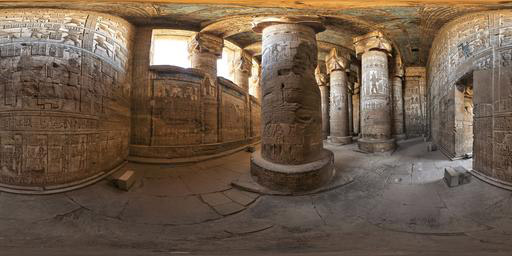}
    & \includegraphics[width=\mywidth]{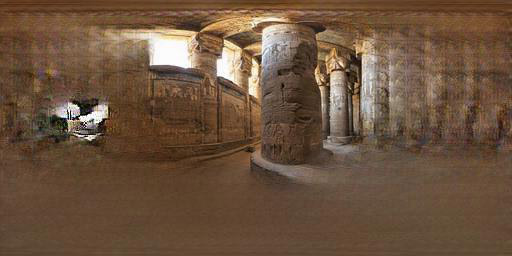}
    % & \includegraphics[width=\mywidth]{figs/quant/symmetry/random/rec/744364_0_partial_out_0.jpg}
    & \includegraphics[width=\mywidth]{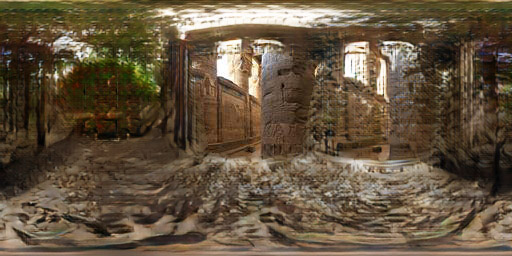}
    & \includegraphics[width=\mywidth]{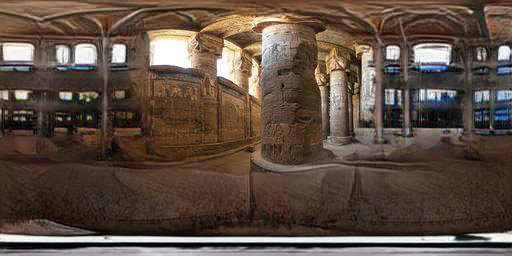}
    & \includegraphics[width=\mywidth]{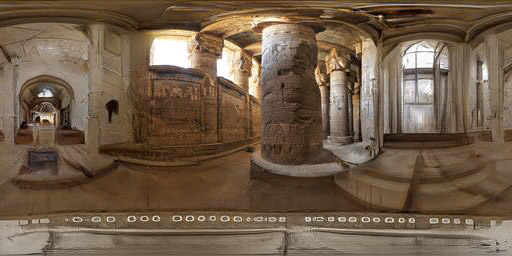}
    & \includegraphics[width=\mywidth]{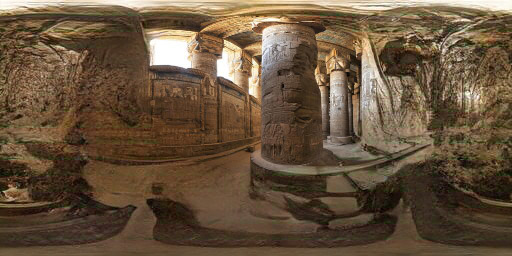}
    \\
    % \rotatebox{90}{\parbox{1.3cm}{\centering $h_\theta = 77^\circ$ \\ $\beta=-30^\circ$}}
    % & \includegraphics[width=\mywidth]{figs/quant/ours/random/partial/89000_0_full_77_30_partial.jpg}
    % & \includegraphics[width=\mywidth]{figs/quant/ours/40d/original/89000_0_full_original.jpg}
    % & \includegraphics[width=\mywidth]{figs/quant/pix2pixHD/random/generated/89000_0_full_77_30_0_synthesized_image.jpg}
    % % & \includegraphics[width=\mywidth]{figs/quant/symmetry/random/rec/89000_0_partial_out_0.jpg}
    % & \includegraphics[width=\mywidth]{figs/quant/symmetry/random/gen/89000_0_partial_out_0.jpg}
    % & \includegraphics[width=\mywidth]{figs/quant/vanilla_comodgan/random/89000_0_full_77_30.0_0_full.jpg}
    % & \includegraphics[width=\mywidth]{figs/quant/CoModGAN++/random/generated/89000_0_full_77_30_generated.jpg}
    % & \includegraphics[width=\mywidth]{figs/quant/ours/random/generated/89000_original_generated.jpg}
    % \parbox[b]{1.5cm}{\centering $h_\theta = 85^\circ$ \\ $\beta=-28^\circ$} &
    \includegraphics[width=\mywidth]{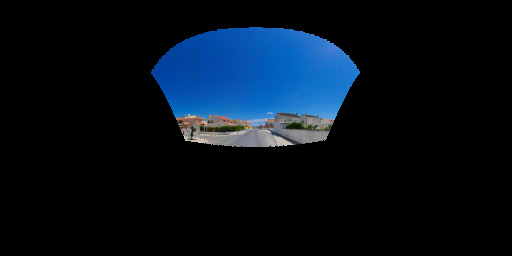}
    & \includegraphics[width=\mywidth]{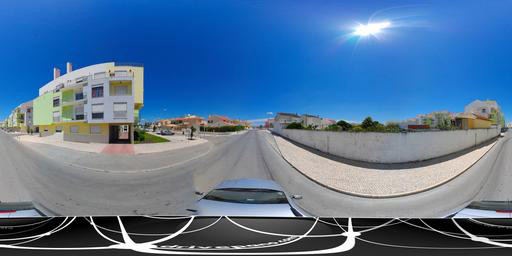}
    & \includegraphics[width=\mywidth]{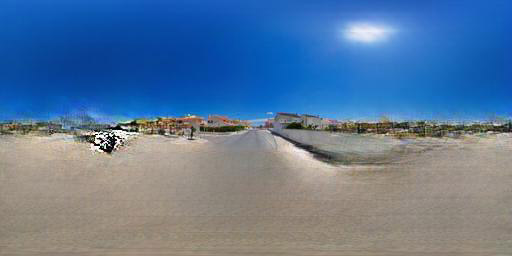}
    % & \includegraphics[width=\mywidth]{figs/quant/symmetry/random/rec/89000_0_partial_out_0.jpg}
    & \includegraphics[width=\mywidth]{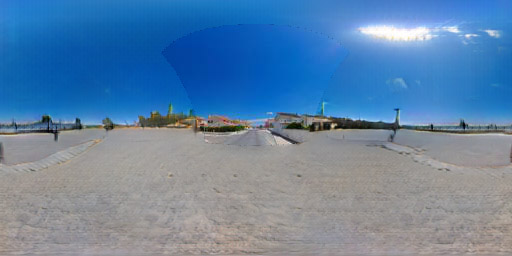}
    & \includegraphics[width=\mywidth]{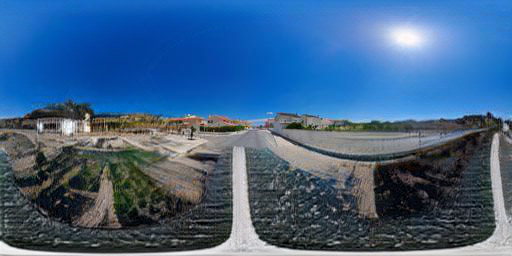}
    & \includegraphics[width=\mywidth]{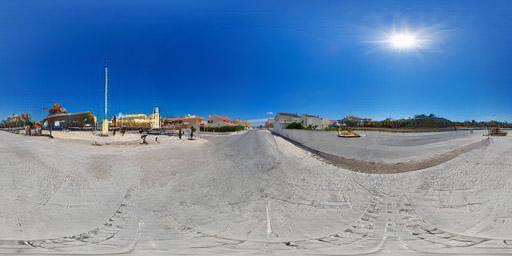}
    & \includegraphics[width=\mywidth]{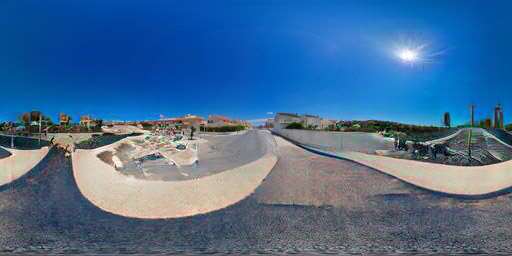}
    \\
    % \parbox[b]{1.5cm}{\centering $h_\theta = 92^\circ$ \\ $\beta=10^\circ$} &
    \includegraphics[width=\mywidth]{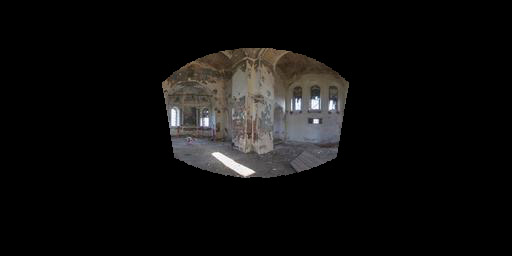}
    & \includegraphics[width=\mywidth]{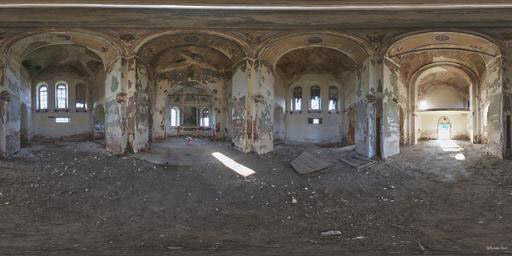}
    & \includegraphics[width=\mywidth]{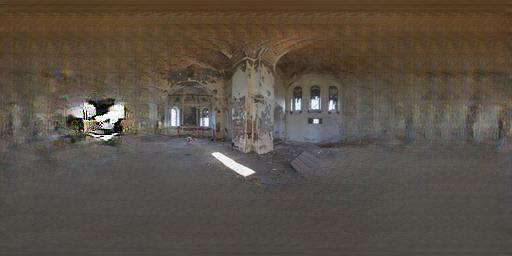}
    % & \includegraphics[width=\mywidth]{figs/quant/symmetry/random/rec/736147_0_partial_out_0.jpg}
    & \includegraphics[width=\mywidth]{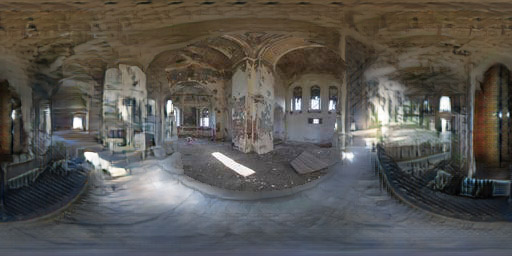}
    & \includegraphics[width=\mywidth]{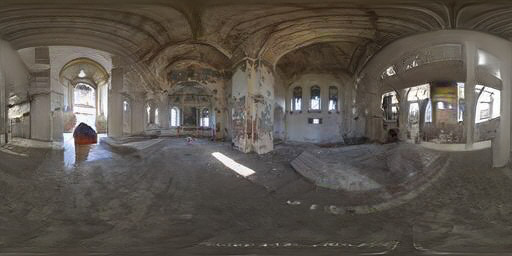}
    & \includegraphics[width=\mywidth]{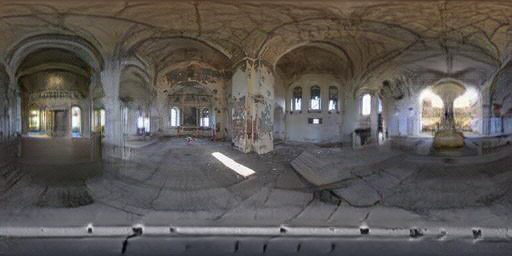}
    & \includegraphics[width=\mywidth]{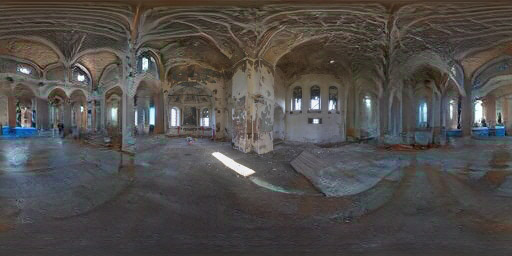}
    \\
    % \parbox[b]{1.5cm}{\centering $h_\theta = 94^\circ$ \\ $\beta=-17^\circ$} &
    \includegraphics[width=\mywidth]{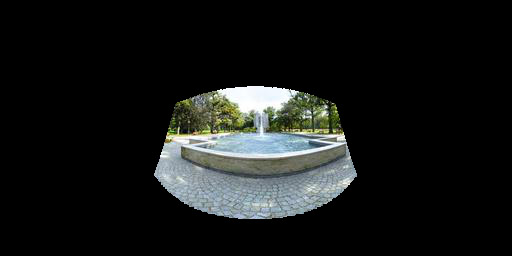}
    & \includegraphics[width=\mywidth]{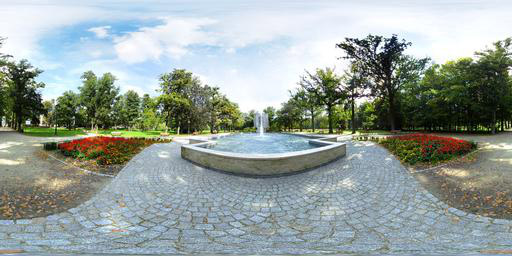} 
    & \includegraphics[width=\mywidth]{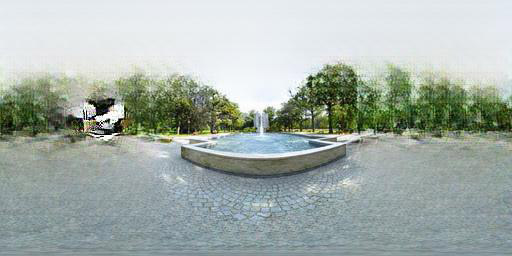}
    % & \includegraphics[width=\mywidth]{figs/quant/symmetry/random/rec/429560_0_partial_out_0.jpg}
    & \includegraphics[width=\mywidth]{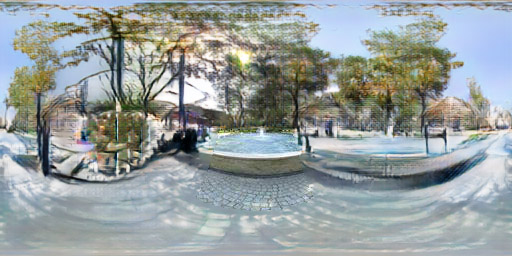}
    & \includegraphics[width=\mywidth]{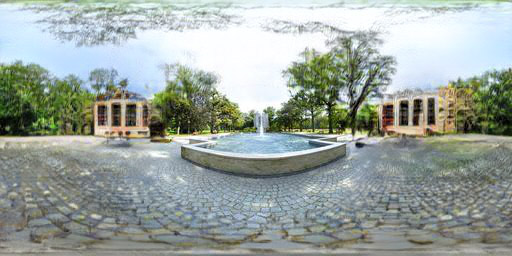}
    & \includegraphics[width=\mywidth]{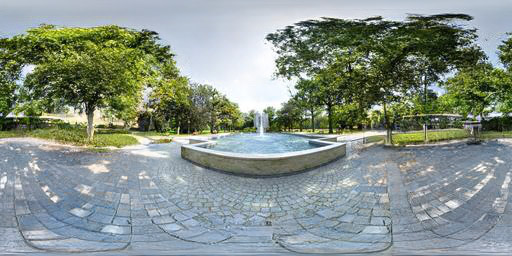}
    & \includegraphics[width=\mywidth]{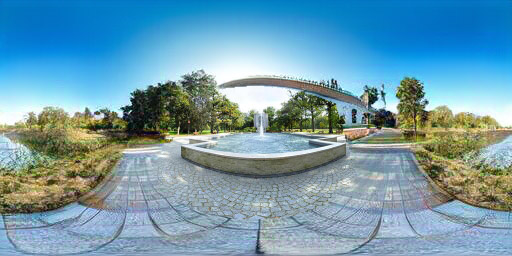}
    \\
    % \parbox[b]{1.5cm}{\centering $h_\theta = 97^\circ$ \\ $\beta=-4^\circ$} &
    \includegraphics[width=\mywidth]{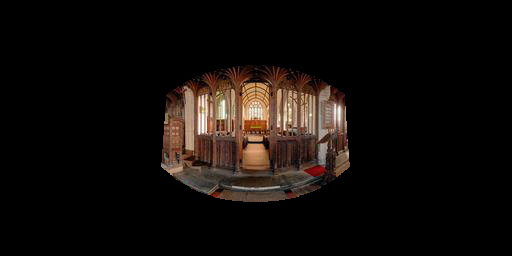}
    & \includegraphics[width=\mywidth]{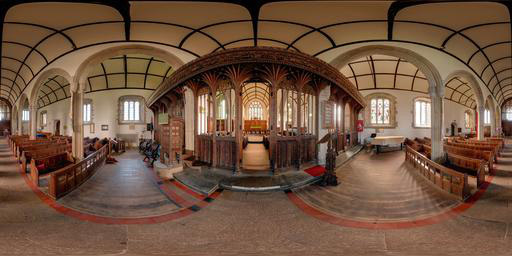}
    & \includegraphics[width=\mywidth]{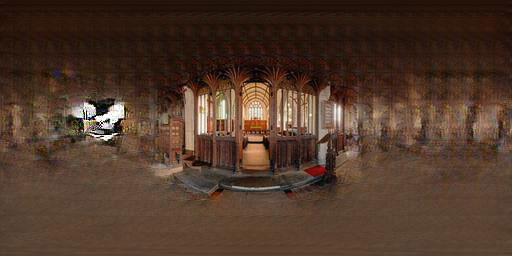}
    % & \includegraphics[width=\mywidth]{figs/quant/symmetry/random/rec/405479_0_partial_out_0.jpg}
    & \includegraphics[width=\mywidth]{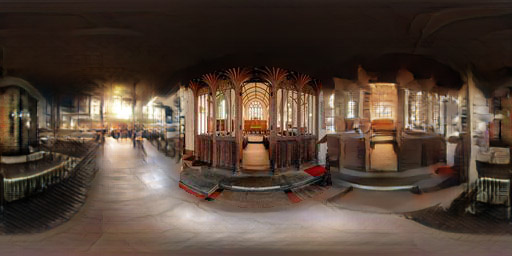}
    & \includegraphics[width=\mywidth]{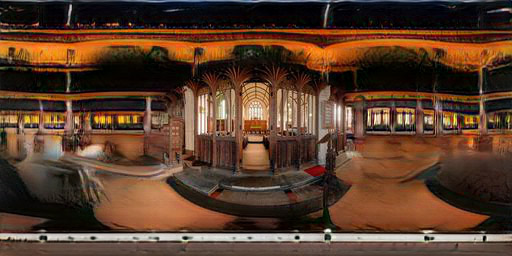}
    & \includegraphics[width=\mywidth]{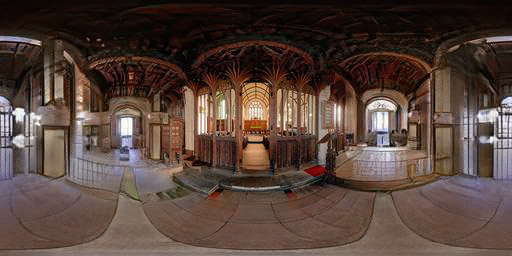}
    & \includegraphics[width=\mywidth]{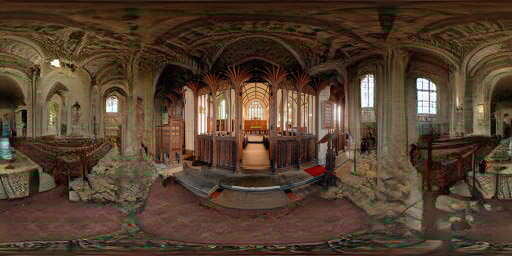}
    \\
    \end{tabular}}
    \caption{Qualitative field of view extrapolation results. Previous works~\cite{wang2018high,Hara_Mukuta_Harada_2021,zhao2021comodgan} present obvious visual artifacts, including mode collapse~\cite{wang2018high}, blurriness~\cite{wang2018high,Hara_Mukuta_Harada_2021}, and semantic mismatch ~\cite{wang2018high,Hara_Mukuta_Harada_2021}. In contrast, our \thename enables high-quality results (penultimate column), which are preserved by our novel guided co-modulation framework (last column). The first four rows contain examples from the ``fixed-fov'' subset of our test set, while the last six are taken from the ``mixed'' subset.}
    \label{fig:qual-comparison}
\end{figure*}

%% file: fig_comodgan_vanilla_vs_360.tex
\begin{figure} [t]
    \centering
    \scriptsize
    \setlength{\tabcolsep}{1pt}
    \newcommand{\mywidth}{0.19\linewidth}
    \begin{tabular}{ccccc}
    & 
    \multicolumn{2}{c}{CoModGAN~\cite{zhao2021comodgan}} & 
    \multicolumn{2}{c}{\thename (ours)} \\
    Input & Original & Rotated & Original & Rotated \\
    \includegraphics[width=\mywidth]{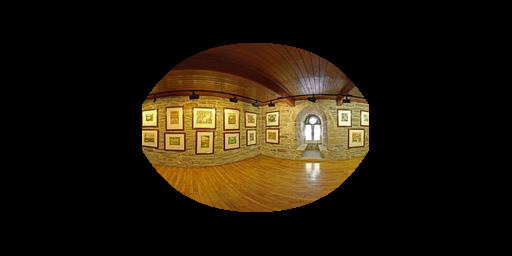}
    & \includegraphics[width=\mywidth]{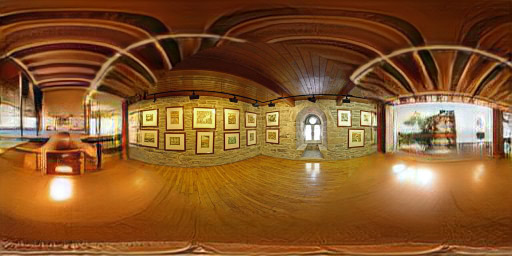}
    & \includegraphics[width=\mywidth]{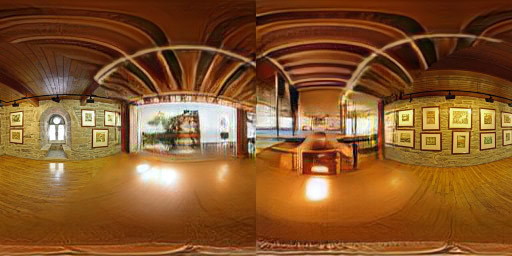}
    & \includegraphics[width=\mywidth]{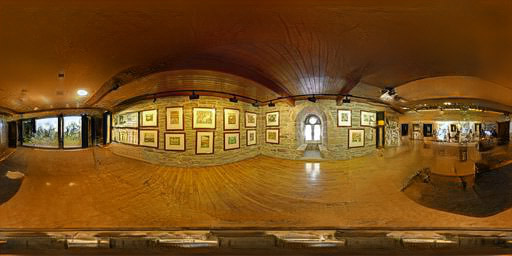}
    & \includegraphics[width=\mywidth]{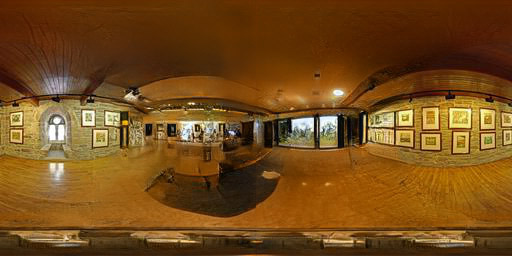}
    \\
    \includegraphics[width=\mywidth]{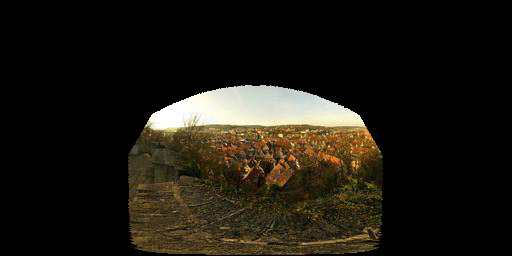}
    & \includegraphics[width=\mywidth]{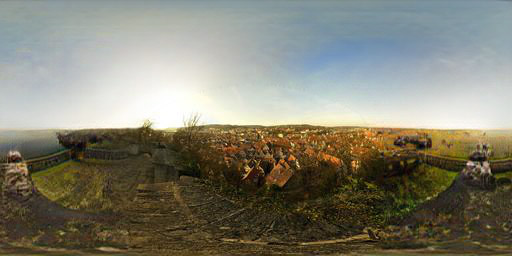}
    & \includegraphics[width=\mywidth]{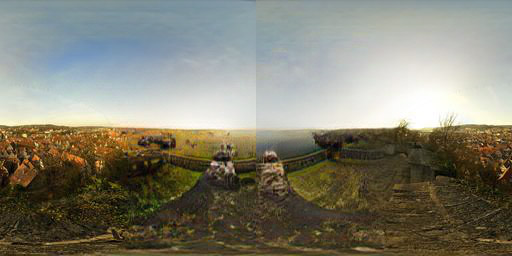}
    & \includegraphics[width=\mywidth]{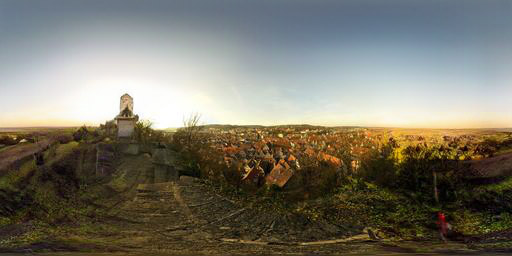}
    & \includegraphics[width=\mywidth]{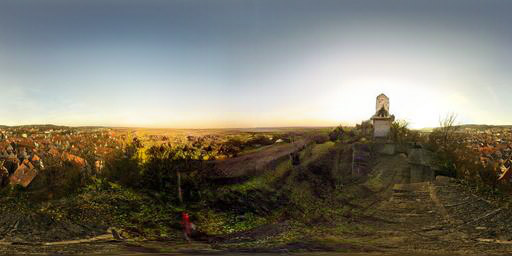}
    \\
    \end{tabular}
    \setlength{\tabcolsep}{5pt}
    \begin{tabular}{lccccc}
    \toprule
    Method & $40^\circ$ & $60^\circ$ & $90^\circ$ & $120^\circ$ & Mixed\\
    \midrule
    CoModGAN \cite{zhao2021comodgan} & 81.10 & 70.74 & 50.72 & 40.63 & 51.54 \\
    \thename (ours) & \cellcolor{best}{38.58} & \cellcolor{best}{36.37} & \cellcolor{best}{34.72} & \cellcolor{best}{33.34} & \cellcolor{best}{34.48} \\
    \bottomrule
    \end{tabular}
    \caption{\emph{Top}: Rotating panoramas by 180\degree{} azimuth reveals that CoModGAN (left) generates a vertical seam, which is removed with our technique (right). \emph{Bottom}: FID computed after rotating the generated panoramas by 180\degree{} azimuth.}
    \label{fig:comodgan-vs-ours}
\end{figure}

%% file: fig_editingresult.tex
\begin{figure*} [t!]
    \centering
    \renewcommand{\tabcolsep}{1pt}
    \newcommand{\mywidth}{0.2\linewidth}
    \resizebox{\linewidth}{!}{% take the[] entire width, but still find a good per-image width otherwise text gets compressed
    \begin{tabular}{cccccc}
    Outdoor image 
    & Image-conditioned
    & $\mapsto \text{sky}$
    & $\mapsto \text{snow field}$
    & $\mapsto \text{promenade}$
    & $\mapsto \text{lawn}$
    \\
    \includegraphics[width=\mywidth]{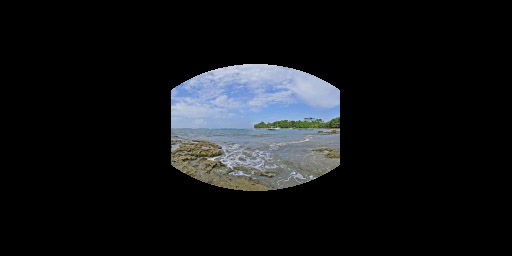}
    & \includegraphics[width=\mywidth]{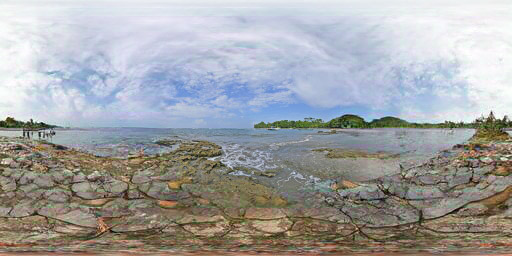}
    & \includegraphics[width=\mywidth]{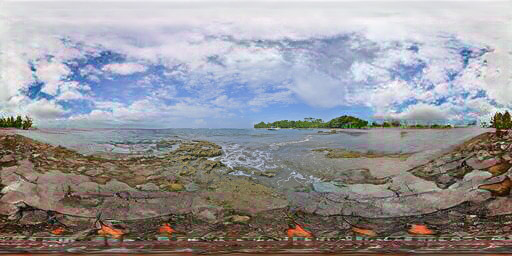}
    & \includegraphics[width=\mywidth]{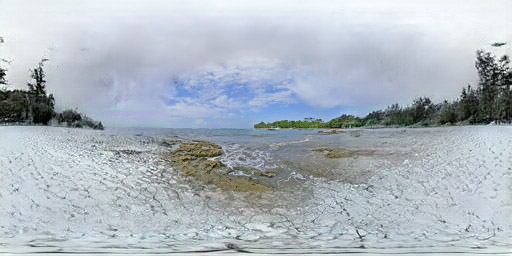}
    & \includegraphics[width=\mywidth]{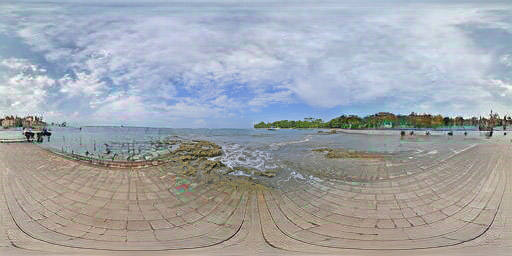}
    & \includegraphics[width=\mywidth]{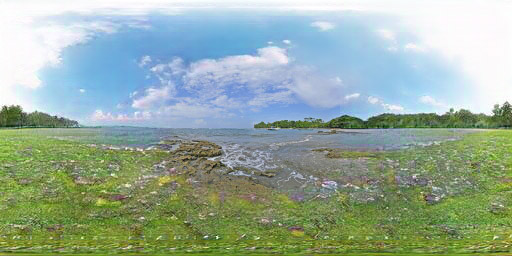}
    \\
    % & \includegraphics[width=\mywidth]{figs/editing/qual/630110_partial.jpg}
    % & \includegraphics[width=\mywidth]{figs/editing/qual/630110_src_generated.jpg}
    % & \includegraphics[width=\mywidth]{figs/editing/qual/630110_sky_generated.jpg}
    % & \includegraphics[width=\mywidth]{figs/editing/qual/630110_snowfield_generated.jpg}
    % & \includegraphics[width=\mywidth]{figs/editing/qual/630110_promenade_generated.jpg}
    % & \includegraphics[width=\mywidth]{figs/editing/qual/630110_lawn_generated.jpg}
    % \\
    % & \includegraphics[width=\mywidth]{figs/editing/qual/152837_partial.jpg}
    % & \includegraphics[width=\mywidth]{figs/editing/qual/152837_src_generated.jpg}
    % & \includegraphics[width=\mywidth]{figs/editing/qual/152837_sky_generated.jpg}
    % & \includegraphics[width=\mywidth]{figs/editing/qual/152837_snowfield_generated.jpg}
    % & \includegraphics[width=\mywidth]{figs/editing/qual/152837_promenade_generated.jpg}
    % & \includegraphics[width=\mywidth]{figs/editing/qual/152837_lawn_generated.jpg}
    % \\
    \includegraphics[width=\mywidth]{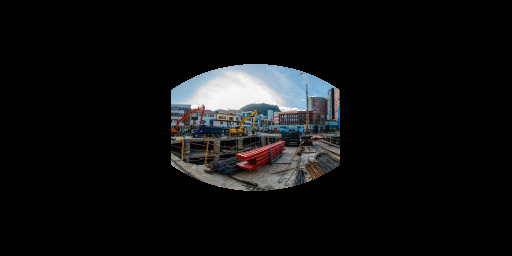}
    & \includegraphics[width=\mywidth]{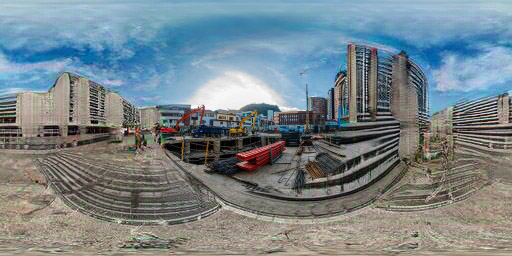}
    & \includegraphics[width=\mywidth]{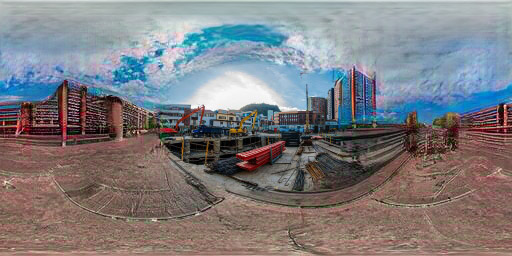}
    & \includegraphics[width=\mywidth]{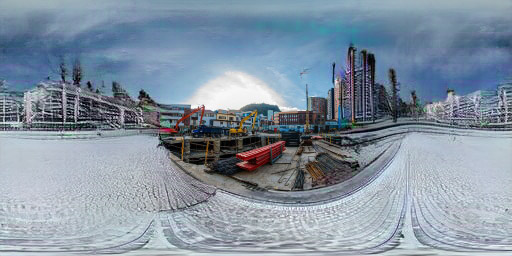}
    & \includegraphics[width=\mywidth]{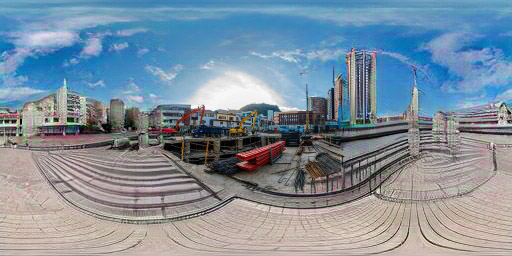}
    & \includegraphics[width=\mywidth]{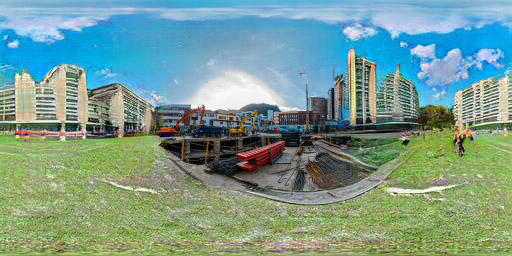}
    \\
    % & \includegraphics[width=\mywidth]{figs/editing/qual/622157_partial.jpg}
    % & \includegraphics[width=\mywidth]{figs/editing/qual/622157_src_generated.jpg}
    % & \includegraphics[width=\mywidth]{figs/editing/qual/622157_sky_generated.jpg}
    % & \includegraphics[width=\mywidth]{figs/editing/qual/622157_snowfield_generated.jpg}
    % & \includegraphics[width=\mywidth]{figs/editing/qual/622157_promenade_generated.jpg}
    % & \includegraphics[width=\mywidth]{figs/editing/qual/622157_lawn_generated.jpg}
    % \\
    % & \includegraphics[width=\mywidth]{figs/editing/qual/600389_partial.jpg}
    % & \includegraphics[width=\mywidth]{figs/editing/qual/600389_src_generated.jpg}
    % & \includegraphics[width=\mywidth]{figs/editing/qual/600389_sky_generated.jpg}
    % & \includegraphics[width=\mywidth]{figs/editing/qual/600389_snowfield_generated.jpg}
    % & \includegraphics[width=\mywidth]{figs/editing/qual/600389_promenade_generated.jpg}
    % & \includegraphics[width=\mywidth]{figs/editing/qual/600389_lawn_generated.jpg}
    % \\
    % & \includegraphics[width=\mywidth]{figs/editing/qual/581272_partial.jpg}
    % & \includegraphics[width=\mywidth]{figs/editing/qual/581272_src_generated.jpg}
    % & \includegraphics[width=\mywidth]{figs/editing/qual/581272_sky_generated.jpg}
    % & \includegraphics[width=\mywidth]{figs/editing/qual/581272_snowfield_generated.jpg}
    % & \includegraphics[width=\mywidth]{figs/editing/qual/581272_promenade_generated.jpg}
    % & \includegraphics[width=\mywidth]{figs/editing/qual/581272_lawn_generated.jpg}
    % \\
    % \midrule
    Indoor image 
    & Image-conditioned
    & $\mapsto \text{entrance hall}$
    & $\mapsto \text{corridor}$
    & $\mapsto \text{artists loft}$
    & $\mapsto \text{throne room}$
    \\
    % \multirow{2}{*}{\adjustbox{valign=m}{\rotatebox{90}{Indoor examples}}} 
    % % & \includegraphics[width=\mywidth]{figs/editing/qual/128847_partial.jpg}
    % & \includegraphics[width=\mywidth]{figs/editing/qual/128847_src_generated.jpg}
    % & \includegraphics[width=\mywidth]{figs/editing/qual/128847_entrance_hall_generated.jpg}
    % & \includegraphics[width=\mywidth]{figs/editing/qual/128847_church_indoor_generated.jpg}
    % & \includegraphics[width=\mywidth]{figs/editing/qual/128847_artists_loft_generated.jpg}
    % & \includegraphics[width=\mywidth]{figs/editing/qual/128847_throne_room_generated.jpg}
    % \\
    \includegraphics[width=\mywidth]{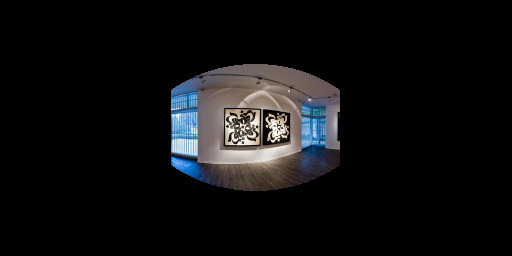}
    & \includegraphics[width=\mywidth]{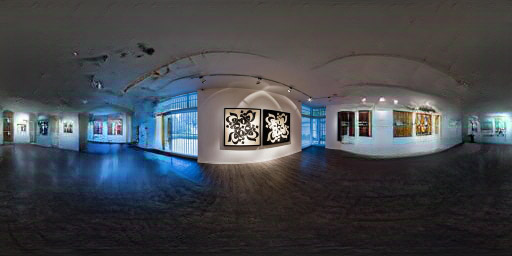}
    & \includegraphics[width=\mywidth]{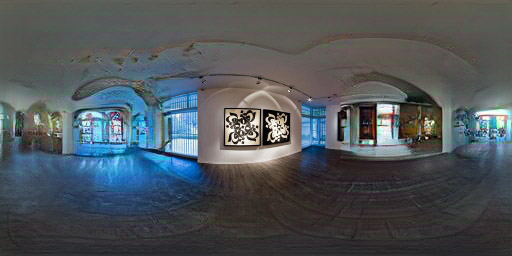}
    & \includegraphics[width=\mywidth]{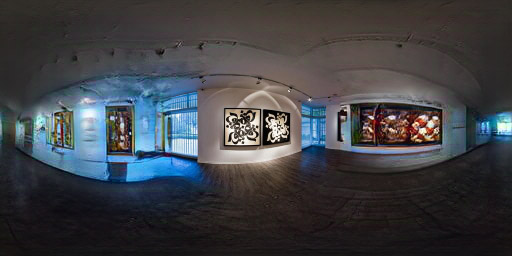}
    & \includegraphics[width=\mywidth]{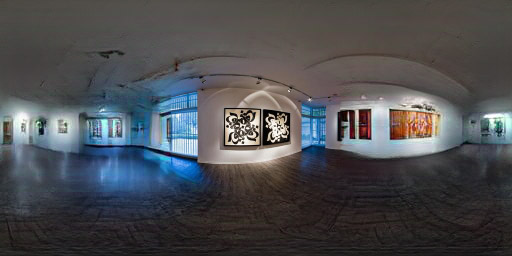}
    & \includegraphics[width=\mywidth]{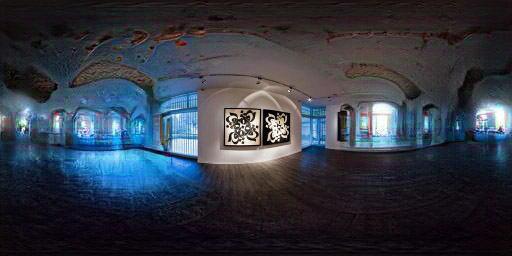}
    \\
    % & \includegraphics[width=\mywidth]{figs/editing/qual/299274_partial.jpg}
    % & \includegraphics[width=\mywidth]{figs/editing/qual/299274_src_generated.jpg}
    % & \includegraphics[width=\mywidth]{figs/editing/qual/299274_entrance_hall_generated.jpg}
    % & \includegraphics[width=\mywidth]{figs/editing/qual/299274_church_indoor_generated.jpg}
    % & \includegraphics[width=\mywidth]{figs/editing/qual/299274_artists_loft_generated.jpg}
    % & \includegraphics[width=\mywidth]{figs/editing/qual/299274_throne_room_generated.jpg}
    % \\
    \includegraphics[width=\mywidth]{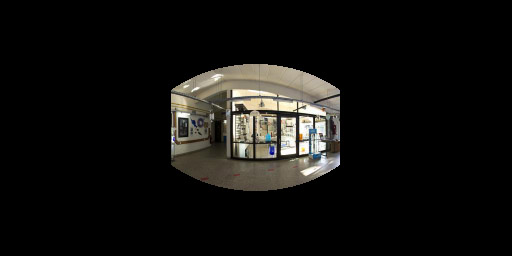}
    & \includegraphics[width=\mywidth]{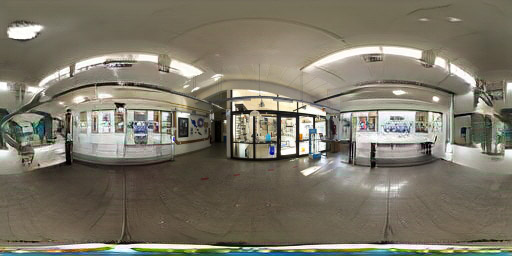}
    & \includegraphics[width=\mywidth]{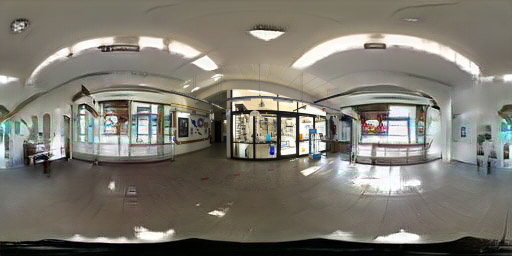}
    & \includegraphics[width=\mywidth]{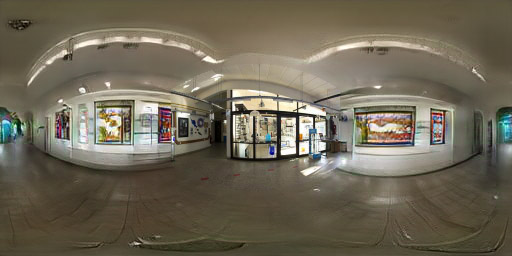}
    & \includegraphics[width=\mywidth]{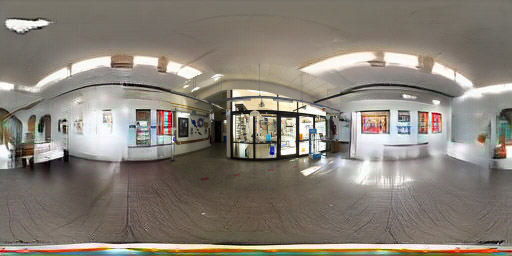}
    & \includegraphics[width=\mywidth]{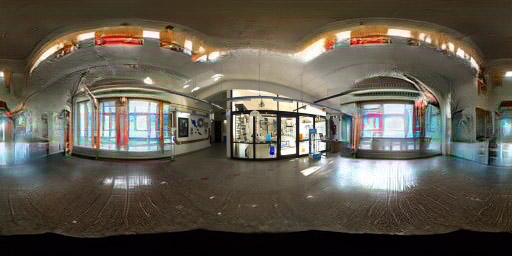}
    \\
    \end{tabular}}
    \caption{Qualitative editing results via our novel guided style co-modulation mechanism. From different input images (``Image''), an extrapolated panorama is first obtained by our network (``Image-conditioned''). A user then selects different target labels, which are used to determine the optimal vector $\mathbf{g}^*$  for style co-modulation. Observe how the network realistically adapts to the (sometimes drastic) change of semantics between the image contents and the desired target label. The random input to the mapper $\mathbf{z}$ is kept constant across rows to ensure the differences are only due to the guidance process.}
    \label{fig:qual-editing}
\end{figure*}

%% file: 6_applications.tex
\section{Application: virtual object compositing}

\input{fig_objectinsertion2}

\paragraph{Approach}
% \label{sec:compositing-approach}
One promising application of FOV extrapolation is that of virtual object compositing~\cite{Debevec1998}, especially when the virtual objects are highly reflective (low roughness). In this case, unseen parts of the environment are reflected onto the virtual object. Here, we leverage recent developments in depth estimation and employ a combination of existing architectures~\cite{xie2021segformer} and datasets~\cite{diode_dataset,Gehrig21ral,madai2016revisiting,vankadari2019unsupervised,tartanair2020iros,silberman2012indoor,MegaDepthLi18,wang2019irs,roberts2021hypersim,garg2019learning} for estimating the depth from the panoramas generated using our technique, and to fit a 3D mesh to the resulting depth map. Details of this approach are given in the supplementary material. This 3D mesh allows for the generation of spatially-varying results, where the reflections off the objects are dependent upon their position in the scene. We use Blender Cycles~\cite{blender} to render virtual objects within the extrapolated 3D mesh and perform the compositing with the original input image. 

\paragraph{Experimental results}
% \label{sec:compositing-results}
Several object insertion results are shown in \cref{fig:teaser,fig:object-insertion}. The panoramas predicted by our method are detailed enough to enable the generation of realistic reflections off of the surface of shiny objects. Also note the spatially-varying reflections that vary as a function of the position of the virtual object in the scene, thereby further increasing the realism. Our method generates sharp, realistic results even for images outside of the training domain (\cref{fig:comparison-obj-insert}), in contrast to \cite{wang2021learning} which produces blurry results for out-of-domain images. Because of its guided editing capabilities, our approach enables users to modify the appearance of the extrapolated FOV, and thus control the appearance of the virtually inserted objects (\cref{fig:teaser,subfig:insertion-edit}). See the supp. material for animations showcasing the results. 

\input{fig_obj_insert_comp}

%% file: fig_objectinsertion2.tex
\begin{figure*}[t!]
    \centering
    \scriptsize
    \renewcommand{\tabcolsep}{1pt}
    \newcommand{\myheight}{1.8cm}

    \subfloat[\label{subfig:insertion-orig}]{
    % \resizebox{0.525\linewidth}{!}{
    % \renewcommand{\tabcolsep}{1pt}
    % \newcommand{\myheight}{1.7cm}
    \begin{tabular}{ccc}
    Input 
    & Panorama 
    & Virt. objects 
    \\
    \includegraphics[height=\myheight]{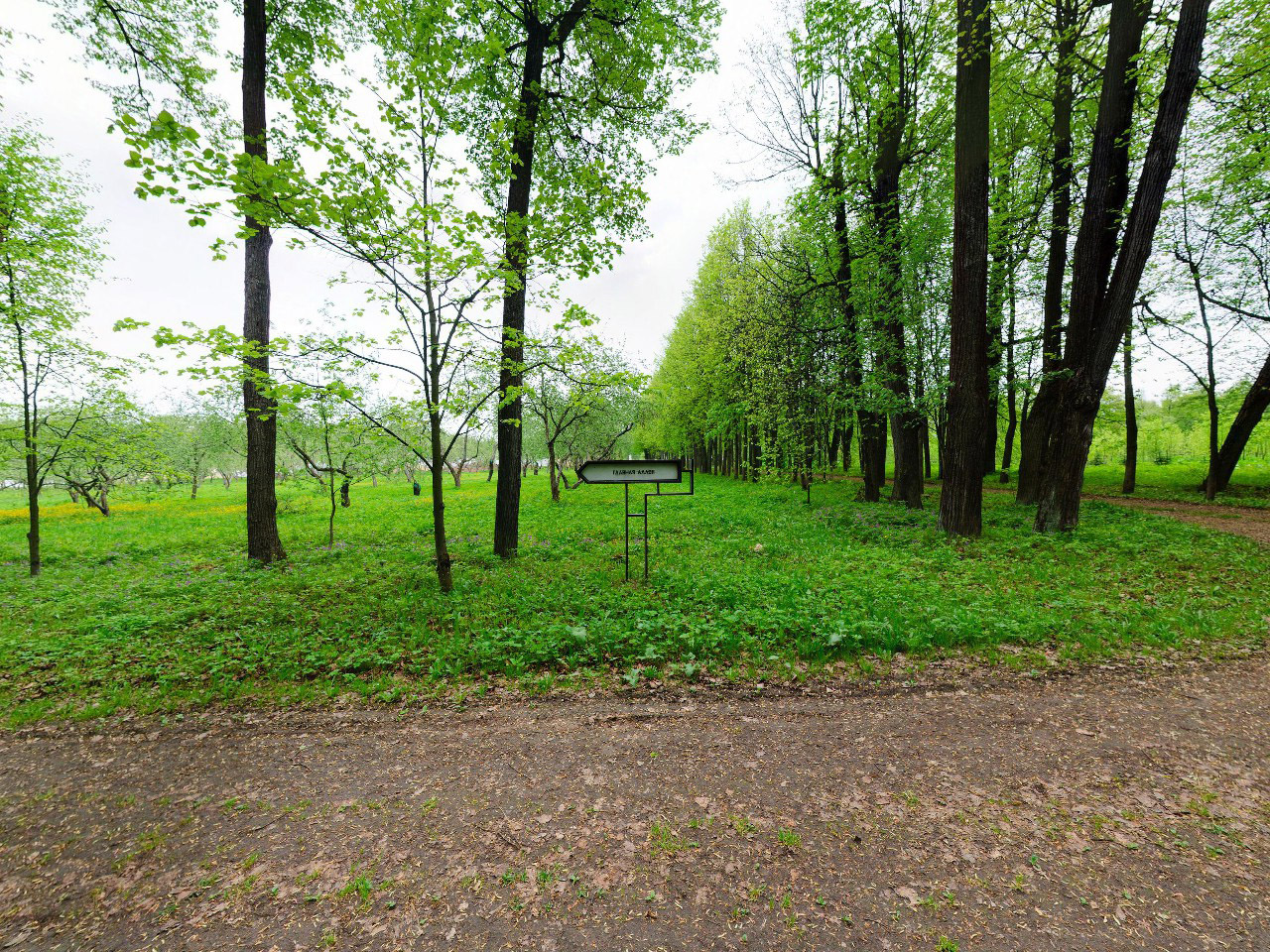}
    & \includegraphics[height=\myheight]{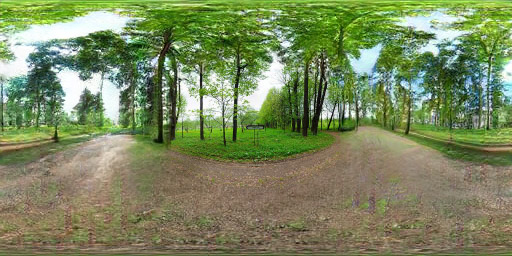}
    & \includegraphics[height=\myheight]{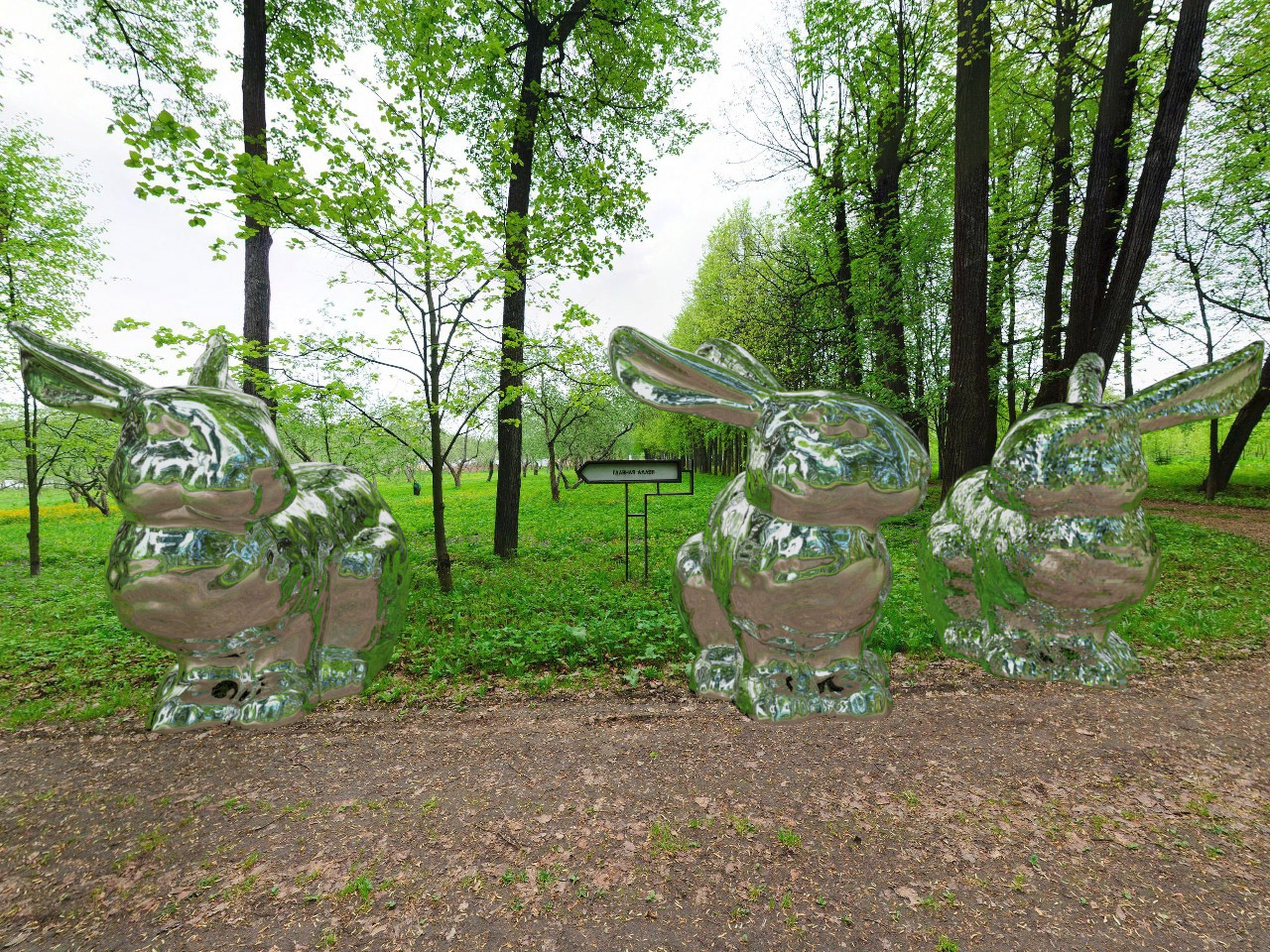} 
    \\
    \includegraphics[height=\myheight]{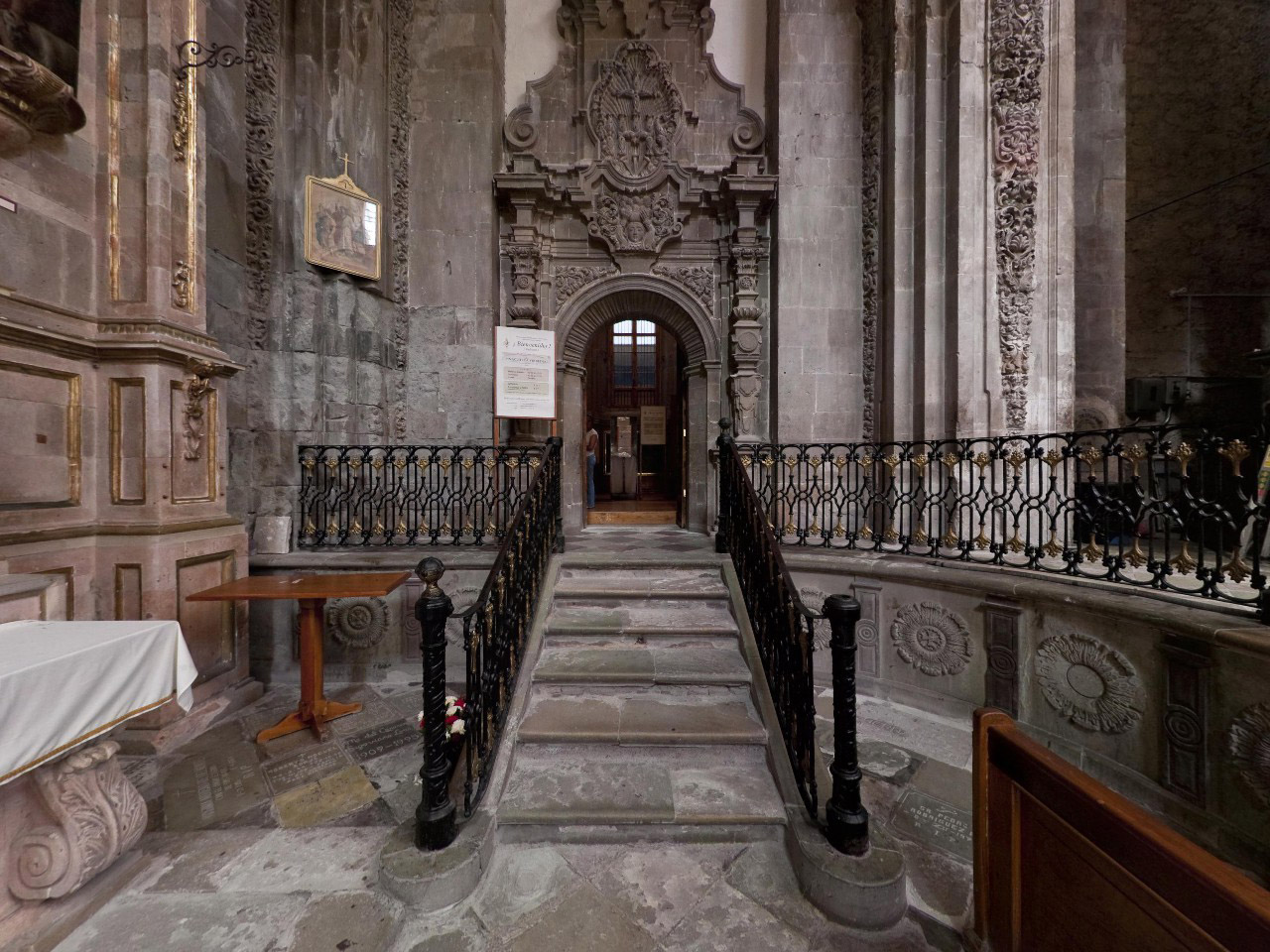}
    & \includegraphics[height=\myheight]{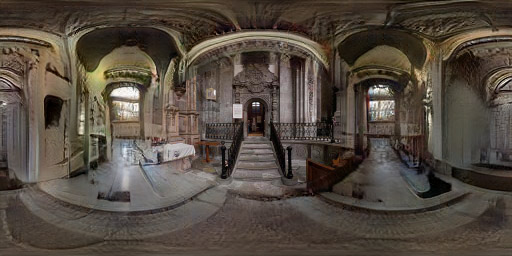}
    & \includegraphics[height=\myheight]{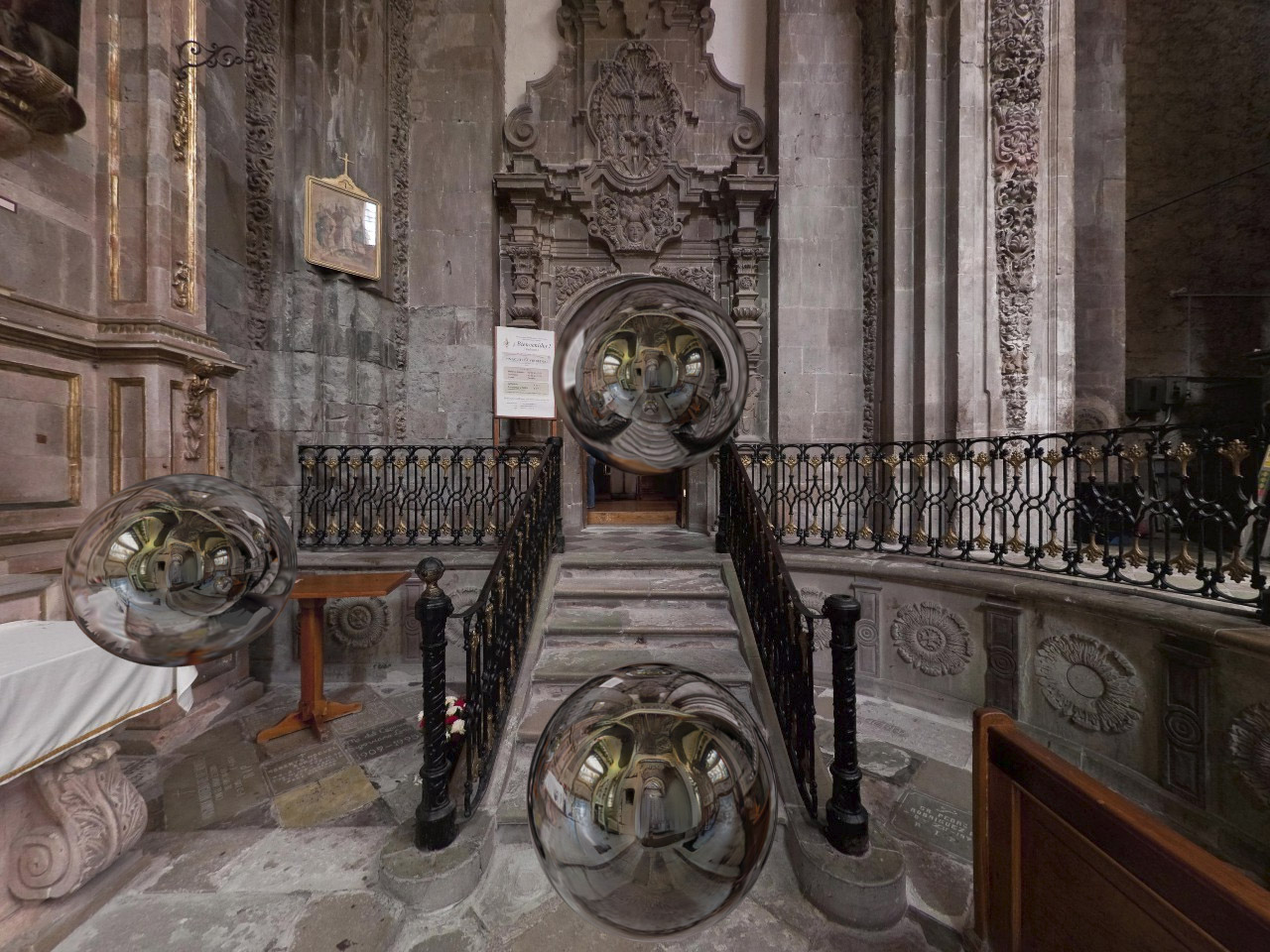} 
    \\
    \includegraphics[height=\myheight]{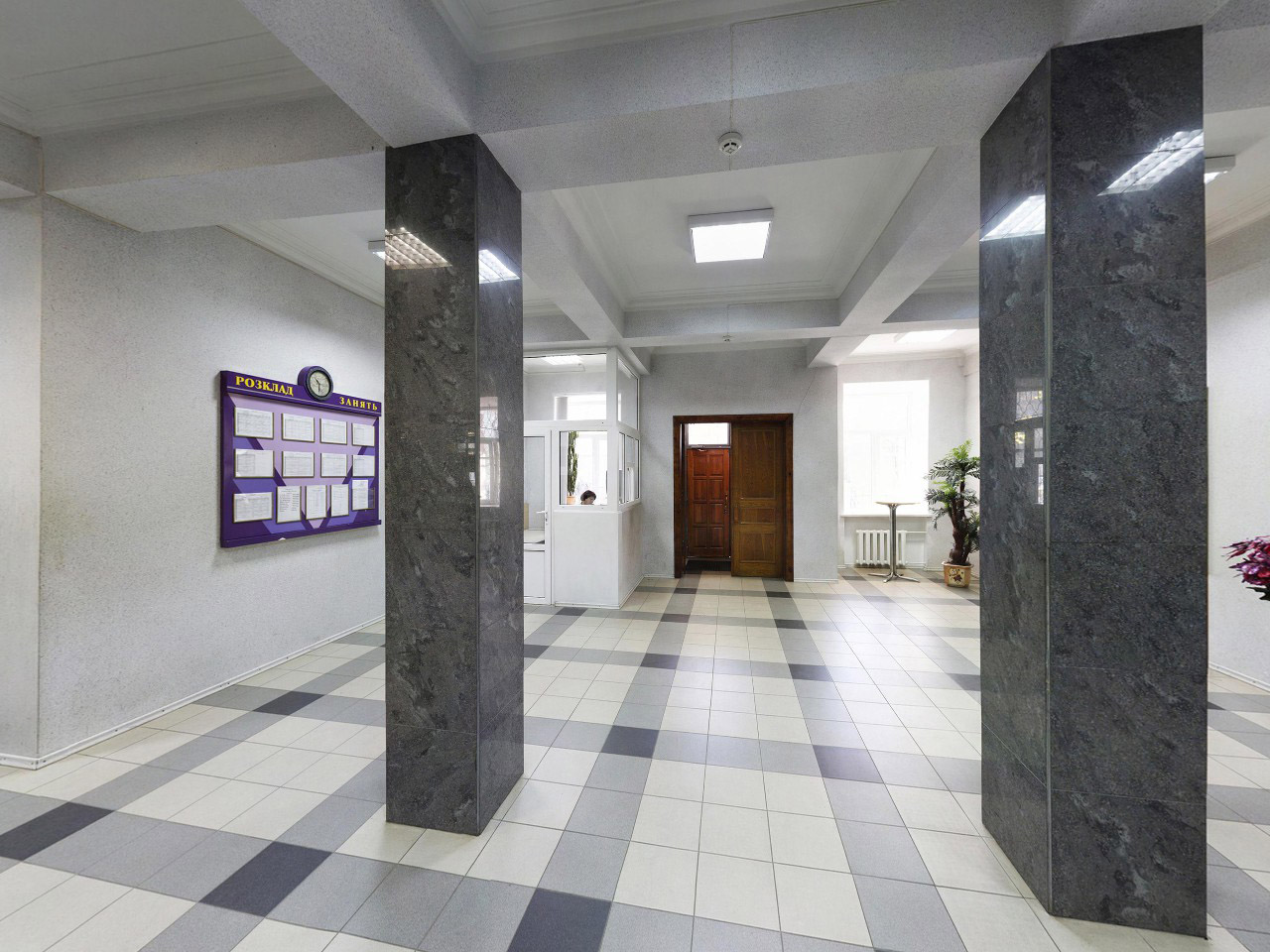}
    & \includegraphics[height=\myheight]{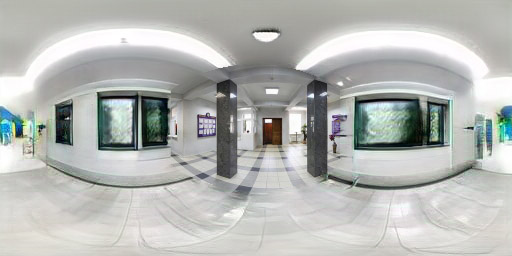}
    & \includegraphics[height=\myheight]{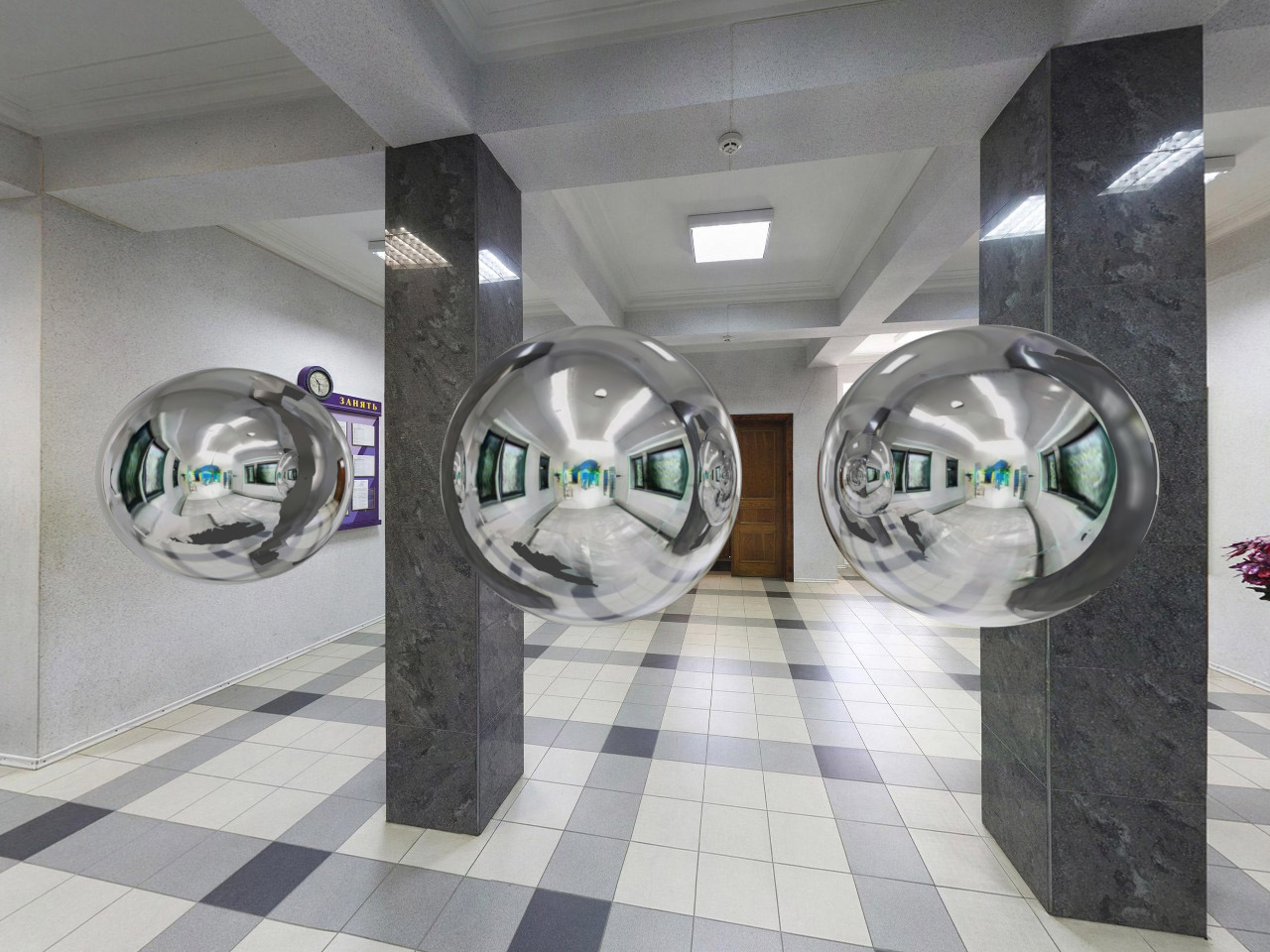} 
    \\
    \end{tabular}
    % }
    }
    \subfloat[\label{subfig:insertion-edit}]{
    % \resizebox{0.46\linewidth}{!}{
    \begin{tabular}{cccc}
    Input
    & \thename
    & 
    & Edited
    \\
    \includegraphics[height=\myheight]{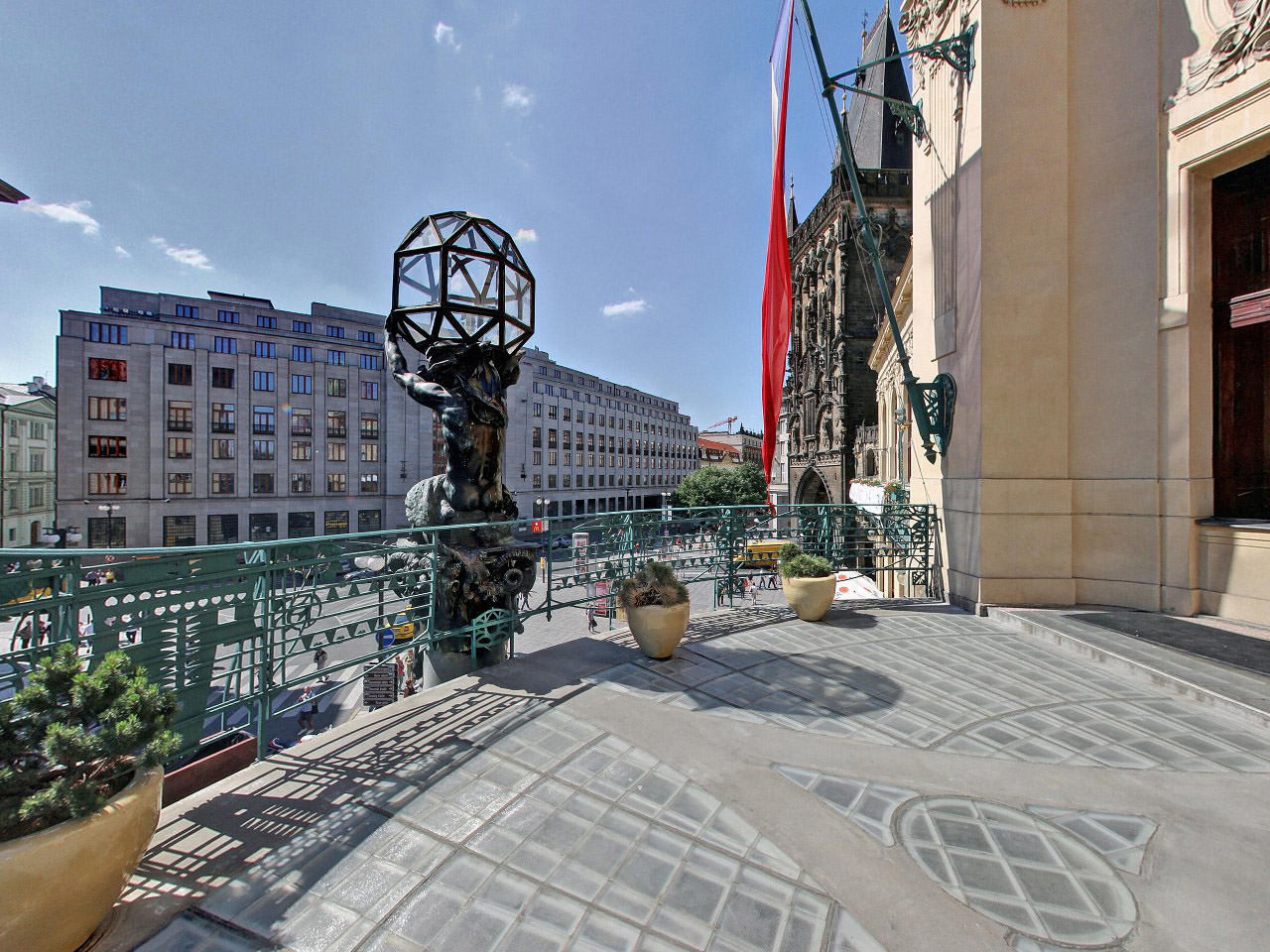}
    & \includegraphics[height=\myheight]{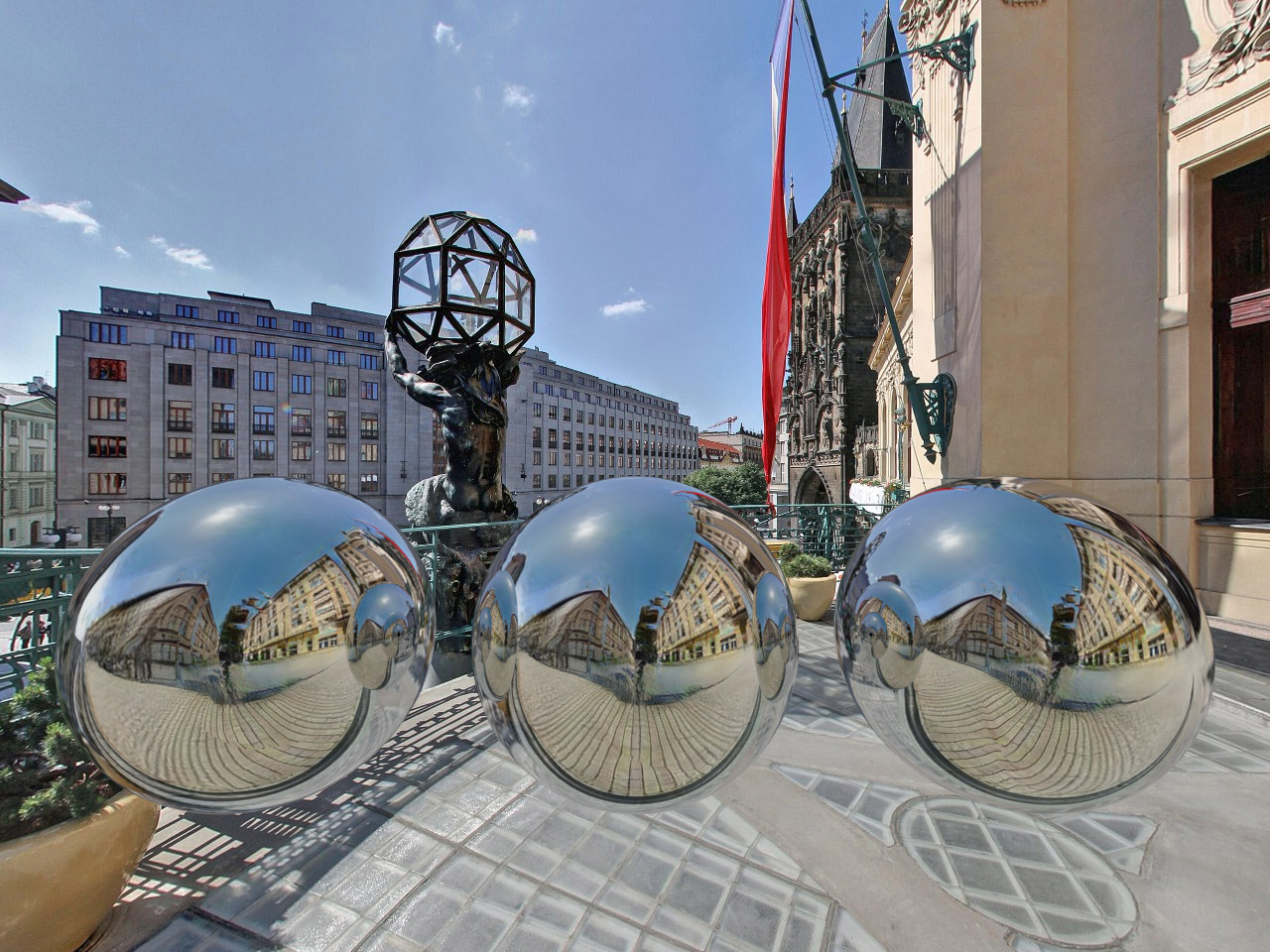} 
    & \rotatebox{90}{\scriptsize $\mapsto \text{``sky''}$} 
    & \includegraphics[height=\myheight]{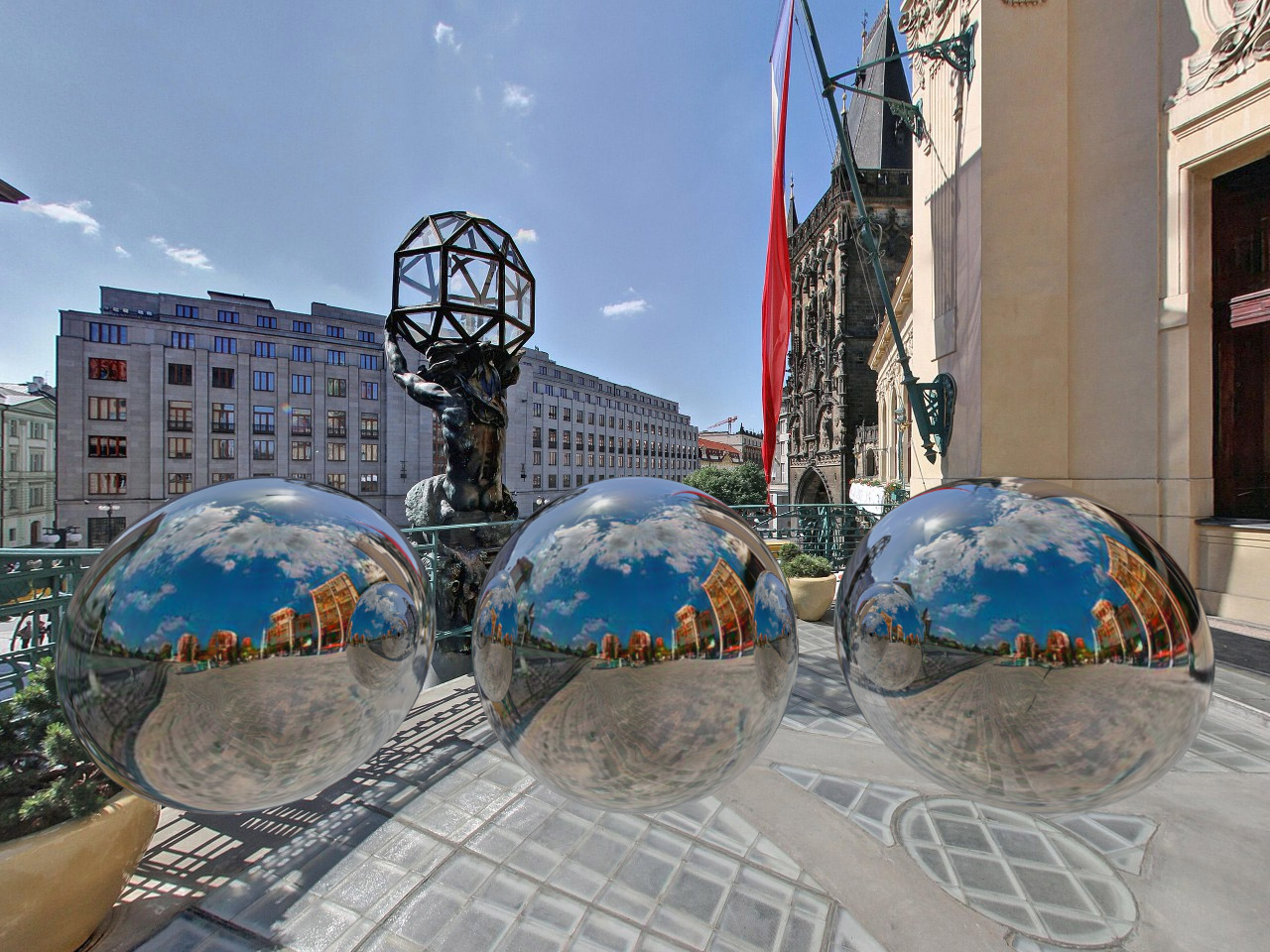} 
    \\
    \includegraphics[height=\myheight]{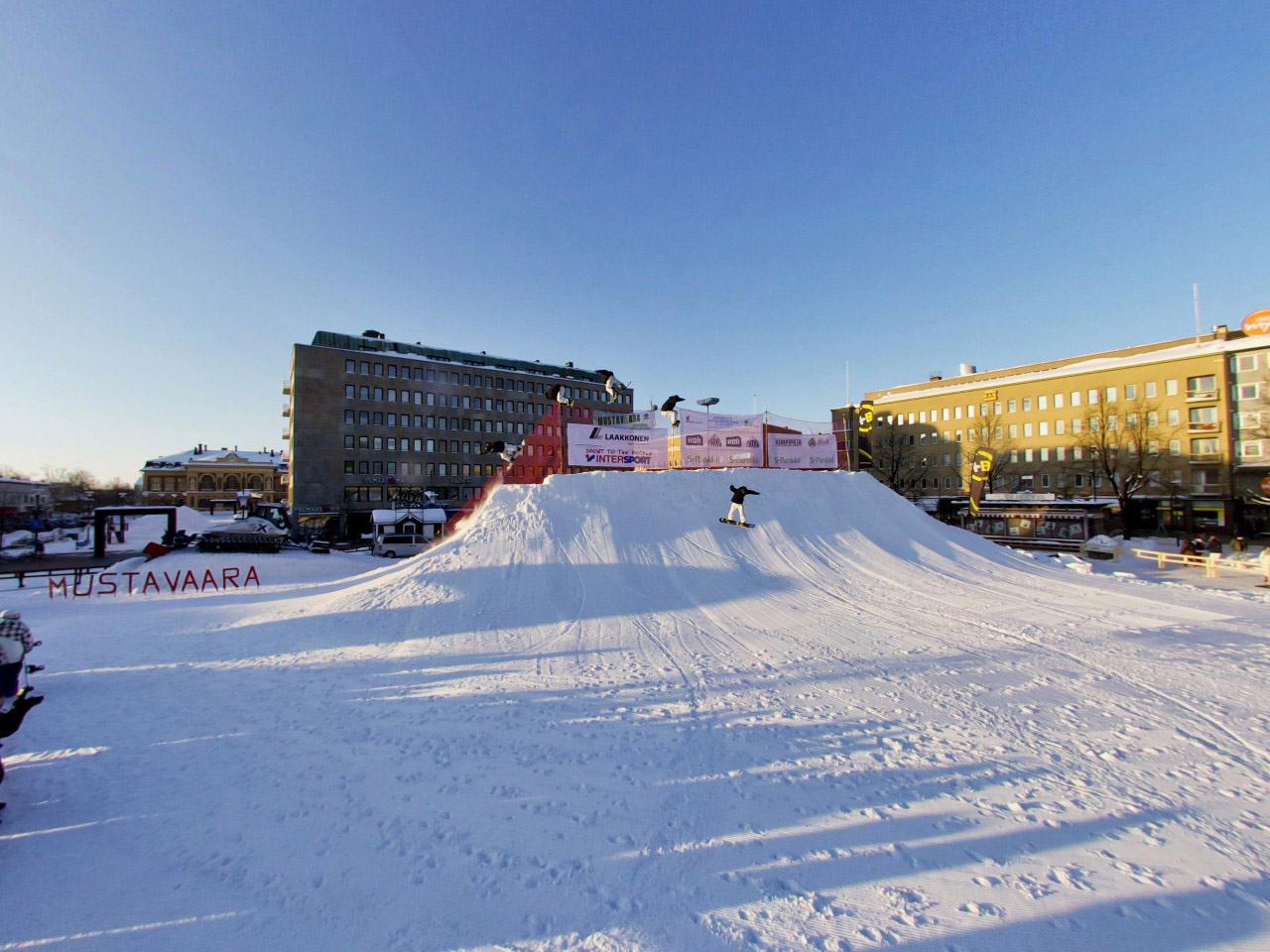}
    & \includegraphics[height=\myheight]{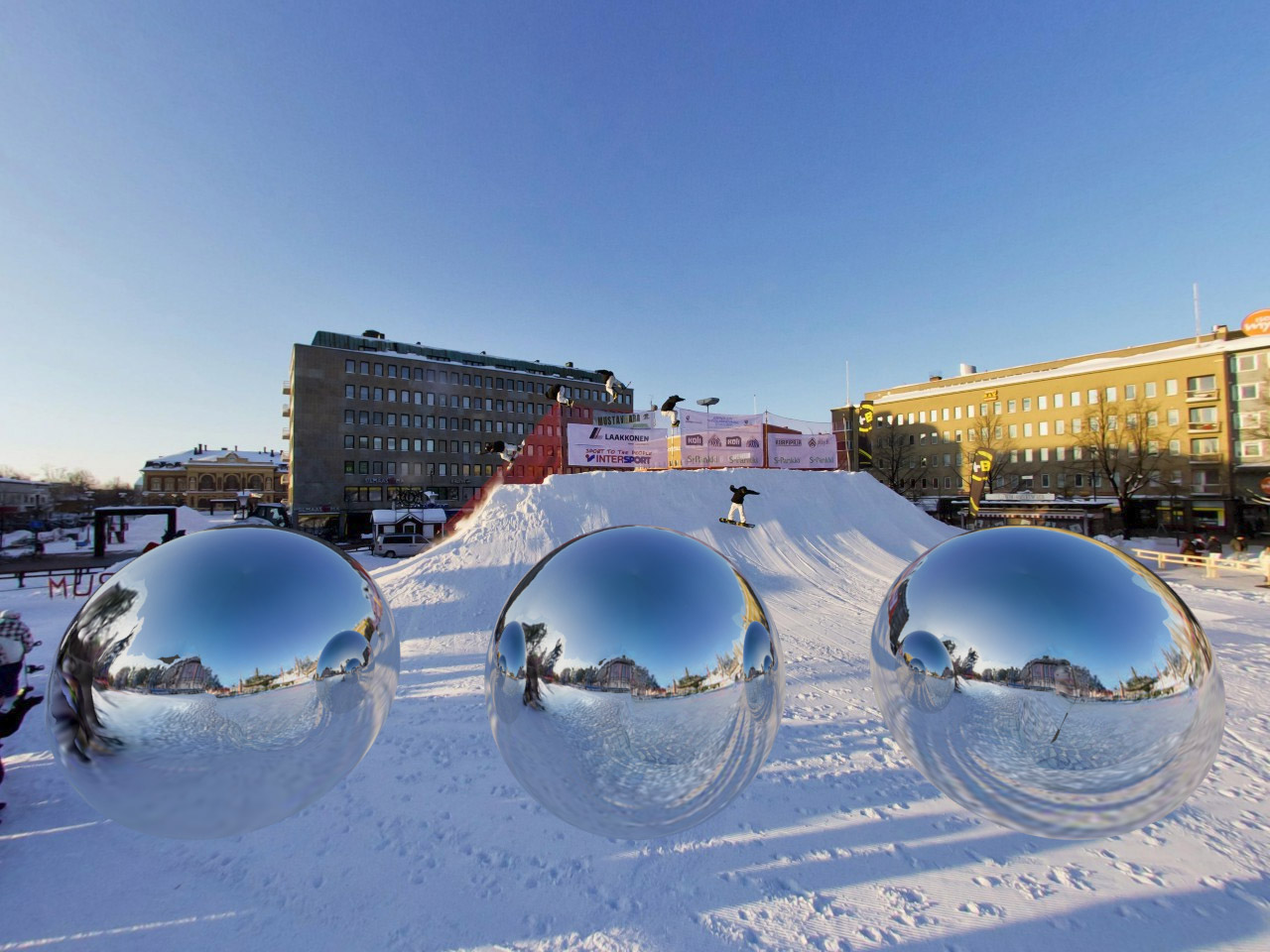} 
    & \rotatebox{90}{\scriptsize $\mapsto \text{``crtyard''}$} 
    & \includegraphics[height=\myheight]{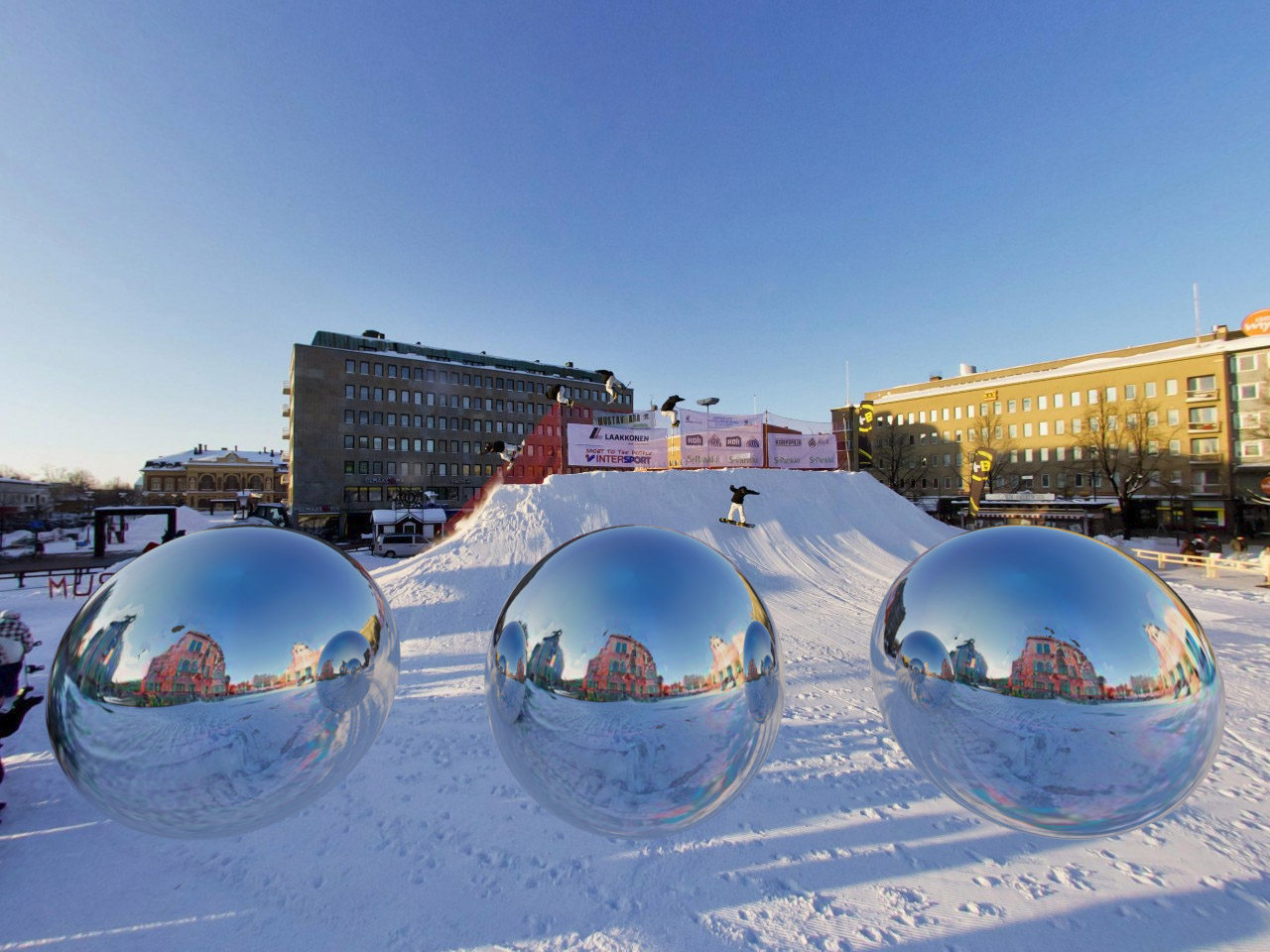} 
    \\
    \includegraphics[height=\myheight]{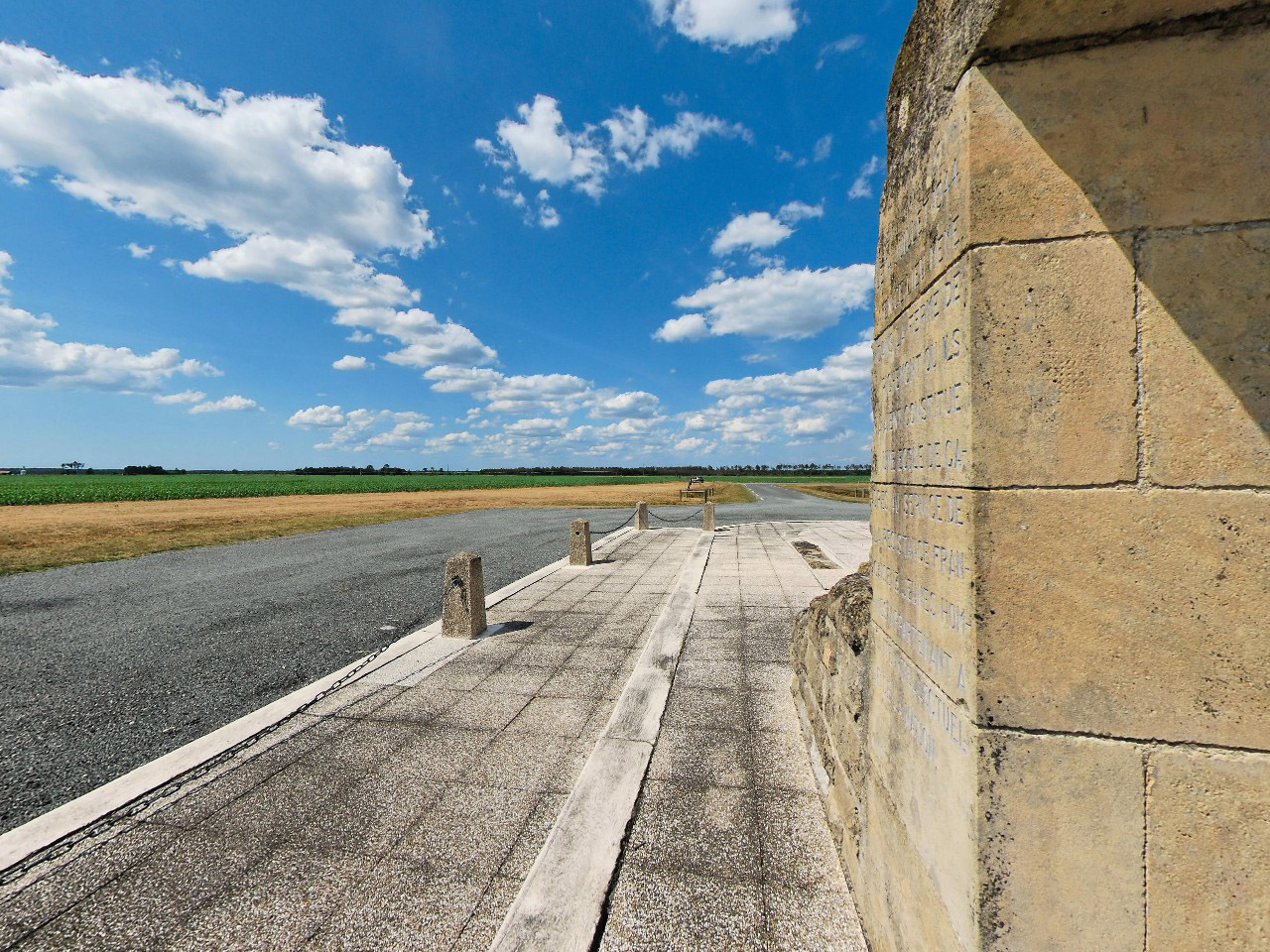}
    & \includegraphics[height=\myheight]{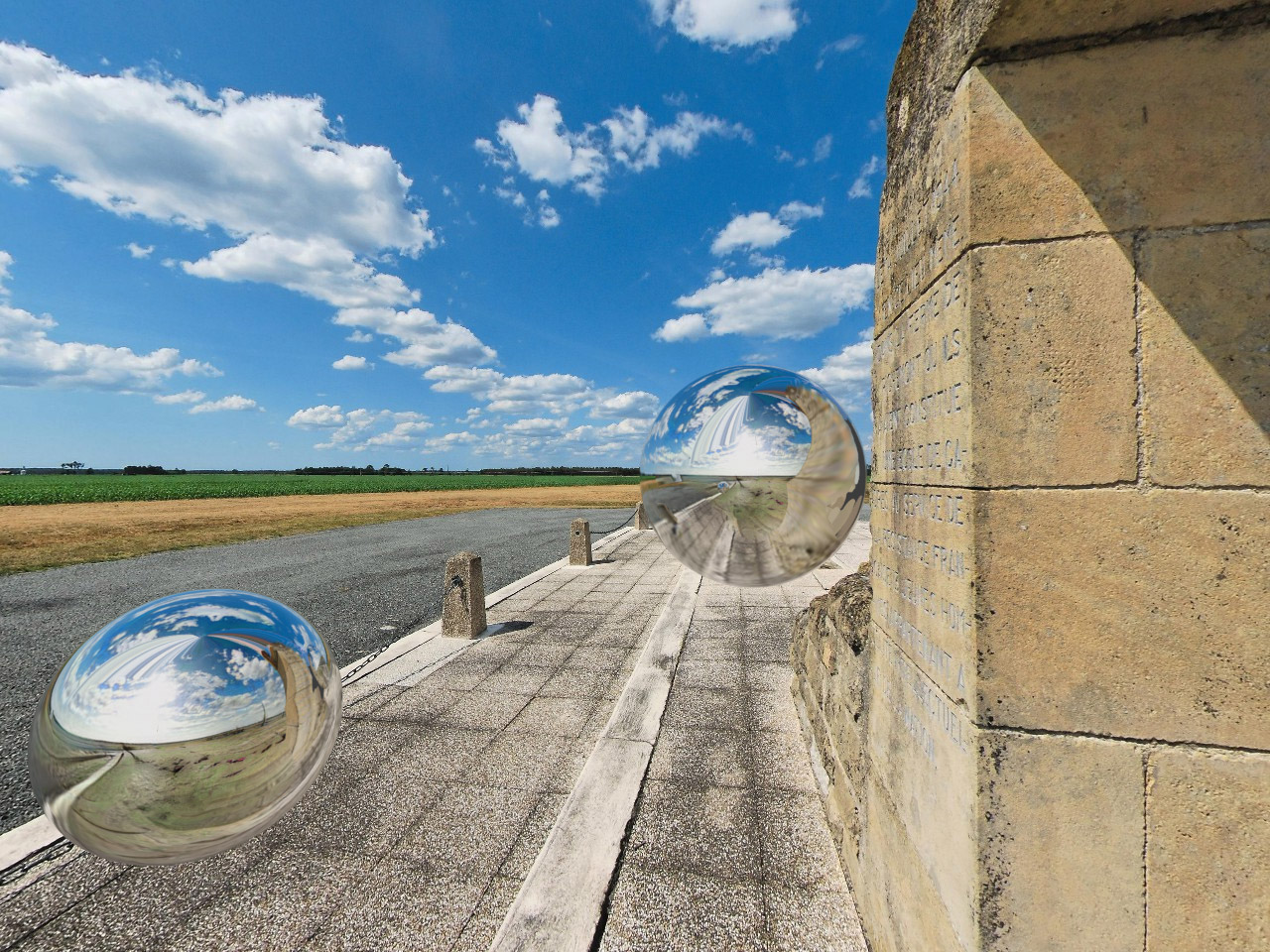} 
    & \rotatebox{90}{\scriptsize $\mapsto \text{``beach''}$} 
    & \includegraphics[height=\myheight]{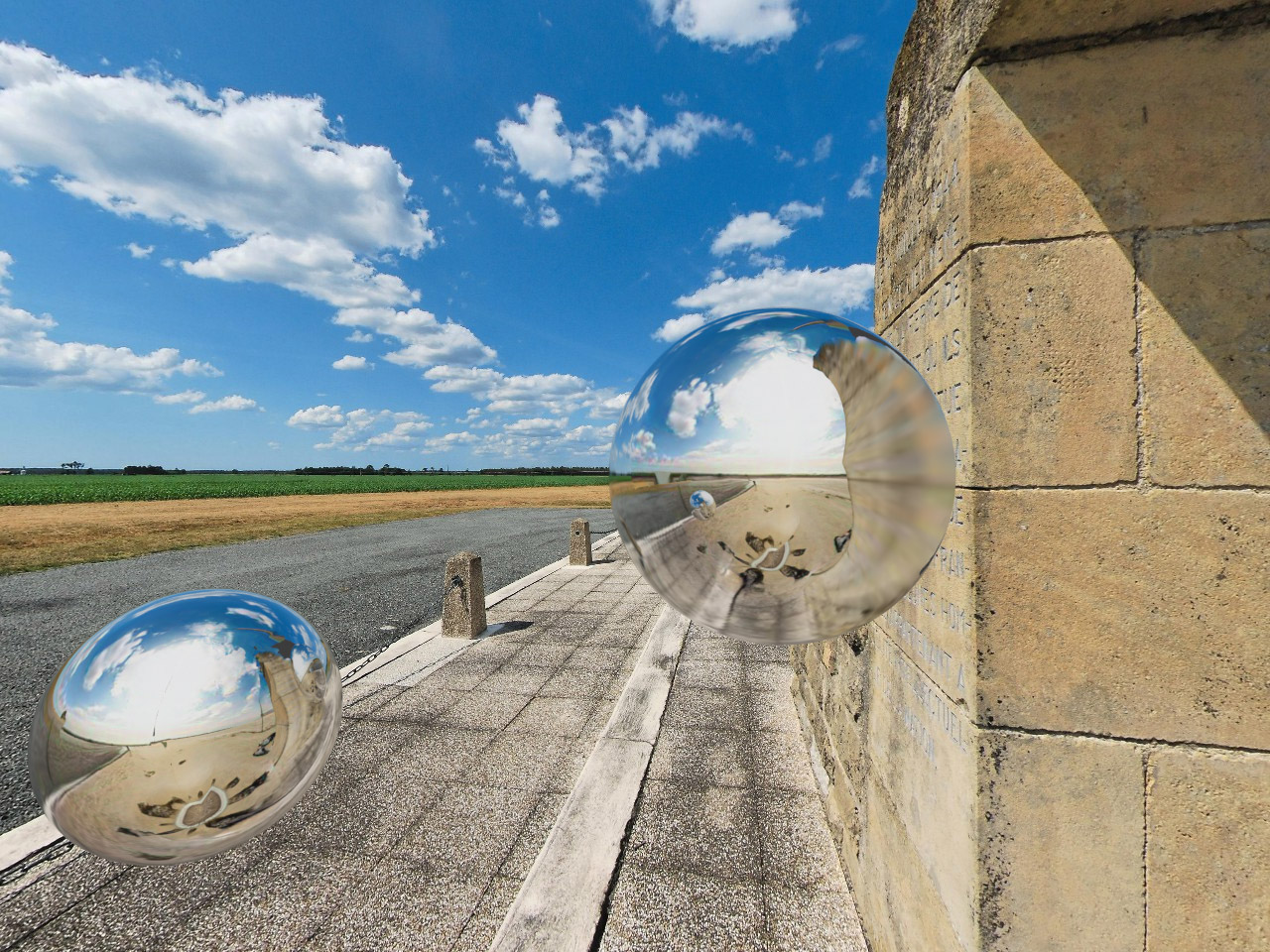} 
    \\
    \end{tabular}
    % }
    }
    \caption[]{Virtual object insertion results. \subref{subfig:insertion-orig} Our extrapolated panoramas can be used to realistically insert shiny virtual objects, which are particularly challenging since they reflect the entire environment. \subref{subfig:insertion-edit} With our guided co-modulation technique, a user can control the semantic content of the reflections on virtual objects, by generating more dramatic clouds ($\mapsto \text{``sky''}$), suggesting a more urban environment ($\mapsto \text{``courtyard''}$), and adding more sand ($\mapsto \text{``beach''}$).}
     % By estimating a depth map from the panoramas and fitting a mesh to the output, we can simulate spatially-varying reflections. }
    \label{fig:object-insertion}
\end{figure*}

%% file: fig_obj_insert_comp.tex
\begin{figure} [t]
\scriptsize
\centering
\newcommand{\myheight}{0.16\linewidth}
\newcommand{\mywidth}{0.16\linewidth}
\setlength{\tabcolsep}{1pt}
\begin{tabular}{ccccc}
Input image & Wang~\etal~\cite{wang2021learning} &\cite{wang2021learning} zoom & Ours & Ours zoom \\
\includegraphics[height=\myheight]{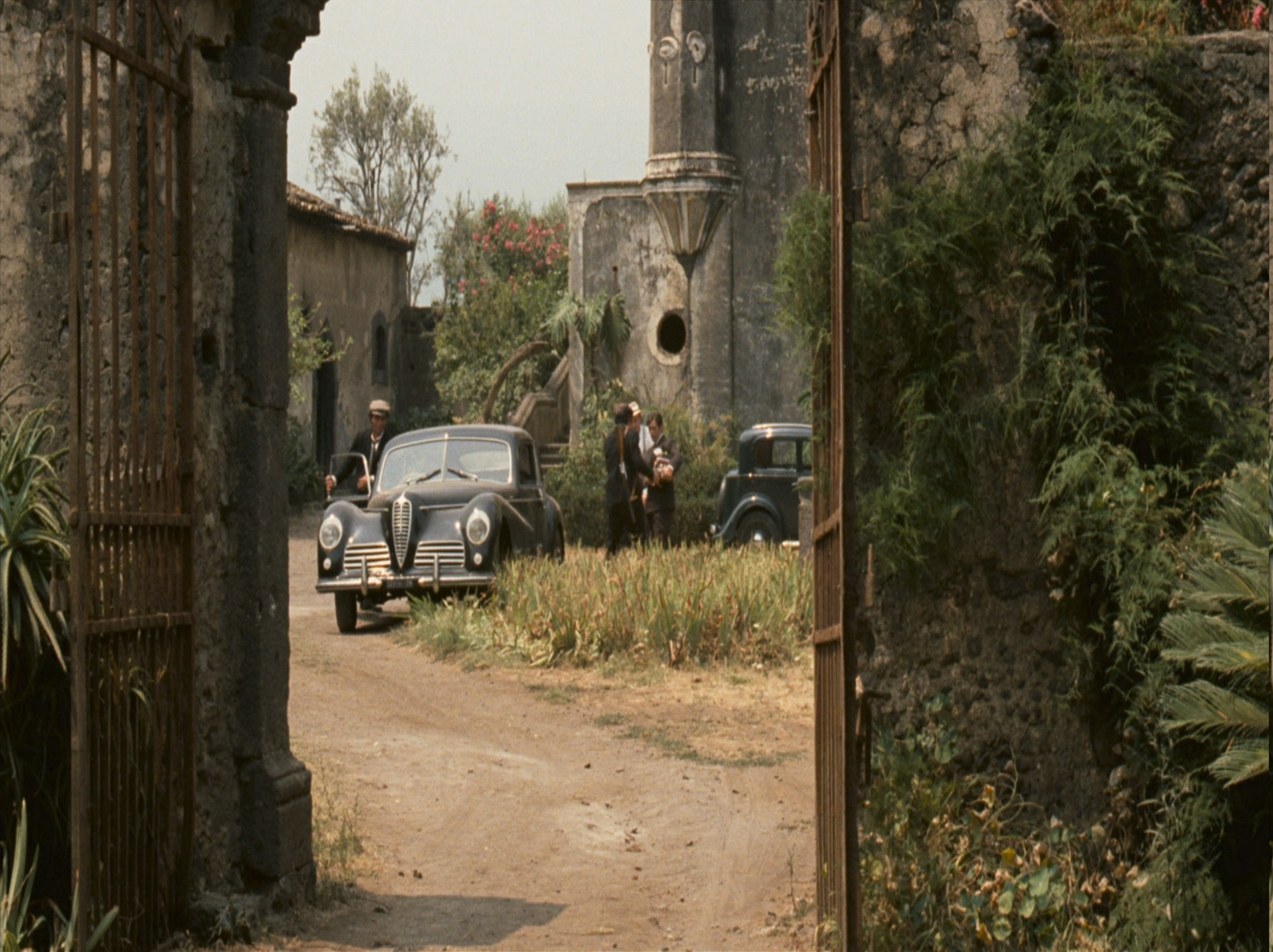} & 
\includegraphics[height=\myheight]{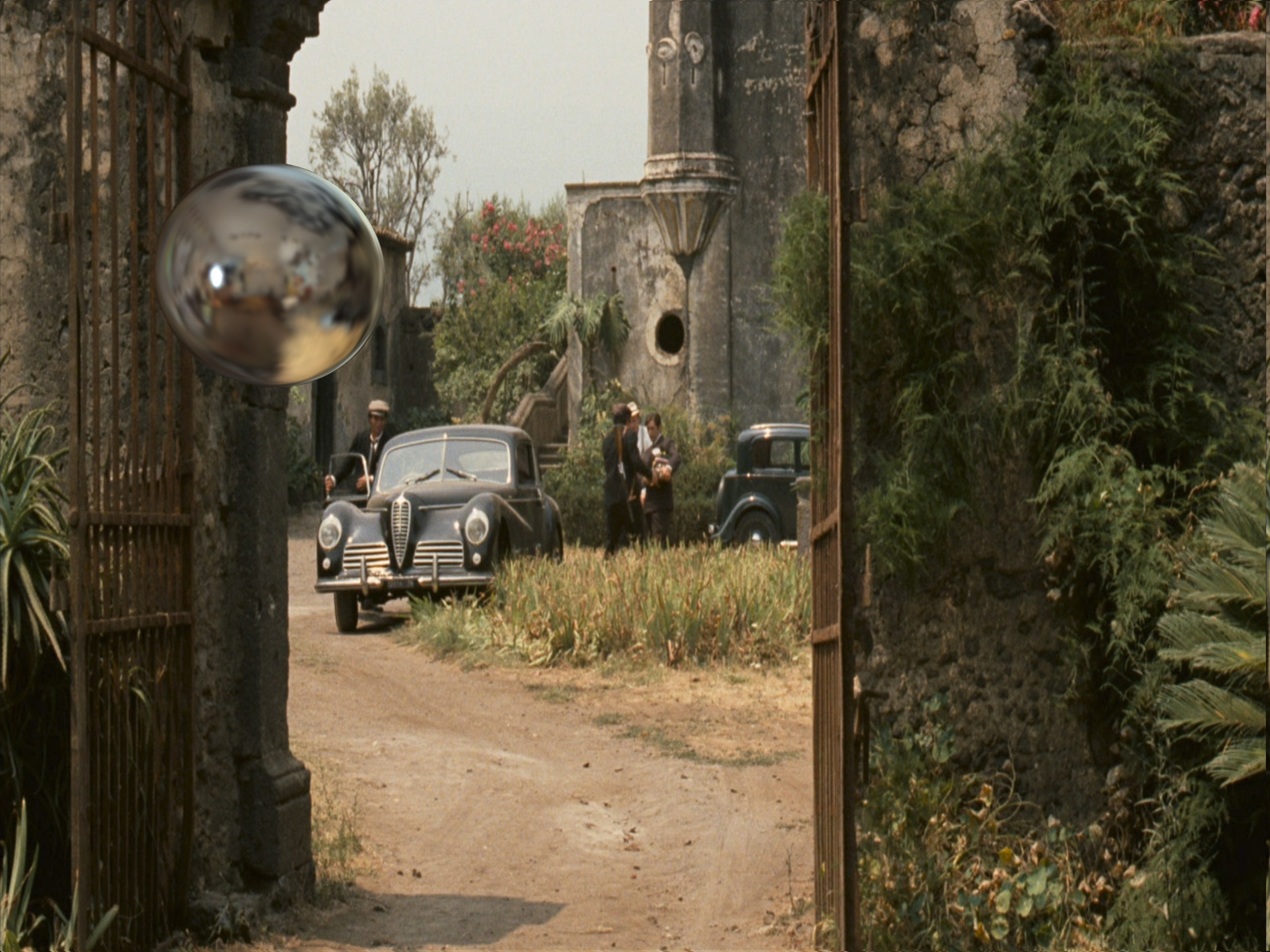} &
\includegraphics[height=\myheight, width=\mywidth]{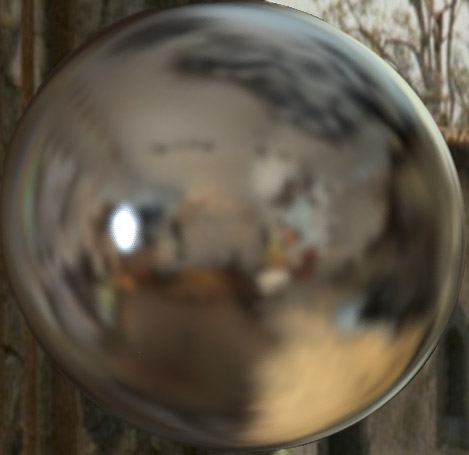} & 
\includegraphics[height=\myheight]{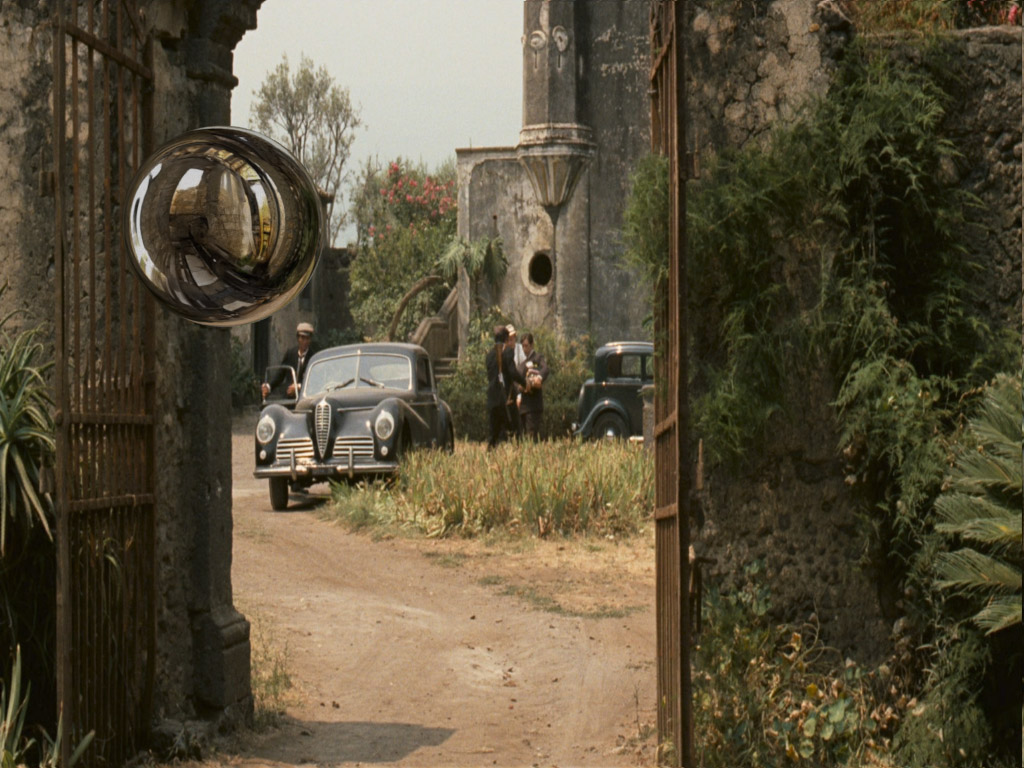} &
\includegraphics[height=\myheight, width=\mywidth]{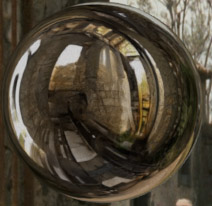} \\ 
\includegraphics[height=\myheight]{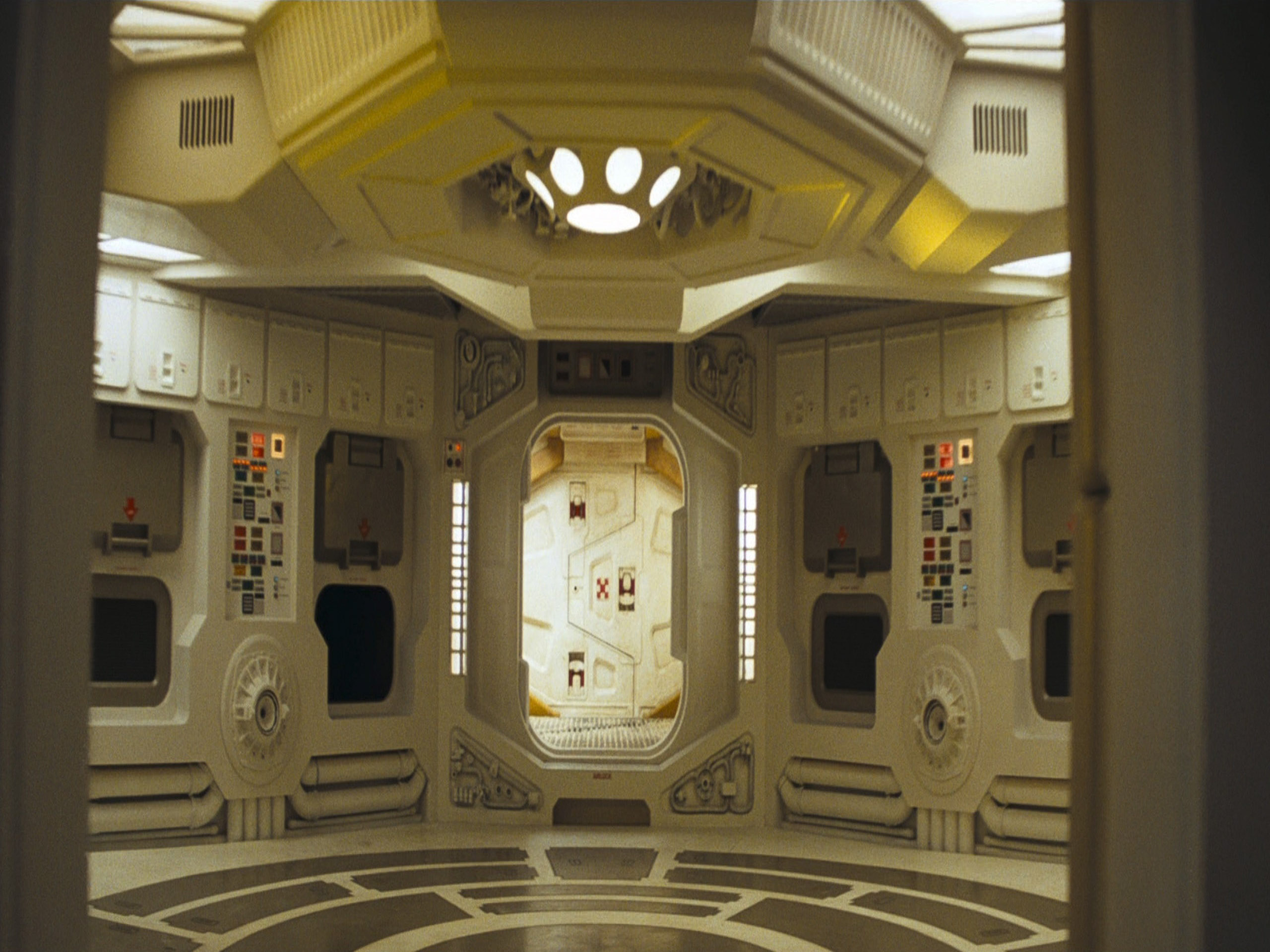} & 
\includegraphics[height=\myheight]{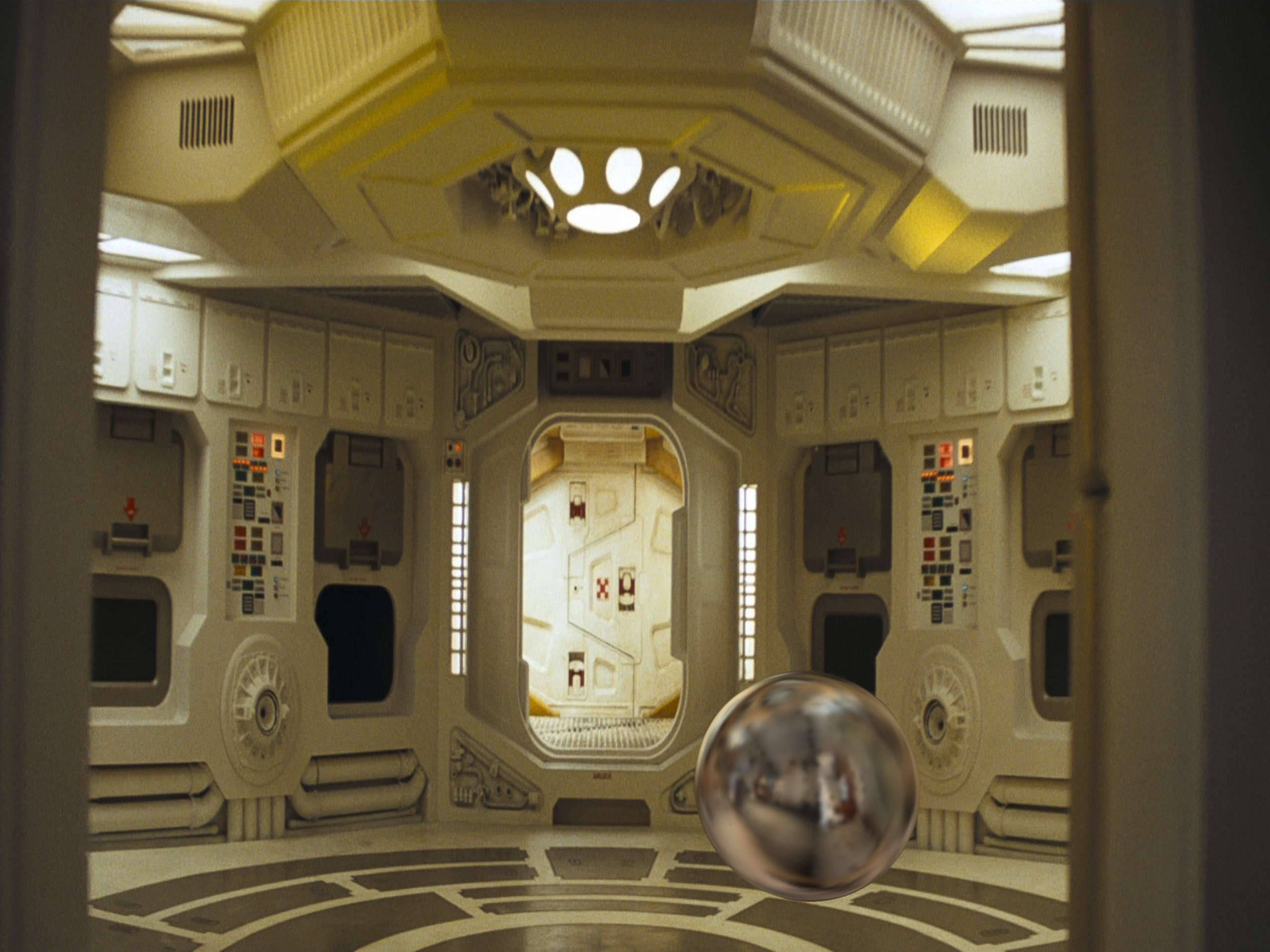} &
\includegraphics[height=\myheight, width=\mywidth]{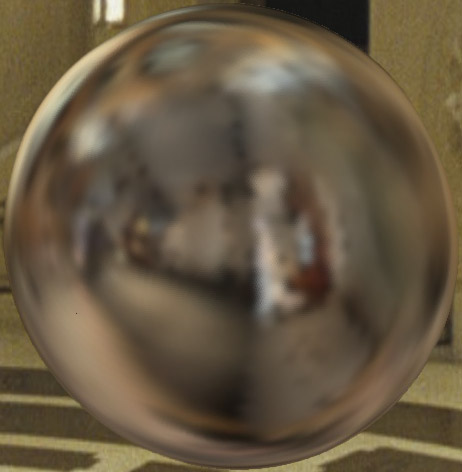} &
\includegraphics[height=\myheight]{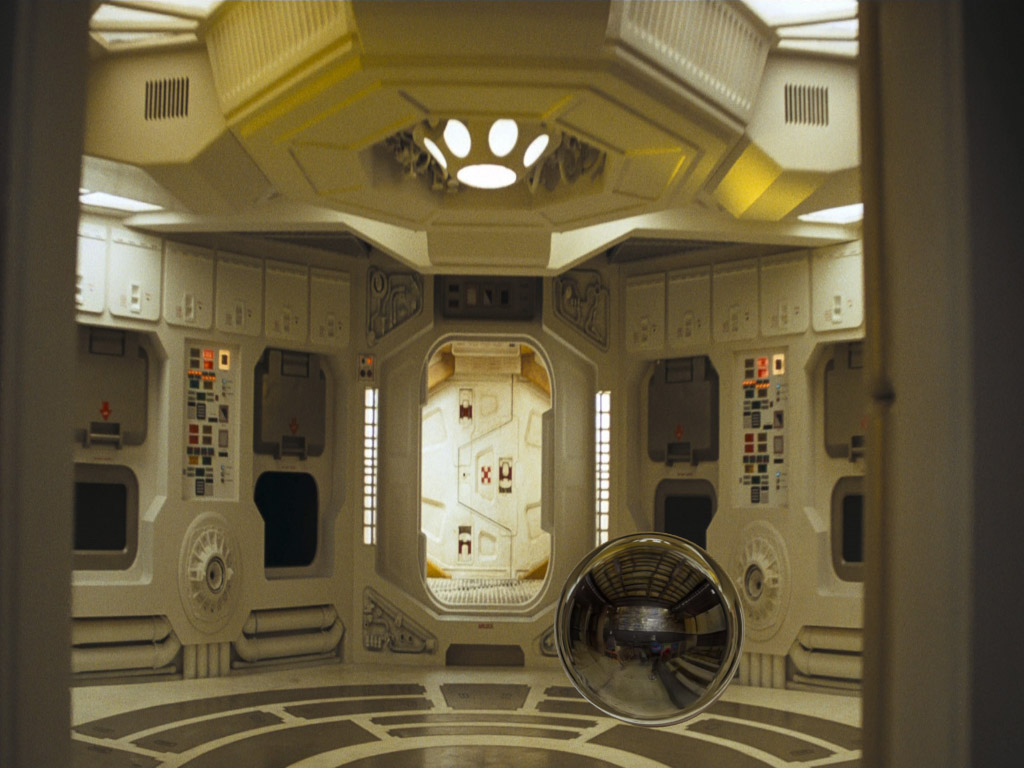} &
\includegraphics[height=\myheight, width=\mywidth]{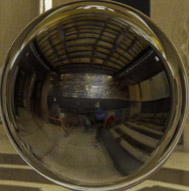} \\
\end{tabular}
\caption[]{Comparison to Wang~\etal~\cite{wang2021learning} as the state-of-the-art for virtual object insertion. Our method generates sharp results, even for out-of-domain images (images from \cite{wang2021learning}).}
\label{fig:comparison-obj-insert}
\end{figure}

%% file: 7_discussion.tex
\section{Discussion}

%\paragraph{Limitations}
As observed in \cref{fig:editing-details}, not all editing scenarios yield satisfying results. In particular, cross-domain translations (\eg, extrapolating an outdoor panorama from an input indoor image), yield, perhaps as expected, performance that is significantly below that of in-domain results. We also note an overall better performance for outdoor labels---achieving a better indoor/outdoor balance in the training dataset might help alleviate this. We also observe not all labels work equally well. It would be interesting to explore whether there is a relation between the semantic similarity between labels (\eg, with NLP tools~\cite{radford2021learning,patashnik2021styleclip}) and the visual quality of the result. Nevertheless, the analysis performed in \cref{sec:editing-experiments} could help determining ``good'' labels to suggest in an interactive application, for example. 
Another limitation is that our proposed method allows for a global edit that modifies the entire extrapolated FOV. We are confident this framework can be extended to local edits either by leveraging the spatiality of the noise of the StyleGAN backbone or by coupling it with a local editing approach such as GAN dissection~\cite{bau2018gan}. 
Finally, the non-Euclidean nature of the sphere manifold leads to unavoidable distortions when unfolding its surface to 2D. It would be more natural to frame 360\degree{} FOV extrapolation using spherical representations~\cite{lee2019spherephd,perraudin2019deepsphere,fernandez2020corners}, which presents another potential exciting direction for future work. 

%\paragraph{Conclusion}
In this work, we present a method for FOV extrapolation to a full 360\degree{} around the camera which leverages a novel guided co-modulation mechanism for controlling the appearance of the extrapolated content. The resulting spherical panoramas outperform the state-of-the-art both in terms of the standard FID metric and of visual quality. Through a detailed evaluation of our guidance-based editing mechanism, we quantify the expected visual accuracy of the generated results for a combination of $\text{input image} \mapsto \text{target label}$ mappings, and show that high quality results can be obtained in many semantically meaningful scenarios. We demonstrate the usefulness of our approach on the application of virtual content insertion in the challenging case of highly reflective objects. We hope our approach paves the way for richer environment understanding, high-quality image editing and for helping artists realize their vision more swiftly.

% {\small
\paragraph{Acknowledgements}
This work was partially supported by NSERC grant ALLRP 557208-20. We thank V. Kim, S. Amirghodsi, E. Shechtman and K. Kulkarni for helpful discussions, and everyone at UL who helped with proofreading.
% }
%
% The obtained spherical panoramas are especially useful when inserting virtual objects that bears reflective materials, showcasing the richness of the hallucinated regions. 
%
% While those panoramas are useful, their fixed content can be underwhelming for artists who wants to convey a specific atmosphere through their composite image. To solve this issue, we propose an editing method that allows to tune the generated content to one of the several hundred types of environments available in our off-the-shelf pretrained guide. 
%
% We hope our approach paves the way for environment understanding and high-quality image editing and help artists conceive their works more swiftly. 
% We believe our method show promises in being combined with other guiding signals such as human or face detection, or existing style editing methods. 

%Future work: test other architectures for guidance (e.g. ResNet on Places365) or test the guidance mechanism on other tasks (like faces) and compare to existing ways of performing style editing (can't do it here when input/output don't match).

% \paragraph{Broader impact}

% Any GAN or virtual object insertion method can be employed for pernicious use. By allowing control over the generation of high quality panoramas, our paper enables the creation of more realistic images, making it more difficult to detect them. We condemn misusing these technologies, and promote solutions such as authentication. 

%% file: main.bbl
\begin{thebibliography}{10}\itemsep=-1pt

\bibitem{abbasi2019deep}
Ali Abbasi, Sinan Kalkan, and Yusuf Sahillio{\u{g}}lu.
\newblock Deep \uppercase{3D} semantic scene extrapolation.
\newblock {\em The Visual Computer}, 2019.

\bibitem{abdal2019image2stylegan}
Rameen Abdal, Yipeng Qin, and Peter Wonka.
\newblock {Image2StyleGAN: How to embed images into the StyleGAN latent space?}
\newblock In {\em IEEE Computer Vision and Pattern Recognition Conference},
  2019.

\bibitem{akimoto2019360}
Naofumi Akimoto, Seito Kasai, Masaki Hayashi, and Yoshimitsu Aoki.
\newblock 360-degree image completion by two-stage conditional
  \uppercase{GAN}s.
\newblock In {\em IEEE International Conference on Image Processing}, 2019.

\bibitem{akimoto2022diverse}
Naofumi Akimoto, Yuhi Matsuo, and Yoshimitsu Aoki.
\newblock Diverse plausible 360-degree image outpainting for efficient
  3\uppercase{DCG} background creation.
\newblock {\em IEEE Computer Vision and Pattern Recognition Conference}, 2022.

\bibitem{barnes2009patchmatch}
Connelly Barnes, Eli Shechtman, Adam Finkelstein, and Dan~B Goldman.
\newblock Patch\uppercase{M}atch: A randomized correspondence algorithm for
  structural image editing.
\newblock {\em ACM Transactions on Graphics}, 2009.

\bibitem{bau2019semantic}
David Bau, Hendrik Strobelt, William Peebles, Jonas Wulff, Bolei Zhou, Jun-Yan
  Zhu, and Antonio Torralba.
\newblock Semantic photo manipulation with a generative image prior.
\newblock {\em ACM Transactions on Graphics}, 2019.

\bibitem{bau2018gan}
David Bau, Jun-Yan Zhu, Hendrik Strobelt, Bolei Zhou, Joshua~B. Tenenbaum,
  William~T. Freeman, and Antonio Torralba.
\newblock \uppercase{GAN} dissection: Visualizing and understanding generative
  adversarial networks.
\newblock In {\em International Conference on Learning Representations}, 2019.

\bibitem{chai2021using}
Lucy Chai, Jonas Wulff, and Phillip Isola.
\newblock Using latent space regression to analyze and leverage
  compositionality in \uppercase{GAN}s.
\newblock In {\em International Conference on Learning Representations}, 2021.

\bibitem{cheng2020segvae}
Yen-Chi Cheng, Hsin-Ying Lee, Min Sun, and Ming-Hsuan Yang.
\newblock Controllable image synthesis via {SegVAE}.
\newblock In {\em European Conference on Computer Vision}, 2020.

\bibitem{cheng2021out}
Yen-Chi Cheng, Chieh~Hubert Lin, Hsin-Ying Lee, Jian Ren, Sergey Tulyakov, and
  Ming-Hsuan Yang.
\newblock {In\&Out: Diverse image outpainting via GAN inversion}.
\newblock {\em arXiv:2104.00675}, 2021.

\bibitem{chong2020effectively}
Min~Jin Chong and David Forsyth.
\newblock Effectively unbiased \uppercase{fid} and \uppercase{I}nception
  \uppercase{S}core and where to find them.
\newblock In {\em IEEE Computer Vision and Pattern Recognition Conference},
  2020.

\bibitem{blender}
Blender~Online Community.
\newblock {\em Blender - a 3D modelling and rendering package}.
\newblock Blender Foundation, Stichting Blender Foundation, Amsterdam, 2018.

\bibitem{Debevec1998}
Paul Debevec.
\newblock Rendering synthetic objects into real scenes: Bridging traditional
  and image-based graphics with global illumination and high dynamic range
  photography.
\newblock In {\em Proceedings of ACM SIGGRAPH}, 1998.

\bibitem{demir2018patch}
Ugur Demir and Gozde Unal.
\newblock Patch-based image inpainting with generative adversarial networks.
\newblock {\em arXiv preprint arXiv:1803.07422}, 2018.

\bibitem{efros2001image}
Alexei~A Efros and William~T Freeman.
\newblock Image quilting for texture synthesis and transfer.
\newblock In {\em Proceedings of the 28th annual conference on Computer
  graphics and interactive techniques}, 2001.

\bibitem{efros1999texture}
Alexei~A Efros and Thomas~K Leung.
\newblock Texture synthesis by non-parametric sampling.
\newblock In {\em International Conference on Computer Vision}, 1999.

\bibitem{esser2021taming}
Patrick Esser, Robin Rombach, and Bjorn Ommer.
\newblock Taming transformers for high-resolution image synthesis.
\newblock In {\em IEEE Computer Vision and Pattern Recognition Conference},
  2021.

\bibitem{fernandez2020corners}
Clara Fernandez-Labrador, Jose~M Facil, Alejandro Perez-Yus, C{\'e}dric
  Demonceaux, Javier Civera, and Jose~J Guerrero.
\newblock Corners for layout: End-to-end layout recovery from 360 images.
\newblock {\em IEEE Robotics and Automation Letters}, 2019.

\bibitem{gardner2017learning}
Marc-Andr{\'e} Gardner, Kalyan Sunkavalli, Ersin Yumer, Xiaohui Shen, Emiliano
  Gambaretto, Christian Gagn{\'e}, and Jean-Fran{\c{c}}ois Lalonde.
\newblock Learning to predict indoor illumination from a single image.
\newblock {\em ACM Transactions on Graphics}, 2017.

\bibitem{garg2019learning}
Rahul Garg, Neal Wadhwa, Sameer Ansari, and Jonathan~T Barron.
\newblock Learning single camera depth estimation using dual-pixels.
\newblock In {\em International Conference on Computer Vision}, 2019.

\bibitem{Gehrig21ral}
Mathias Gehrig, Willem Aarents, Daniel Gehrig, and Davide Scaramuzza.
\newblock \uppercase{DSEC}: A stereo event camera dataset for driving
  scenarios.
\newblock {\em IEEE Robotics and Automation Letters}, 2021.

\bibitem{goetschalckx2019ganalyze}
Lore Goetschalckx, Alex Andonian, Aude Oliva, and Phillip Isola.
\newblock Ganalyze: Toward visual definitions of cognitive image properties.
\newblock In {\em International Conference on Computer Vision}, 2019.

\bibitem{goodfellow2014generative}
Ian Goodfellow, Jean Pouget-Abadie, Mehdi Mirza, Bing Xu, David Warde-Farley,
  Sherjil Ozair, Aaron Courville, and Yoshua Bengio.
\newblock Generative adversarial nets.
\newblock In {\em Conference on Neural Information Processing Systems}, 2014.

\bibitem{han2020piinet}
Seo~Woo Han and Doug~Young Suh.
\newblock \uppercase{PIINET}: A 360-degree panoramic image inpainting network
  using a cube map.
\newblock {\em arXiv preprint arXiv:2010.16003}, 2020.

\bibitem{Hara_Mukuta_Harada_2021}
Takayuki Hara, Yusuke Mukuta, and Tatsuya Harada.
\newblock Spherical image generation from a single image by considering scene
  symmetry.
\newblock In {\em Association for the Advancement of Artificial Intelligence},
  2021.

\bibitem{he2016deep}
Kaiming He, Xiangyu Zhang, Shaoqing Ren, and Jian Sun.
\newblock Deep residual learning for image recognition.
\newblock In {\em IEEE Computer Vision and Pattern Recognition Conference},
  2016.

\bibitem{hold2017perceptual}
Yannick Hold-Geoffroy, Kalyan Sunkavalli, Jonathan Eisenmann, Matt Fisher,
  Emiliano Gambaretto, Sunil Hadap, and Jean-Fran{\c{c}}ois Lalonde.
\newblock A perceptual measure for deep single image camera calibration.
\newblock In {\em IEEE Computer Vision and Pattern Recognition Conference},
  2018.

\bibitem{huang2020semantic}
Hsin-Ping Huang, Hung-Yu Tseng, Hsin-Ying Lee, and Jia-Bin Huang.
\newblock {Semantic view synthesis}.
\newblock In {\em European Conference on Computer Vision}, 2020.

\bibitem{harkonen2020ganspace}
Erik Härkönen, Aaron Hertzmann, Jaakko Lehtinen, and Sylvain Paris.
\newblock \uppercase{GANS}pace: Discovering interpretable \uppercase{GAN}
  controls.
\newblock In {\em Conference on Neural Information Processing Systems}, 2020.

\bibitem{isola2017image}
Phillip Isola, Jun-Yan Zhu, Tinghui Zhou, and Alexei~A Efros.
\newblock Image-to-image translation with conditional adversarial networks.
\newblock In {\em IEEE Computer Vision and Pattern Recognition Conference},
  2017.

\bibitem{jahanian2019steerability}
Ali Jahanian, Lucy Chai, and Phillip Isola.
\newblock On the" steerability" of generative adversarial networks.
\newblock {\em International Conference on Learning Representations}, 2019.

\bibitem{jo2021n}
Changho Jo, Woobin Im, and Sung-Eui Yoon.
\newblock In-n-out: Towards good initialization for inpainting and outpainting.
\newblock {\em British Machine Vision Conference}, 2021.

\bibitem{Karras2021_stylegan3}
Tero Karras, Miika Aittala, Samuli Laine, Erik H\"ark\"onen, Janne Hellsten,
  Jaakko Lehtinen, and Timo Aila.
\newblock Alias-free generative adversarial networks.
\newblock In {\em Conference on Neural Information Processing Systems}, 2021.

\bibitem{karras2019_stylegan}
Tero Karras, Samuli Laine, and Timo Aila.
\newblock A style-based generator architecture for generative adversarial
  networks.
\newblock In {\em IEEE Computer Vision and Pattern Recognition Conference},
  2019.

\bibitem{karras2020_stylegan2}
Tero Karras, Samuli Laine, Miika Aittala, Janne Hellsten, Jaakko Lehtinen, and
  Timo Aila.
\newblock Analyzing and improving the image quality of
  \uppercase{S}tyle\uppercase{GAN}.
\newblock In {\em IEEE Computer Vision and Pattern Recognition Conference},
  2020.

\bibitem{kimura2018extvision}
Naoki Kimura and Jun Rekimoto.
\newblock Ext\uppercase{V}ision: augmentation of visual experiences with
  generation of context images for a peripheral vision using deep neural
  network.
\newblock In {\em Proceedings of the CHI Conference on Human Factors in
  Computing Systems}, 2018.

\bibitem{lee2019spherephd}
Yeonkun Lee, Jaeseok Jeong, Jongseob Yun, Wonjune Cho, and Kuk-Jin Yoon.
\newblock \uppercase{S}phere\uppercase{phd}: Applying \uppercase{cnn}s on a
  spherical polyhedron representation of 360 degree images.
\newblock In {\em IEEE Computer Vision and Pattern Recognition Conference},
  2019.

\bibitem{MegaDepthLi18}
Zhengqi Li and Noah Snavely.
\newblock Megadepth: Learning single-view depth prediction from internet
  photos.
\newblock In {\em IEEE Computer Vision and Pattern Recognition Conference},
  2018.

\bibitem{lin2019coco}
Chieh~Hubert Lin, Chia-Che Chang, Yu-Sheng Chen, Da-Cheng Juan, Wei Wei, and
  Hwann-Tzong Chen.
\newblock \uppercase{Coco-gan}: Generation by parts via conditional
  coordinating.
\newblock In {\em International Conference on Computer Vision}, 2019.

\bibitem{lin2021infinitygan}
Chieh~Hubert Lin, Hsin-Ying Lee, Yen-Chi Cheng, Sergey Tulyakov, and Ming-Hsuan
  Yang.
\newblock Infinity{GAN}: Towards infinite-pixel image synthesis.
\newblock In {\em International Conference on Learning Representations}, 2022.

\bibitem{liu2021infinite}
Andrew Liu, Richard Tucker, Varun Jampani, Ameesh Makadia, Noah Snavely, and
  Angjoo Kanazawa.
\newblock {Infinite Nature: Perpetual view generation of natural scenes from a
  single image}.
\newblock In {\em International Conference on Computer Vision}, 2021.

\bibitem{madai2016revisiting}
Lorand Madai-Tahy, Sebastian Otte, Richard Hanten, and Andreas Zell.
\newblock Revisiting deep convolutional neural networks for \uppercase{RGB-D}
  based object recognition.
\newblock In {\em International Conference on Artificial Neural Networks},
  2016.

\bibitem{mirza2014conditional}
Mehdi Mirza and Simon Osindero.
\newblock Conditional generative adversarial nets.
\newblock {\em arXiv preprint arXiv:1411.1784}, 2014.

\bibitem{park2019semantic}
Taesung Park, Ming-Yu Liu, Ting-Chun Wang, and Jun-Yan Zhu.
\newblock Semantic image synthesis with spatially-adaptive normalization.
\newblock In {\em IEEE Computer Vision and Pattern Recognition Conference},
  2019.

\bibitem{patashnik2021styleclip}
Or Patashnik, Zongze Wu, Eli Shechtman, Daniel Cohen-Or, and Dani Lischinski.
\newblock Style\uppercase{clip}: Text-driven manipulation of \uppercase{gan}
  imagery.
\newblock In {\em International Conference on Computer Vision}, 2021.

\bibitem{perraudin2019deepsphere}
Nathana{\"e}l Perraudin, Micha{\"e}l Defferrard, Tomasz Kacprzak, and Raphael
  Sgier.
\newblock {DeepSphere: Efficient spherical convolutional neural network with
  HEALPix sampling for cosmological applications}.
\newblock {\em Astronomy and Computing}, 2019.

\bibitem{qi2018semi}
Xiaojuan Qi, Qifeng Chen, Jiaya Jia, and Vladlen Koltun.
\newblock Semi-parametric image synthesis.
\newblock In {\em IEEE Computer Vision and Pattern Recognition Conference},
  2018.

\bibitem{radford2021learning}
Alec Radford, Jong~Wook Kim, Chris Hallacy, Aditya Ramesh, Gabriel Goh,
  Sandhini Agarwal, Girish Sastry, Amanda Askell, Pamela Mishkin, Jack Clark,
  et~al.
\newblock Learning transferable visual models from natural language
  supervision.
\newblock In {\em International Conference on Machine Learning}, 2021.

\bibitem{radford2015unsupervised}
Alec Radford, Luke Metz, and Soumith Chintala.
\newblock Unsupervised representation learning with deep convolutional
  generative adversarial networks.
\newblock {\em International Conference on Learning Representations}, 2016.

\bibitem{roberts2021hypersim}
Mike Roberts, Jason Ramapuram, Anurag Ranjan, Atulit Kumar, Miguel~Angel
  Bautista, Nathan Paczan, Russ Webb, and Joshua~M Susskind.
\newblock Hypersim: A photorealistic synthetic dataset for holistic indoor
  scene understanding.
\newblock In {\em International Conference on Computer Vision}, 2021.

\bibitem{sabini2018painting}
Mark Sabini and Gili Rusak.
\newblock Painting outside the box: Image outpainting with \uppercase{GAN}s.
\newblock {\em arXiv preprint arXiv:1808.08483}, 2018.

\bibitem{shen2020interpreting}
Yujun Shen, Jinjin Gu, Xiaoou Tang, and Bolei Zhou.
\newblock Interpreting the latent space of \uppercase{gan}s for semantic face
  editing.
\newblock In {\em IEEE Computer Vision and Pattern Recognition Conference},
  2020.

\bibitem{shocher2020semantic}
Assaf Shocher, Yossi Gandelsman, Inbar Mosseri, Michal Yarom, Michal Irani,
  William~T Freeman, and Tali Dekel.
\newblock Semantic pyramid for image generation.
\newblock In {\em IEEE Computer Vision and Pattern Recognition Conference},
  2020.

\bibitem{silberman2012indoor}
Nathan Silberman, Derek Hoiem, Pushmeet Kohli, and Rob Fergus.
\newblock Indoor segmentation and support inference from \uppercase{rgbd}
  images.
\newblock In {\em European Conference on Computer Vision}, 2012.

\bibitem{simonyan2014very}
Karen Simonyan and Andrew Zisserman.
\newblock Very deep convolutional networks for large-scale image recognition.
\newblock In {\em International Conference on Learning Representations}, 2015.

\bibitem{somanath2021hdr}
Gowri Somanath and Daniel Kurz.
\newblock \uppercase{HDR} environment map estimation for real-time augmented
  reality.
\newblock In {\em IEEE Computer Vision and Pattern Recognition Conference},
  2021.

\bibitem{song2018im2pano3d}
Shuran Song, Andy Zeng, Angel~X Chang, Manolis Savva, Silvio Savarese, and
  Thomas Funkhouser.
\newblock {Im2Pano3D: Extrapolating 360 structure and semantics beyond the
  field of view}.
\newblock In {\em IEEE Computer Vision and Pattern Recognition Conference},
  2018.

\bibitem{lighthouse}
Pratul~P. Srinivasan, Ben Mildenhall, Matthew Tancik, Jonathan~T. Barron,
  Richard Tucker, and Noah Snavely.
\newblock Lighthouse: Predicting lighting volumes for spatially-coherent
  illumination.
\newblock In {\em IEEE Computer Vision and Pattern Recognition Conference},
  2020.

\bibitem{sumantri2020360}
Julius~Surya Sumantri and In~Kyu Park.
\newblock {360 panorama synthesis from a sparse set of images with unknown
  field of view}.
\newblock In {\em IEEE Workshop on Applications of Computer Vision}, 2020.

\bibitem{vankadari2019unsupervised}
Madhu~Babu Vankadari, Swagat Kumar, Anima Majumder, and Kaushik Das.
\newblock Unsupervised learning of monocular depth and ego-motion using
  conditional \uppercase{P}atch\uppercase{GAN}s.
\newblock In {\em International Joint Conferences on Artificial Intelligence},
  2019.

\bibitem{diode_dataset}
Igor Vasiljevic, Nick Kolkin, Shanyi Zhang, Ruotian Luo, Haochen Wang,
  Falcon~Z. Dai, Andrea~F. Daniele, Mohammadreza Mostajabi, Steven Basart,
  Matthew~R. Walter, and Gregory Shakhnarovich.
\newblock {DIODE}: {A} dense indoor and outdoor depth dataset.
\newblock {\em arXiv preprint arXiv:1908.00463}, 2019.

\bibitem{wang2019irs}
Qiang Wang, Shizhen Zheng, Qingsong Yan, Fei Deng, Kaiyong Zhao, and Xiaowen
  Chu.
\newblock \uppercase{Irs}: A large synthetic indoor robotics stereo dataset for
  disparity and surface normal estimation.
\newblock {\em arXiv e-prints}, pages arXiv--1912, 2019.

\bibitem{wang2018high}
Ting-Chun Wang, Ming-Yu Liu, Jun-Yan Zhu, Andrew Tao, Jan Kautz, and Bryan
  Catanzaro.
\newblock High-resolution image synthesis and semantic manipulation with
  conditional \uppercase{GAN}s.
\newblock In {\em IEEE Computer Vision and Pattern Recognition Conference},
  2018.

\bibitem{tartanair2020iros}
Wenshan Wang, Delong Zhu, Xiangwei Wang, Yaoyu Hu, Yuheng Qiu, Chen Wang, Yafei
  Hu, Ashish Kapoor, and Sebastian Scherer.
\newblock {TartanAir: A dataset to push the limits of visual SLAM}.
\newblock {\em International Conference on Intelligent Robots and Systems},
  2020.

\bibitem{wang2021learning}
Zian Wang, Jonah Philion, Sanja Fidler, and Jan Kautz.
\newblock Learning indoor inverse rendering with \uppercase{3D}
  spatially-varying lighting.
\newblock In {\em International Conference on Computer Vision}, 2021.

\bibitem{xia2021gan}
Weihao Xia, Yulun Zhang, Yujiu Yang, Jing-Hao Xue, Bolei Zhou, and Ming-Hsuan
  Yang.
\newblock \uppercase{Gan} inversion: A survey.
\newblock {\em arXiv preprint arXiv:2101.05278}, 2021.

\bibitem{xian2019uprightnet}
Wenqi Xian, Zhengqi Li, Matthew Fisher, Jonathan Eisenmann, Eli Shechtman, and
  Noah Snavely.
\newblock Uprightnet: Geometry-aware camera orientation estimation from single
  images.
\newblock In {\em International Conference on Computer Vision}, 2019.

\bibitem{xie2021segformer}
Enze Xie, Wenhai Wang, Zhiding Yu, Anima Anandkumar, Jose~M Alvarez, and Ping
  Luo.
\newblock {SegFormer: Simple and efficient design for semantic segmentation
  with transformers}.
\newblock {\em Conference on Neural Information Processing Systems}, 2021.

\bibitem{yang2019very}
Zongxin Yang, Jian Dong, Ping Liu, Yi Yang, and Shuicheng Yan.
\newblock Very long natural scenery image prediction by outpainting.
\newblock In {\em IEEE Computer Vision and Pattern Recognition Conference},
  2019.

\bibitem{ying2020180}
Zhenqiang Ying and Alan Bovik.
\newblock 180-degree outpainting from a single image.
\newblock {\em arXiv preprint arXiv:2001.04568}, 2020.

\bibitem{yu2018generative}
Jiahui Yu, Zhe Lin, Jimei Yang, Xiaohui Shen, Xin Lu, and Thomas~S Huang.
\newblock Generative image inpainting with contextual attention.
\newblock In {\em IEEE Computer Vision and Pattern Recognition Conference},
  2018.

\bibitem{zhang2021survey}
Yu Zhang, Peter Ti{\v{n}}o, Ale{\v{s}} Leonardis, and Ke Tang.
\newblock A survey on neural network interpretability.
\newblock {\em IEEE Transactions on Emerging Topics in Computational
  Intelligence}, 2021.

\bibitem{zhao2021comodgan}
Shengyu Zhao, Jonathan Cui, Yilun Sheng, Yue Dong, Xiao Liang, Eric~I Chang,
  and Yan Xu.
\newblock Large scale image completion via co-modulated generative adversarial
  networks.
\newblock In {\em International Conference on Learning Representations}, 2021.

\bibitem{zhou2017places}
Bolei Zhou, Agata Lapedriza, Aditya Khosla, Aude Oliva, and Antonio Torralba.
\newblock Places: A 10 million image database for scene recognition.
\newblock {\em IEEE Transactions on Pattern Analysis and Machine Intelligence},
  2017.

\bibitem{zhu2020domain}
Jiapeng Zhu, Yujun Shen, Deli Zhao, and Bolei Zhou.
\newblock In-domain \uppercase{gan} inversion for real image editing.
\newblock In {\em European Conference on Computer Vision}, 2020.

\bibitem{zhu2016generative}
Jun-Yan Zhu, Philipp Kr{\"a}henb{\"u}hl, Eli Shechtman, and Alexei~A Efros.
\newblock Generative visual manipulation on the natural image manifold.
\newblock In {\em European Conference on Computer Vision}, 2016.

\end{thebibliography}
